\documentclass[10pt,twocolumn,letterpaper]{article}

\usepackage{wacv}

\usepackage{graphicx}
\usepackage{booktabs}
\usepackage{multirow}
\usepackage{algorithm}
\usepackage{algpseudocode}
\usepackage{amsmath}
\usepackage{amssymb}
\usepackage{amsthm}   
\usepackage{tikz}
\usepackage{pgfplots}
\usepackage{wrapfig}
\usepackage{xcolor}   
\usepackage{array}
\pgfplotsset{compat=newest}
\usepgfplotslibrary{fillbetween}
\usepgfplotslibrary{groupplots,colorbrewer,polar}
\usepackage{colortbl}
\usepackage{soul}
\usepackage{adjustbox}
\usepackage{placeins}
\usetikzlibrary{positioning}
\usetikzlibrary{arrows.meta,positioning,calc}
\usetikzlibrary{patterns}

\usepackage{eccvabbrv}

\usepackage[pagebackref,breaklinks,colorlinks]{hyperref}
\usepackage[accsupp]{axessibility}

\definecolor{headerblue}{RGB}{220,230,242}
\definecolor{tableblue}{RGB}{220,230,241}
\definecolor{eccvblue}{rgb}{0.20,0.40,0.70}  
\definecolor{claimaccent}{HTML}{2F6FCE}
\definecolor{claimfill}{HTML}{EEF4FB}
\definecolor{insightaccent}{HTML}{1B7A47}
\definecolor{insightfill}{HTML}{EAF5EE}

\newcommand{\claimbox}[2]{%
  \par\smallskip\noindent
  \begin{tikzpicture}
  \node[draw=claimaccent, line width=0.9pt, fill=claimfill, rounded corners=4pt,
        inner sep=6pt, align=left, text width=\dimexpr\linewidth-16pt\relax]
    {\small\textcolor{claimaccent}{\textbf{#1.}}\ #2};
  \end{tikzpicture}%
  \par\smallskip
}

\newcommand{\insightbox}[2]{%
  \par\smallskip\noindent
  \begin{tikzpicture}
  \node[draw=insightaccent, line width=0.9pt, fill=insightfill, rounded corners=4pt,
        inner sep=5pt, align=left, text width=\dimexpr\linewidth-14pt\relax]
    {\small\textcolor{insightaccent}{\textbf{#1.}}\ #2};
  \end{tikzpicture}%
  \par\smallskip
}

\newcommand{\name}{{\textsf{Voyager}}}

\def\wacvPaperID{832} 
\def\paperID{\wacvPaperID}  
\def\confName{WACV}
\def\confYear{2027}

\begin{document}

\title{Not All Layers Are Equal: Dynamic Layer Routing for Reliable CLIP OOD Detection}

\author{
Ignacio M. De la Jara$^{1,3}$ \quad
Cristian Rodriguez-Opazo$^{2}$ \quad
Damith Ranasinghe$^{1,3}$\\
$^{1}$University of Adelaide \\
$^{2}$Australian National University \\
$^{3}$Naval Group Pacific \\
{\tt\small ignacio.mezadelajara@adelaide.edu.au}
}

\maketitle


\begin{abstract}
Information aggregation across model layers are revealed to improve ODD detection. In contrast to crafting a method for layer-wise information aggregation in recent work, we investigate if layer selection is a learnable problem. In other words, we transpose the question from how to fuse layers to one asking which layers to trust for an input. Using a generalizable, weak, out of distribution context crafting approach for supervision---shown to be more effective than state of the art methods' mechanisms---we formulate learning a lightweight router to select a sparse, final-layer-anchored expert, over CLIP's layer depth for OOD detection. 
Across three diverse benchmarks we demonstrate our learnable routing method dubbed \name{} improves OOD detection. On ImageNet-1K, \name{} achieves an average FPR@95 of \textbf{18.86}, outperforming the strongest, comparable, prompt-learning method by $8.8$ points. These gains persist across multiple supervision sources, including those used by existing state-of-the-art prompt-learning methods, demonstrating, whilst our weak ODD supervision context is highly effective, the key advantage is realized from the learnable router component rather than the supervision source. Significantly, \name{} is highly practical---router learning takes approximately two minutes using less than 1\,GB of memory, making it $\approx20\times$ more efficient than current prompt-learning approaches. Anonymized Code: \url{https://anonymous.4open.science/r/Voyager/}.
\end{abstract}
\section{Introduction}

Out-of-distribution (OOD) detection is essential when vision systems face open-world inputs that do not belong to their intended label space. Vision-language models such as CLIP~\cite{radford2021learning} are attractive in this setting because they provide strong zero-shot image-text confidence scores without task-specific backbone training~\cite{Esmaeilpour_2022, ming2022delvingoutofdistributiondetectionvisionlanguage, wang2023clipnzeroshotooddetection, miyai2023locoopfewshotoutofdistributiondetection}. Most CLIP OOD detectors, however, still behave as if the model were shallow: they score the final representation and discard the sequence of intermediate states that produced it.

\begin{figure}[t]
    \centering
    \includegraphics[width=\linewidth,trim=30pt 12pt 34pt 12pt,clip]{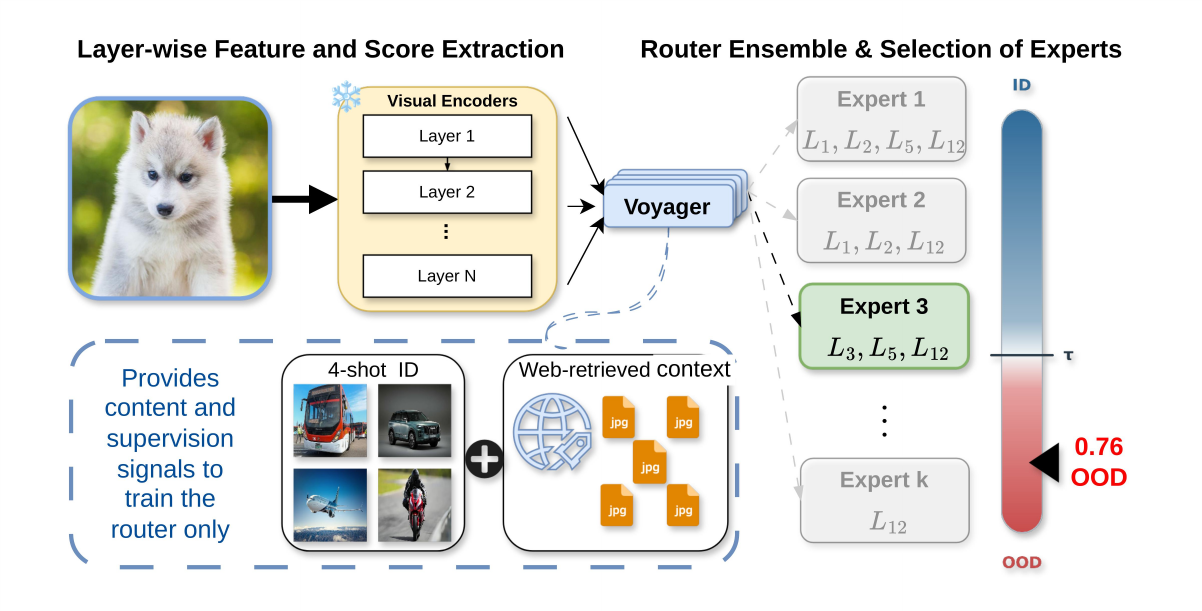}
    \vspace{-0.8em}
    \caption{
        \textbf{Main idea.} \name{} treats CLIP depth as an input-conditioned choice: a lightweight router trained with few-shot ID data and our WebOE retrieval-based weak OOD supervision selects a sparse layer-subset expert for each image.
    }
    \label{fig:intro_main}
    \vspace{-2em}
\end{figure}

This final-layer view is increasingly incomplete. Intermediate representations have long been useful for OOD detection in discriminative networks~\cite{lee2018simpleunifiedframeworkdetecting}, and recent CLIP analyses show that different depths encode complementary, shift-sensitive evidence~\cite{uselis2025intermediatelayerclassifiersood, jelenic2024outofdistributiondetectionleveragingbetweenlayer, delajara2025mysteriesdeeproleintermediate}. \emph{Mysteries of the Deep}~\cite{delajara2025mysteriesdeeproleintermediate} demonstrates that fixed fusion of intermediate CLIP layers can outperform final-layer MCM. Yet its fusion rule is static: once a subset is selected, the same layer combination is used for every image and every shift. Prompt-learning methods such as CoOp, LoCoOp, ID-Like, and Local-Prompt~\cite{Zhou_2022, miyai2023locoopfewshotoutofdistributiondetection, bai2024idlikepromptlearningfewshot, zeng2025localpromptextensiblelocalprompts} adapt CLIP with few examples, and several of them manufacture OOD context to sharpen the decision boundary---but that context is synthesized purely from ID data, and they still act on prompts or final-layer scores rather than on the depth at which evidence is read. Outlier Exposure~\cite{hendrycks2019deepanomalydetectionoutlier} shows that exposing a detector to genuine external outliers is an effective alternative; we adopt this principle, but instead of curating an outlier set by hand we \emph{automatically retrieve real web images} as weak context. As we show later (Sec.~\ref{main-results}), simply handing this external context to prompt-learning detectors barely helps---a limitation rooted in how they are trained---whereas using it to choose which frozen layers to read clearly does.

\claimbox{Observation 1: No single set of layers works for every image}
{There is no universally best CLIP layer subset for OOD detection. Some images are already well separated by the final layer, while others need specific intermediate cues; the right layers should therefore be chosen per image rather than fixed once for the whole dataset.}

The retrospective diagnostics in Sec.~\ref{sec:motivation} make this limitation concrete without treating an oracle as a deployable baseline. Figure~\ref{fig:layer_contribution_radar} shows that useful depths are structured but shift dependent: both ViT-B/16 and ViT-B/32 keep the final layer as a semantic anchor, while their auxiliary-layer usage differs by backbone and OOD shift. Figure~\ref{fig:margin-dist} then shows that choosing the expert separately for each image can push the ID--OOD margin beyond any fixed expert, indicating that the missing ingredient is \emph{choosing the layers per image} rather than merely adding more layers.

\claimbox{Observation 2: Retrieved OOD context should guide the choice, not become the score}
{The retrieved web images should teach the router \emph{which frozen expert to trust} for a given image; they should not be turned into a test-time OOD score. Keeping the two separate isolates the benefit of the choice itself and avoids reducing the method to a classifier of the retrieved set.}

\begin{figure*}[t]
    \centering
    \includegraphics[width=\textwidth]{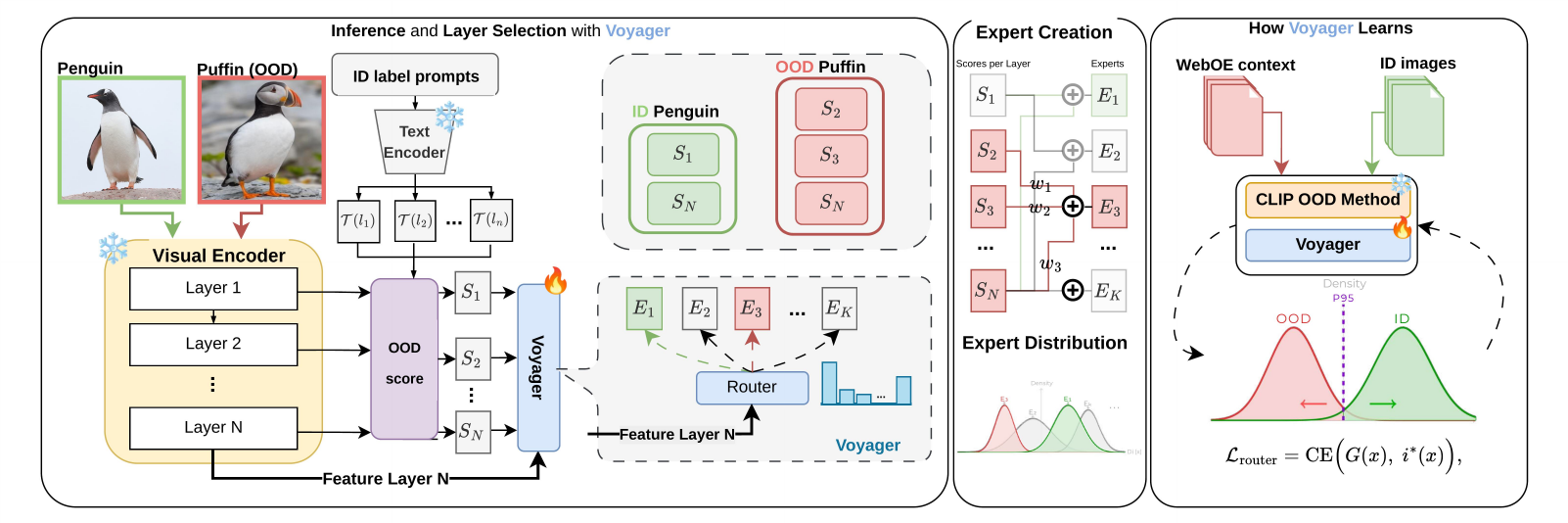}
    \vspace{-1.0em}
    \caption{
        \textbf{\name{} method overview.} Frozen CLIP produces layer-wise OOD scores. A lightweight router selects one sparse, final-layer-anchored expert, and the selected expert score becomes the final detector score.
    }
    \label{fig:dyrod_overview}
    \vspace{-0.8em}
\end{figure*}

We introduce \name{}, a lightweight dynamic layer-routing plug-in for frozen CLIP. \name{} is not another few-shot prompt model: it leaves the prompts untouched and instead acts one level deeper, on \emph{which intermediate layers} produce the score for each image. Figure~\ref{fig:intro_main} gives the high-level idea, and Figure~\ref{fig:dyrod_overview} details the deployed architecture. \name{} builds a dictionary of sparse, final-layer-anchored experts, where each expert averages layer-wise CLIP OOD scores over one layer subset. A small router then maps each input to one expert. The CLIP image encoder, text encoder, prompts, and expert scoring functions remain frozen; only the router is learned. To train this router without target-OOD examples, we also introduce \textbf{WebOE}, a retrieval-based weak OOD supervision procedure that uses the ID class names to retrieve WordNet-filtered Wikimedia Commons images. WebOE supplies only a binary domain label for routing supervision; no WebOE image is used as a test-time score, and no target-OOD test instance is used to build the proxy.

Our contributions are: 
\begin{itemize} 
    \item We show that CLIP OOD detection benefits from \textbf{input-conditioned layer selection}, demonstrating that dynamic routing consistently outperforms fixed intermediate-layer fusion. 
    \item We propose \textbf{\name{}} and \textbf{WebOE}: \name{} routes each image to a sparse, final-layer-anchored expert, while WebOE provides retrieval-based weak supervision for learning the routing policy. 
    \item We introduce a stable routing strategy based on top-$k$ expert supervision, strict separation between routing supervision and inference, and a lightweight router ensemble. 
    \item Across different benchmarks, \name{} achieves the best average FPR@95 among zero-shot, prompt-learning, and few-shot CLIP baselines while training roughly $20\times$ more efficiently than prompt-learning methods. Learned routing further outperforms the best score-based selector by ${\sim}10$ FPR@95 and fixed-context experts by ${\sim}16$. 
\end{itemize}

\section{Related Work}

\noindent \textbf{Out-of-distribution Detection.}
OOD detection identifies inputs that deviate from the training distribution, a key challenge for safe deployment. Early approaches rely on confidence scores such as MSP~\cite{hendrycks2018baselinedetectingmisclassifiedoutofdistribution} and ODIN~\cite{liang2020enhancingreliabilityoutofdistributionimage}, and statistical distances like Mahalanobis scoring~\cite{lee2018simpleunifiedframeworkdetecting} or Energy-based scoring~\cite{liu2020energybasedoutofdistribution}. Outlier Exposure~\cite{hendrycks2019deepanomalydetectionoutlier} improves robustness by training models to assign low confidence to a small set of external non-ID samples, effectively regularizing decision boundaries and enhancing OOD detection. Additional advances explore residual features~\cite{wang2022vimoutofdistributionvirtuallogitmatching}, hyperspherical embeddings~\cite{ming2023exploithypersphericalembeddingsoutofdistribution}, and activation shaping~\cite{djurisic2023extremelysimpleactivationshaping}. Foundational analyses examine learnability and scoring principles~\cite{bitterwolf2022breaking, fang2022learnable, morteza2022provable}, while modality-specific taxonomies~\cite{arora2021types} contextualize anomaly types.

\noindent \textbf{Vision--language Models for OOD Detection.}
Large vision--language models such as CLIP~\cite{radford2021learningtransferablevisualmodels} enable training-free OOD scoring via image--text similarity. Extensions include concept matching~\cite{ming2022delvingoutofdistributiondetectionvisionlanguage, miyai2025glmcmgloballocalmaximum} and prompt tuning~\cite{miyai2023locoopfewshotoutofdistributiondetection, li2024learningtransferablenegativeprompts}. Synthetic outlier generation~\cite{bai2024idlikepromptlearningfewshot, cao2024envisioningoutlierexposurelarge} has also been explored, showing that prompt-based models can learn from ID-derived synthetic anomalies but struggle to benefit from external data: related or unrelated outliers provide only marginal gains, limiting their effectiveness for broader OOD generalization. Training-free approaches such as SeTAR~\cite{li2024setaroutofdistributiondetectionselective} avoid prompt learning but still rely on an optimization procedure that requires labeled data and a dedicated loss function.

\noindent \textbf{Intermediate-layer Representations.}
Early distance-based detectors used intermediate features~\cite{lee2018simpleunifiedframeworkdetecting}, but most modern approaches rely on final-layer outputs. Recent work shows that intermediate CLIP layers encode transferable, shift-stable signals~\cite{uselis2025intermediatelayerclassifiersood, jelenic2024outofdistributiondetectionleveragingbetweenlayer}, often with added supervision or retraining. Lin et al.~\cite{lin2021moodmultileveloutofdistributiondetection} introduce multi-exit energy scoring for closed-set classifiers, while Fayyad et al.~\cite{exploitin_classifier_interlevel} attach supervised auxiliary heads. Guglielmo and Masana~\cite{Leveraging_Intermediate_Representations} rank layers to choose a single discriminative depth on small-scale datasets rather than fusing them. \emph{Mysteries of the Deep}~\cite{delajara2025mysteriesdeeproleintermediate} shows that aggregating intermediate layers improves training-free OOD detection, but uses a fixed fusion rule that cannot adapt to input-dependent variability.


\section{Motivation: Why Static Fusion Fails}
\label{sec:motivation}

OOD detection with CLIP commonly uses Maximum Concept Matching (MCM)~\cite{ming2022delvingoutofdistributiondetectionvisionlanguage}: an image embedding is compared with prompt-based text prototypes, a softmax is applied over ID classes, and the maximum confidence is used as the ID score. Recent work shows that applying this idea beyond the final layer can improve detection because CLIP's intermediate representations retain low-level, semantic, and shift-specific cues~\cite{delajara2025mysteriesdeeproleintermediate}. The remaining question is whether a \emph{fixed} layer subset can fully exploit this diversity.

\claimbox{Question: Is one layer subset enough?}
{If the best intermediate evidence changes across inputs and OOD shifts, static fusion can improve over the final layer but still leave substantial performance unused.}

\subsection{Limitations of Static Layer Selection}

\definecolor{radarColor0}{RGB}{102,194,165}
\definecolor{radarColor1}{RGB}{252,141,98}
\definecolor{radarColor2}{RGB}{141,160,203}
\definecolor{radarColor3}{RGB}{231,138,195}

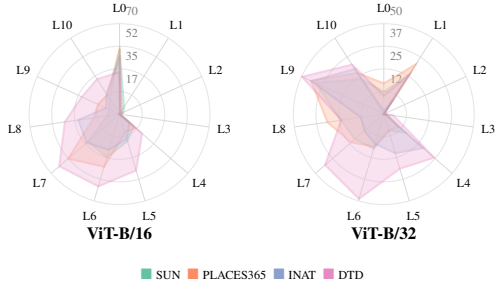
\begin{figure}[t]
    \centering
    \begin{tikzpicture}[scale=0.92, transform shape]
    
    \def\numaxes{11}
    \def\maxval{70}
    \def\radius{1.3}
    
    \begin{scope}[shift={(-1.9,0)}]
        \foreach \r in {0, 17.5, 35, 52.5, 70} {
            \pgfmathsetmacro\circrad{\r/\maxval*\radius}
            \draw[gray!20, thin] (0,0) circle (\circrad);
            \ifdim\r pt>0pt
                \pgfmathtruncatemacro\rint{\r}
                \node[gray, font=\tiny] at (90:\circrad) [above right, xshift=-1pt] {\rint};
            \fi
        }
        \foreach \i in {0,...,10} {
            \pgfmathsetmacro\angle{90 - \i*360/\numaxes}
            \draw[gray!30, very thin] (0,0) -- (\angle:\radius);
        }
        \foreach \i in {0,...,10} {
            \pgfmathsetmacro\angle{90 - \i*360/\numaxes}
            \node[font=\tiny] at (\angle:\radius+0.22) {L\i};
        }
        
        \pgfmathsetmacro\aid{90}
        \pgfmathsetmacro\bid{90-1*360/11}
        \pgfmathsetmacro\cid{90-2*360/11}
        \pgfmathsetmacro\did{90-3*360/11}
        \pgfmathsetmacro\eid{90-4*360/11}
        \pgfmathsetmacro\fid{90-5*360/11}
        \pgfmathsetmacro\gid{90-6*360/11}
        \pgfmathsetmacro\hid{90-7*360/11}
        \pgfmathsetmacro\iid{90-8*360/11}
        \pgfmathsetmacro\jid{90-9*360/11}
        \pgfmathsetmacro\kid{90-10*360/11}

        \draw[fill=radarColor0!60, draw=radarColor0, thick, opacity=0.40]
            (\aid:49.18/\maxval*\radius) -- (\bid:6.33/\maxval*\radius) --
            (\cid:3.08/\maxval*\radius) -- (\did:0.07/\maxval*\radius) --
            (\eid:14.37/\maxval*\radius) -- (\fid:21.67/\maxval*\radius) --
            (\gid:34.81/\maxval*\radius) -- (\hid:32.19/\maxval*\radius) --
            (\iid:9.41/\maxval*\radius) -- (\jid:8.74/\maxval*\radius) --
            (\kid:14.31/\maxval*\radius) -- cycle;
        \draw[fill=radarColor1!60, draw=radarColor1, thick, opacity=0.40]
            (\aid:49.99/\maxval*\radius) -- (\bid:2.79/\maxval*\radius) --
            (\cid:2.47/\maxval*\radius) -- (\did:0.0/\maxval*\radius) --
            (\eid:16.30/\maxval*\radius) -- (\fid:14.74/\maxval*\radius) --
            (\gid:42.45/\maxval*\radius) -- (\hid:52.50/\maxval*\radius) --
            (\iid:19.96/\maxval*\radius) -- (\jid:18.31/\maxval*\radius) --
            (\kid:17.18/\maxval*\radius) -- cycle;
        \draw[fill=radarColor2!60, draw=radarColor2, thick, opacity=0.40]
            (\aid:43.00/\maxval*\radius) -- (\bid:0.18/\maxval*\radius) --
            (\cid:0.22/\maxval*\radius) -- (\did:0.0/\maxval*\radius) --
            (\eid:10.25/\maxval*\radius) -- (\fid:19.91/\maxval*\radius) --
            (\gid:28.77/\maxval*\radius) -- (\hid:33.70/\maxval*\radius) --
            (\iid:32.24/\maxval*\radius) -- (\jid:11.46/\maxval*\radius) --
            (\kid:18.67/\maxval*\radius) -- cycle;
        \draw[fill=radarColor3!60, draw=radarColor3, thick, opacity=0.40]
            (\aid:32.59/\maxval*\radius) -- (\bid:1.15/\maxval*\radius) --
            (\cid:0.53/\maxval*\radius) -- (\did:0.02/\maxval*\radius) --
            (\eid:23.03/\maxval*\radius) -- (\fid:45.44/\maxval*\radius) --
            (\gid:58.24/\maxval*\radius) -- (\hid:61.58/\maxval*\radius) --
            (\iid:42.70/\maxval*\radius) -- (\jid:32.06/\maxval*\radius) --
            (\kid:32.09/\maxval*\radius) -- cycle;

        \node[font=\scriptsize\bfseries] at (0,-1.72) {ViT-B/16};
    \end{scope}

    \def\maxvalR{50}
    \begin{scope}[shift={(1.9,0)}]
        \foreach \r in {0, 12.5, 25, 37.5, 50} {
            \pgfmathsetmacro\circrad{\r/\maxvalR*\radius}
            \draw[gray!20, thin] (0,0) circle (\circrad);
            \ifdim\r pt>0pt
                \pgfmathtruncatemacro\rint{\r}
                \node[gray, font=\tiny] at (90:\circrad) [above right, xshift=-1pt] {\rint};
            \fi
        }
        \foreach \i in {0,...,10} {
            \pgfmathsetmacro\angle{90 - \i*360/\numaxes}
            \draw[gray!30, very thin] (0,0) -- (\angle:\radius);
        }
        \foreach \i in {0,...,10} {
            \pgfmathsetmacro\angle{90 - \i*360/\numaxes}
            \node[font=\tiny] at (\angle:\radius+0.22) {L\i};
        }

        \pgfmathsetmacro\aid{90}
        \pgfmathsetmacro\bid{90-1*360/11}
        \pgfmathsetmacro\cid{90-2*360/11}
        \pgfmathsetmacro\did{90-3*360/11}
        \pgfmathsetmacro\eid{90-4*360/11}
        \pgfmathsetmacro\fid{90-5*360/11}
        \pgfmathsetmacro\gid{90-6*360/11}
        \pgfmathsetmacro\hid{90-7*360/11}
        \pgfmathsetmacro\iid{90-8*360/11}
        \pgfmathsetmacro\jid{90-9*360/11}
        \pgfmathsetmacro\kid{90-10*360/11}

        \draw[fill=radarColor0!60, draw=radarColor0, thick, opacity=0.40]
            (\aid:12.31/\maxvalR*\radius) -- (\bid:25.71/\maxvalR*\radius) --
            (\cid:0/\maxvalR*\radius) -- (\did:1.21/\maxvalR*\radius) --
            (\eid:15.47/\maxvalR*\radius) -- (\fid:9.48/\maxvalR*\radius) --
            (\gid:19.83/\maxvalR*\radius) -- (\hid:21.07/\maxvalR*\radius) --
            (\iid:23.89/\maxvalR*\radius) -- (\jid:41.38/\maxvalR*\radius) --
            (\kid:23.04/\maxvalR*\radius) -- cycle;
        \draw[fill=radarColor1!60, draw=radarColor1, thick, opacity=0.40]
            (\aid:16.91/\maxvalR*\radius) -- (\bid:32.61/\maxvalR*\radius) --
            (\cid:0.01/\maxvalR*\radius) -- (\did:1.43/\maxvalR*\radius) --
            (\eid:10.71/\maxvalR*\radius) -- (\fid:9.61/\maxvalR*\radius) --
            (\gid:18.69/\maxvalR*\radius) -- (\hid:23.78/\maxvalR*\radius) --
            (\iid:31.45/\maxvalR*\radius) -- (\jid:44.10/\maxvalR*\radius) --
            (\kid:26.97/\maxvalR*\radius) -- cycle;
        \draw[fill=radarColor2!60, draw=radarColor2, thick, opacity=0.40]
            (\aid:1.61/\maxvalR*\radius) -- (\bid:9.69/\maxvalR*\radius) --
            (\cid:0/\maxvalR*\radius) -- (\did:0.84/\maxvalR*\radius) --
            (\eid:28.63/\maxvalR*\radius) -- (\fid:22.48/\maxvalR*\radius) --
            (\gid:17.11/\maxvalR*\radius) -- (\hid:14.12/\maxvalR*\radius) --
            (\iid:12.75/\maxvalR*\radius) -- (\jid:43.85/\maxvalR*\radius) --
            (\kid:29.62/\maxvalR*\radius) -- cycle;
        \draw[fill=radarColor3!60, draw=radarColor3, thick, opacity=0.40]
            (\aid:10.23/\maxvalR*\radius) -- (\bid:25.69/\maxvalR*\radius) --
            (\cid:0.02/\maxvalR*\radius) -- (\did:5.27/\maxvalR*\radius) --
            (\eid:36.72/\maxvalR*\radius) -- (\fid:31.42/\maxvalR*\radius) --
            (\gid:48.79/\maxvalR*\radius) -- (\hid:42.94/\maxvalR*\radius) --
            (\iid:23.65/\maxvalR*\radius) -- (\jid:49.59/\maxvalR*\radius) --
            (\kid:32.45/\maxvalR*\radius) -- cycle;

        \node[font=\scriptsize\bfseries] at (0,-1.72) {ViT-B/32};
    \end{scope}

    \node[font=\tiny, anchor=center] at (0,-2.3) {%
        \textcolor{radarColor0}{$\blacksquare$}~SUN\enspace
        \textcolor{radarColor1}{$\blacksquare$}~PLACES365\enspace
        \textcolor{radarColor2}{$\blacksquare$}~INAT\enspace
        \textcolor{radarColor3}{$\blacksquare$}~DTD
    };

    \end{tikzpicture}
    \vspace{-0.8em}
    \caption{Retrospective layer-selection diagnostic: percentage of times each layer is selected per OOD dataset.}
    \label{fig:layer_contribution_radar}
    \vspace{-0.8em}
\end{figure}

We first examine which layers would be selected by a retrospective diagnostic under heterogeneous distribution shifts. For each input image, we select the layer subset that gives the strongest ID--OOD separation under MCM, using test labels only for analysis, and summarize the selected layers per OOD dataset. Figure~\ref{fig:layer_contribution_radar} shows interpretable depth signatures rather than a uniform preference for the final representation. ViT-B/16 often assigns mass to early and mid-depth layers, especially for texture-centric shifts such as DTD, while scene-oriented shifts distribute mass more smoothly. ViT-B/32 shows a different profile, with stronger concentration on late layers. Thus OOD evidence is distributed across depth, varies with anomaly type, and is architecture-specific.

\subsection{Routing Sparsity}
\label{sec:impact_sparsity}

To quantify the opportunity created by depth-adaptive routing, we define a retrospective selector that chooses, for each input $x$, a layer subset $\mathcal{L}^\star(x)\in\mathcal{C}_s$ under a sparsity budget $s$. This selector is an analysis tool: it uses labels unavailable at deployment and is not compared as a practical method. The candidate set $\mathcal{C}_s$ always includes the final layer and allows up to $s-1$ auxiliary intermediate layers.

Figure~\ref{fig:margin-dist} (right) makes this concrete on ViT-B/16: as $s$ grows the deployed router's FPR@95 keeps falling (to $18.86$ at $s{=}9$) while the best fixed expert plateaus near $33.5$ --- the gain comes from per-input choice, not a larger pool (full per-dataset retrospective breakdown in App.~\ref{appendix:extra-results-complementarity}, Fig.~\ref{fig:sparsity_vs_fpr}).

\input{plots/margin_figure_plot_vit16and32}

\subsection{Margin-Based Separability}
\label{sec:margin_analysis}

Layer frequency alone does not quantify how much routing changes the ID--OOD decision boundary. We therefore analyze MCM margin distributions. For each sample, let $m(x)=\tau-s(x)$, where $\tau$ is an ID boundary estimated from in-distribution data and $s(x)$ is the score; larger positive margins indicate stronger OOD separation.

Figure~\ref{fig:margin-dist} (left) compares final-layer MCM, static fusion (MoD)~\cite{delajara2025mysteriesdeeproleintermediate}, the best static expert, and the retrospective per-instance selector. MoD shifts mass toward positive margins (intermediate layers help) and the best static expert improves further; the retrospective selector shows the same pool of subsets still holds additional input-specific signal.


\insightbox{Motivation insight: Selection matters beyond fusion}
{Intermediate layers are useful, but fixed fusion does not use them equally well for every input. The retrospective diagnostics motivate learning a router, rather than only designing another fixed fusion heuristic.}

\section{Methodology}
\label{sec:methodology}

We consider OOD detection with a frozen CLIP-style model and an ID label set $\mathcal{Y}=\{1,\dots,C\}$. Training uses a few-shot ID set $\mathcal{D}_{\mathrm{ID}}$ and a weak OOD-context set $\mathcal{D}_{\mathrm{ctx}}$ used only to learn routing. The context set is built without target-OOD labels or test instances. Each training sample has a binary domain label
\[
y(x)=0 \ \text{if } x\in\mathcal{D}_{\mathrm{ID}}, \qquad
y(x)=1 \ \text{if } x\in\mathcal{D}_{\mathrm{ctx}} .
\]

\subsection{Layer-Wise Frozen Scores}
\label{sec:backbone_scores}

For a frozen vision encoder with $L$ layers, let $f_\ell(x)\in\mathbb{R}^{d}$ be the normalized global image representation at depth $\ell$, and let $\mathcal{T}=\{t_c\}_{c=1}^{C}$ be normalized text prototypes from the frozen text encoder. For layer $\ell$, MCM-style logits are
\[
z_{\ell,c}(x)=\gamma\langle f_\ell(x),t_c\rangle,
\qquad
p_\ell(c\mid x)=\frac{\exp z_{\ell,c}(x)}{\sum_{c'}\exp z_{\ell,c'}(x)}.
\]
The layer-wise confidence is
\begin{equation}
S_\ell(x)=\max_{c\in\mathcal{Y}} p_\ell(c\mid x).
\label{eq:mcm_layer_score}
\end{equation}
In the main ImageNet-1K experiments we use the same construction with GL-MCM scores, adding the local patch term described in Sec.~\ref{subsec:setting}; the routing machinery is unchanged.

\subsection{Anchored Sparse Experts}
\label{sec:experts}

Let $\ell^\star=L-1$ be the final layer. Each expert is a sparse subset of layers that always includes $\ell^\star$, preserving CLIP's high-level semantic anchor while allowing intermediate evidence to contribute. With sparsity budget $s$, the expert dictionary is
\begin{equation}
\Lambda(L,s)=
\left\{
A\cup\{\ell^\star\} :
A\subseteq\{0,\dots,L-2\},\ |A|\le s-1
\right\}.
\label{eq:expert_set}
\end{equation}
Thus
\begin{equation}
K=|\Lambda(L,s)|=\sum_{r=0}^{s-1}\binom{L-1}{r}.
\label{eq:expert_count_main}
\end{equation}
Each expert $i$ with layer set $L_i\in\Lambda(L,s)$ returns the frozen score
\begin{equation}
\mathrm{E}_i(x)=
\frac{1}{|L_i|}\sum_{\ell\in L_i} S_\ell(x).
\label{eq:expert_score}
\end{equation}
We use mean aggregation to isolate the effect of \emph{selection}; more complex fusion rules are unnecessary for defining the expert family.

\subsection{ID Boundary Normalization}
\label{sec:warmup}

Raw expert scores are not directly comparable because different layer subsets have different ID score scales. For each expert, we estimate a 95\% ID-retention boundary using only $\mathcal{D}_{\mathrm{ID}}$:
\begin{align}
b_i &= Q_{0.05}\big(\{\mathrm{E}_i(x):x\in\mathcal{D}_{\mathrm{ID}}\}\big),\\
\sigma_i&=\mathrm{Std}\big(\{\mathrm{E}_i(x):x\in\mathcal{D}_{\mathrm{ID}}\}\big).
\end{align}
Since higher confidence is more ID-like, $b_i$ is the low-tail ID threshold that keeps approximately 95\% of ID samples above the boundary. We define the normalized signed distance
\begin{equation}
D_i(x)=\frac{\mathrm{E}_i(x)-b_i}{\sigma_i+\epsilon}.
\label{eq:normalized_distance}
\end{equation}
Positive values are ID-like for expert $i$; negative values indicate that the sample falls below that expert's ID boundary.

\subsection{Learning the Router}
\label{sec:routing}

The router $g_\theta:\mathbb{R}^{d_r}\rightarrow\Delta^{K-1}$ takes frozen image features $\phi(x)$, implemented as a fixed aggregation of layer features, and outputs an expert distribution $G_\theta(x)=g_\theta(\phi(x))$. Importantly, the router does not receive the test label or any test-time OOD context; context data are used only to form the training objective.

For a labelled training sample, we convert expert distances into a policy metric:
\begin{equation}
M_i(x)=
\begin{cases}
D_i(x), & y(x)=0\quad(\mathrm{ID}),\\
-D_i(x), & y(x)=1\quad(\mathrm{OOD\ context}).
\end{cases}
\label{eq:policy_metric}
\end{equation}
This metric selects experts that keep ID samples above their ID boundary and push context OOD samples below it. A hard target $\arg\max_i M_i(x)$ is brittle because many sparse experts are near-tied, so we use an FPR@95-aligned soft target restricted to the top-$k$ near-optimal experts. Let $Z_k(x)=\sum_{j\in\mathcal{N}_k(x)}\exp(M_j(x)/\tau_{\mathrm{tgt}})$. Then
\begin{equation}
q_i(x)=
\begin{cases}
\exp(M_i(x)/\tau_{\mathrm{tgt}})/Z_k(x), & i\in\mathcal{N}_k(x),\\
0, & i\notin\mathcal{N}_k(x),
\end{cases}
\label{eq:soft_target}
\end{equation}
where $\mathcal{N}_k(x)$ is the set of the $k$ experts with the largest $M_i(x)$.
We use $k{=}32$ in the deployed detector (ablated in App.~\ref{appendix:target}); $k{=}1$ recovers the brittle hard target and $k{=}K$ the fully soft target.
Here $\tau_{\mathrm{tgt}}$ spreads probability mass across the $k$ near-best experts. The routing loss is
\begin{equation}
\mathcal{L}_{\mathrm{route}}(\theta)=
-\mathbb{E}_{x\sim\mathcal{D}_{\mathrm{ID}}\cup\mathcal{D}_{\mathrm{ctx}}}
\sum_{i=1}^{K}q_i(x)\log [G_\theta(x)]_i .
\label{eq:router_loss}
\end{equation}
The implemented objective is
\begin{equation}
\mathcal{L}=\mathcal{L}_{\mathrm{route}}+\mathcal{L}_{\mathrm{sep}}+0.01\,\mathcal{L}_{\mathrm{bal}},
\label{eq:total_router_loss}
\end{equation}
with constant weights throughout training. The auxiliary separation term is differentiable and is applied to the router-weighted normalized distance, not to the hard test-time route:
\begin{equation}
\widetilde{D}_{\theta}(x)=
\sum_{i=1}^{K}[G_\theta(x)]_i\,D_i(x).
\label{eq:soft_training_score}
\end{equation}
For a mini-batch $B$, the direct separation term is
\begin{equation}
\begin{aligned}
\mathcal{L}_{\mathrm{sep}}=
&\frac{1}{|B_{\mathrm{ID}}|}\sum_{x\in B_{\mathrm{ID}}}
(\widetilde{D}_{\theta}(x)-1)^2\\
&+\frac{1}{|B_{\mathrm{ctx}}|}\sum_{x\in B_{\mathrm{ctx}}}
(\widetilde{D}_{\theta}(x)+1)^2,
\end{aligned}
\label{eq:separation_loss}
\end{equation}
where positive $\widetilde{D}_{\theta}$ is ID-like and negative $\widetilde{D}_{\theta}$ is context-OOD-like. This term is used only during training; no straight-through estimator is used. The load-balancing term discourages expert collapse:
\begin{equation}
\begin{aligned}
\mathcal{L}_{\mathrm{bal}}&=\sum_{i=1}^{K} f_i p_i,\\
p_i&=\frac{1}{|B|}\sum_{x\in B}[G_\theta(x)]_i,\\
f_i&=\frac{1}{|B|}\sum_{x\in B}
\mathbf{1}\{\arg\max_j [G_\theta(x)]_j=i\}.
\end{aligned}
\label{eq:balance_loss}
\end{equation}
We use target temperature $\tau_{\mathrm{tgt}}{=}2.0$ and no loss-weight schedule. These terms define the router training only; the backbone and expert scores remain frozen.

\subsection{Inference, Ensembling, and Cost}
\label{sec:inference}

At test time, \name{} computes the layer scores once, evaluates all sparse expert scores by cheap reductions, and selects
\begin{equation}
\widehat{i}(x)=\arg\max_i [G_\theta(x)]_i,
\qquad
\mathrm{Score}(x)=\mathrm{E}_{\widehat{i}(x)}(x).
\label{eq:final_score}
\end{equation}
No WebOE sample, OOD context image, or target-domain OOD example is used at inference.

\paragraph{Router ensemble.}
The deployed \name{} detector uses a small score-level ensemble of $R$ routers (giving deterministic test-time inference once trained), all trained with the same frozen features, expert dictionary, and routing objective, but with independent random initializations and data order. For router $r$, let
\[
\widehat{i}_r(x)=\arg\max_i [G_{\theta_r}(x)]_i,
\qquad
\mathrm{Score}_r(x)=\mathrm{E}_{\widehat{i}_r(x)}(x).
\]
The final OOD score is the arithmetic mean of the selected expert scores:
\begin{equation}
\mathrm{Score}_{\mathrm{ens}}(x)=
\frac{1}{R}\sum_{r=1}^{R}\mathrm{Score}_r(x),
\qquad R=5 \text{ by default}.
\label{eq:ensemble_score}
\end{equation}
This score-level ensemble keeps the interpretation of each member as a hard layer-subset route, reduces seed variance, and does not require additional CLIP forward passes because all routers share the same precomputed layer scores and expert table.

\noindent\textbf{Cost.}
The dominant computation is the frozen CLIP forward pass and class-similarity scores already needed by the base detector. Expert evaluation is a matrix of sparse averages, and routing is a small MLP forward pass. Appendix Sec.~\ref{sec:overhead} reports the resulting latency overhead.

\section{WebOE: Retrieval-Based Weak OOD Supervision}
\label{sec:weboe}

\name{} needs weak OOD context only to learn which frozen expert to route to. We introduce \textbf{WebOE}, a retrieval-based weak OOD supervision procedure that can be instantiated for any labeled ID distribution. Given the ID class names, WebOE uses WordNet filtering and basic image-quality checks to select semantically separated web labels, retrieves balanced Wikimedia Commons images for those labels, and uses the images only as binary routing context ($y{=}1$). WebOE images are never used as a test-time score, and no target-OOD test instance is used to build the proxy.

Concretely, WebOE has three stages (Appendix Algorithm~\ref{alg:weboe}). First, we map the ID labels and candidate web labels to WordNet synsets, then discard candidates that overlap with, or are immediate neighbors of, an ID synset. We keep candidates whose shortest-path distance to the ID label set is at least two hops, reducing direct semantic leakage while retaining nearby categories that can act as challenging context. Second, we query Wikimedia Commons with the retained lemmas and keep valid images that pass basic resolution and file-integrity checks. Sampling is capped per label/query so the proxy set is not dominated by a few visual concepts. Third, we cache frozen CLIP features and per-layer scores for the retrieved images and train the router with binary domain labels only. Thus WebOE supervises \emph{which frozen layer expert to trust}; it does not introduce new class labels, tune prompts, alter CLIP, or appear at test time.

\begin{figure}[t]
    \centering
    \includegraphics[width=\linewidth]{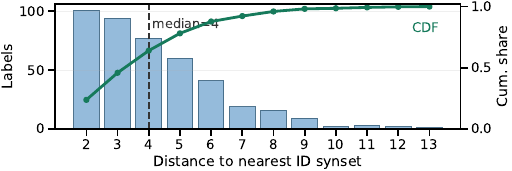}
    \vspace{-0.7em}
    \caption{\textbf{WebOE label filtering.} Bars count candidate WebOE labels by shortest-path distance to the ID label set, and the green curve shows cumulative share. In our ImageNet-1K instance, retained labels have no direct ID overlap and are at least two WordNet hops away.}
    \label{fig:weboe_construction}
    \label{fig:wordnet_dist}
    \vspace{-1.0em}
\end{figure}

Figure~\ref{fig:weboe_construction} summarizes the label-filtering step. WordNet filtering encourages semantic separation from ID classes, while still retaining nearby concepts that can act as challenging near-OOD context. Example retrieved images are shown in Appendix Fig.~\ref{fig:examplesofimages}. Since unknown OOD shifts cannot be known at training time, our claim is not semantic disjointness from every possible evaluation concept; instead, WebOE is constructed from ID labels alone and its transfer is tested by the selector ablations (Appendix Sec.~\ref{appendix:selector}).

\begin{figure}[t]
    \centering
    \captionsetup[subfigure]{skip=1pt}
    \begin{subfigure}{0.49\linewidth}
        \centering
        \includegraphics[width=\linewidth]{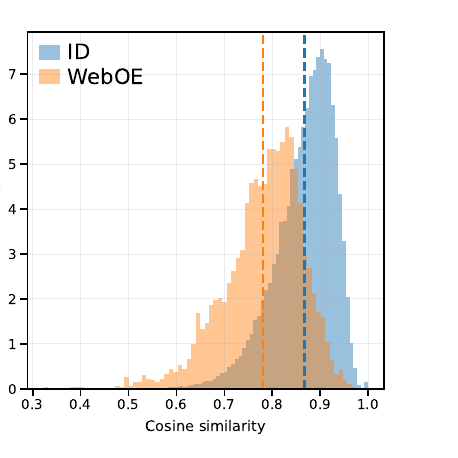}
        \caption{Retrieval similarity}
        \label{fig:sim_b16}
    \end{subfigure}
    \hspace{-0.10\linewidth}
    \begin{subfigure}{0.49\linewidth}
        \centering
        \includegraphics[width=\linewidth]{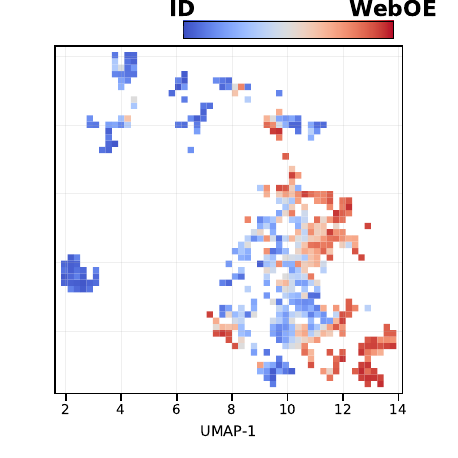}
        \caption{Local structure (UMAP)}
        \label{fig:umap_b16}
    \end{subfigure}
    \vspace{-0.35em}
    \caption{\textbf{WebOE retrieval analysis (ViT-B/16).} WebOE samples lie near the ID manifold but are less similar than ID matches. UMAP colors indicate ID-dominated (blue), WebOE-dominated (red), and mixed (pale) neighborhoods. ViT-B/32 is similar (Appendix Fig.~\ref{fig:combined_visualizations_b32}).}
    \label{fig:combined_visualizations}
    \vspace{-1.0em}
\end{figure}

Figure~\ref{fig:combined_visualizations} explains why this weak supervision is useful. WebOE samples are not arbitrary negatives: in CLIP space they remain near the ID manifold while shifting away from exact ID matches. This makes them suitable for the routing objective in Eq.~\ref{eq:soft_target}, which asks the router to choose experts that keep ID samples above their FPR@95 boundary and push context samples below it.

\definecolor{blue1}{HTML}{3498DB}
\definecolor{green1}{HTML}{1B7A47}
\begin{figure}[t]
\centering
\begin{tikzpicture}
\begin{axis}[
    width=0.84\linewidth, height=3.2cm,
    scale only axis=true,
    xmode=log, log basis x=10,
    xlabel={WebOE images used},
    ylabel={FPR@95 $\downarrow$},
    ylabel style={yshift=-0.5ex},
    axis line style={thick, black!80},
    tick style={thick, black!80}, tick align=outside,
    label style={font=\small, color=black!90},
    tick label style={font=\footnotesize, color=black!80},
    ymajorgrids=true, grid style={dashed, gray!30},
    axis x line*=bottom, axis y line*=left,
    enlarge x limits=0.06,
    ymin=18.2, ymax=25.5,
    xtick={250,500,1000,2000,5078},
    xticklabels={250,500,1000,2000,5078},
    xticklabel style={font=\scriptsize, rotate=45, anchor=east, color=black!80},
    legend style={draw=none, fill=none, font=\scriptsize, at={(0.98,0.98)}, anchor=north east, row sep=-1pt},
    every axis plot/.append style={thick},
]
\addplot[name path=up, draw=none, forget plot] coordinates {(250,24.78) (500,24.58) (1000,23.33) (2000,19.98) (5078,18.86)};
\addplot[name path=lo, draw=none, forget plot] coordinates {(250,24.11) (500,23.47) (1000,21.41) (2000,18.45) (5078,18.86)};
\addplot[blue1!17, forget plot] fill between[of=up and lo];        
\addplot[blue1, very thick, mark=*, mark size=1.7pt, mark options={fill=white, draw=blue1}]
    coordinates {(250,24.44) (500,24.05) (1000,22.43) (2000,19.4) (5078,18.86)};
\addlegendentry{mean ($N{=}5$)}
\addlegendimage{area legend, fill=blue1!17, draw=blue1!17}
\addlegendentry{subsample range}
\addplot[green1, only marks, mark=star, mark size=3.0pt, mark options={fill=green1, draw=green1}]
    coordinates {(5078,18.86)};
\addlegendentry{deployed (Table~\ref{tab:main_results})}
\end{axis}
\end{tikzpicture}
\vspace{-0.9em}
\caption{\textbf{WebOE supervision-size sweep (ViT-B/16).}
ImageNet-1K FPR@95 as the deployed router ($N{=}5$, $s{=}9$, top-$32$) is trained on increasing
subsamples of the WebOE pool (mean over four nested subsamples; shading is min--max).}
\label{fig:weboe_proportion}
\end{figure}
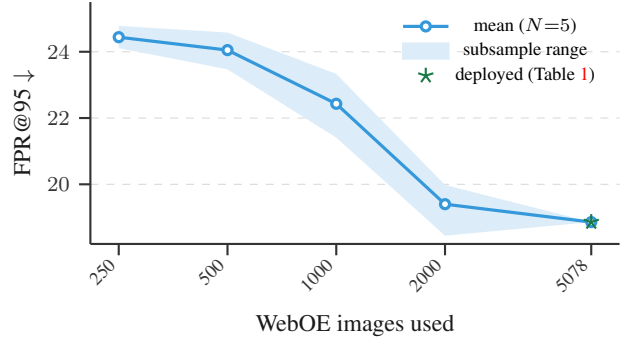

Figure~\ref{fig:weboe_proportion} sweeps the deployed router ($N{=}5$, $s{=}9$, top-$32$) over increasing subsamples of the WebOE pool. FPR@95 decreases as more context is added and reaches the deployed detector at the full $5{,}078$-image pool, reproducing the Table~\ref{tab:main_results} headline $18.86$. The trend is robust to the particular subset sampled: the min--max band over four nested subsamples stays within ${\sim}2$ FPR@95 points.

\noindent\textbf{Proxy-source check.}
Holding the router, context size, and training recipe fixed, WebOE improves over the best single external proxy and over ID-only proxy contexts. Notably, local crops from the few-shot ID images---a Local-Prompt-style source of negative regions---also improve \name{} over static fusion, showing that ID-derived local context can supervise useful routing even without external images. Full per-source results are in Appendix Table~\ref{tab:proxy_ablation}.
\begin{table}[t]
\centering
\caption*{\textbf{Proxy-source check.} Same router, context size, and training recipe; only the proxy source changes. Full per-dataset results are in Table~\ref{tab:proxy_ablation}.}
\vspace{-0.25em}
\begingroup
\footnotesize
\setlength{\tabcolsep}{3.4pt}
\renewcommand{\arraystretch}{1.04}
\setlength{\aboverulesep}{0.25ex}
\setlength{\belowrulesep}{0.25ex}
\begin{tabular}{@{}>{\raggedright\arraybackslash}p{0.54\linewidth}>{\centering\arraybackslash}p{0.20\linewidth}>{\centering\arraybackslash}p{0.18\linewidth}@{}}
\toprule
\rowcolor{headerblue}
\textbf{Proxy source} & \textbf{Avg. FPR}$\downarrow$ & \textbf{Gap} \\
\midrule
\rowcolor{tableblue}
\textbf{WebOE} & \cellcolor{green!34}\textbf{18.86} & \textbf{--} \\
Best external dataset & \cellcolor{green!18}23.74 & +4.88 \\
ID-only local crops (Local-Prompt-style) & \cellcolor{green!13}26.71 & +7.85 \\
Generic ID-only synthetic & \cellcolor{yellow!14}27.38 & +8.52 \\
MoD static fusion & \cellcolor{yellow!18}30.70 & +11.84 \\
\bottomrule
\end{tabular}
\endgroup
\vspace{-0.85em}
\end{table}

\section{Experimental Settings}
\label{subsec:setting}

\textit{Benchmarks.}
We evaluate on the standard ImageNet-1K CLIP OOD suite in the same ID-positive metric convention used by MCM~\cite{ming2022delvingoutofdistributiondetectionvisionlanguage} and OpenOOD.
ImageNet-1K~\cite{deng2009imagenet} is ID, and the OOD sets are iNaturalist~\cite{Horn2023iNaturalist}, SUN~\cite{xiao2010sun}, Places~\cite{zhou2017places}, and Textures~\cite{cimpoi2013describingtextureswild}, following the protocol of~\cite{huang2021mosscalingoutofdistributiondetection}.

\textit{Scoring.}
We keep CLIP ViT-B/16~\cite{radford2021learning} frozen (image and text encoders) and use \textbf{GL-MCM}~\cite{miyai2025glmcmgloballocalmaximum} as the per-layer score, combining the global \texttt{[CLS]} confidence with a local patch term, $S_\ell=S^{\mathrm{global}}_\ell+\lambda\,S^{\mathrm{local}}_\ell$ with $\lambda{=}0.5$.

\textit{Few-shot ID and web-retrieved outlier context.}
The router is trained on \textbf{4-shot} ID features extracted with the frozen backbone. For weak outlier supervision we use the selected \textbf{WebOE} proxy (Sec.~\ref{sec:weboe}), a retrieval-based weak OOD supervision source built from WordNet-filtered Wikimedia Commons images for the chosen ID label set. WebOE is built without target-OOD labels or test instances, and each training example carries only a binary domain label ($y{=}0$ ID, $y{=}1$ context); it is used solely to learn expert/layer routing, never for classification or as a test-time score.

\textit{Router and ensemble.}
The default WebOE router uses mean-pooled layer features, a small MLP router, a mandatory final-layer anchor, AdamW with learning rate $10^{-2}$ and weight decay $10^{-4}$, batch size 64, and the top-$k$ soft target of Eq.~\ref{eq:soft_target} ($k{=}32$; see App.~\ref{appendix:target}). The deployed ImageNet-1K detector uses one fixed global recipe: WebOE, hidden widths $256,128,32$, 6 epochs, sparsity budget $s{=}9$, and routing target top-$32$; Fig.~\ref{fig:ablation_study} sweeps these design choices. Our default detector, \name{}, averages the per-sample scores of \textbf{5} independently trained routers with seeds fixed in advance ($0$--$4$), not selected by benchmark performance (shared frozen features), yielding \emph{deterministic test-time inference} (a fixed scalar score per input) at negligible added cost. Appendix Table~\ref{tab:configs_used} maps each reported result to its WebOE pool, router count, and shot setting.

\textit{Metrics.}
We report AUROC and FPR@95, where FPR@95 is the false-positive rate on OOD when the ID true-positive rate is fixed at $95\%$ (lower is better).

\section{Main Results}
\label{main-results}

\begin{table}[!t]
\centering
\caption{
\textbf{ImageNet-1K OOD detection with frozen CLIP ViT-B/16.}
Each cell reports FPR@95$\downarrow$/AUROC$\uparrow$; Avg. is the macro-average across OOD datasets.
}
\vspace{-0.2em}
\label{tab:main_results}

\begingroup
\footnotesize
\setlength{\tabcolsep}{2.0pt}
\renewcommand{\arraystretch}{1.02}
\setlength{\aboverulesep}{0.25ex}
\setlength{\belowrulesep}{0.25ex}
\newcommand{\fprauc}[2]{#1{\scriptsize/}#2}
\resizebox{\linewidth}{!}{%
\begin{tabular}{@{}lccccc@{}}
\toprule
\rowcolor{headerblue}
\textbf{Method} &
\textbf{iNat.} & \textbf{SUN} & \textbf{Places} & \textbf{DTD} & \textbf{Avg.} \\
\midrule
\multicolumn{6}{@{}l}{\textit{Zero-shot, training free}} \\
MCM \cite{ming2022delvingoutofdistributiondetectionvisionlanguage} &
\fprauc{30.91}{94.61} & \fprauc{37.59}{92.57} & \fprauc{44.69}{89.77} & \fprauc{57.77}{86.11} & \cellcolor{orange!20}\fprauc{42.74}{90.77} \\
GL-MCM \cite{miyai2025glmcmgloballocalmaximum} &
\fprauc{17.42}{96.44} & \fprauc{30.75}{93.44} & \fprauc{37.62}{90.63} & \fprauc{55.20}{85.54} & \cellcolor{orange!14}\fprauc{35.25}{91.51} \\
SeTAR+MCM \cite{li2024setaroutofdistributiondetectionselective} &
\fprauc{26.92}{94.67} & \fprauc{35.57}{92.79} & \fprauc{42.64}{90.16} & \fprauc{55.83}{86.58} & \cellcolor{orange!18}\fprauc{40.24}{91.05} \\
SeTAR+GL \cite{li2024setaroutofdistributiondetectionselective} &
\fprauc{13.36}{96.92} & \fprauc{28.17}{93.36} & \fprauc{36.80}{90.40} & \fprauc{54.17}{84.59} & \cellcolor{yellow!18}\fprauc{33.12}{91.32} \\
MoD \cite{delajara2025mysteriesdeeproleintermediate} &
\fprauc{15.98}{96.90} & \fprauc{45.58}{89.69} & \fprauc{35.71}{92.72} & \fprauc{\textbf{25.51}}{\textbf{94.84}} & \cellcolor{yellow!18}\fprauc{30.70}{93.54} \\
\midrule
\multicolumn{6}{@{}l}{\textit{One-shot}} \\
CoOp \cite{Zhou_2022} &
\fprauc{18.95}{95.52} & \fprauc{29.58}{92.90} & \fprauc{38.72}{89.64} & \fprauc{48.03}{85.87} & \cellcolor{yellow!18}\fprauc{33.82}{90.98} \\
LoCoOp \cite{miyai2023locoopfewshotoutofdistributiondetection} &
\fprauc{21.67}{95.69} & \fprauc{22.98}{95.07} & \fprauc{31.41}{92.10} & \fprauc{49.79}{87.85} & \cellcolor{yellow!16}\fprauc{31.46}{92.68} \\
ID-Like \cite{bai2024idlikepromptlearningfewshot} &
\fprauc{14.57}{97.35} & \fprauc{44.02}{91.08} & \fprauc{41.74}{91.15} & \fprauc{26.77}{94.38} & \cellcolor{yellow!16}\fprauc{31.78}{93.49} \\
\rowcolor{tableblue}
\textbf{\name{}} &
\fprauc{4.38}{98.48} & \fprauc{25.64}{94.21} & \fprauc{24.64}{92.36} & \fprauc{32.85}{88.74} & \cellcolor{green!24}\fprauc{21.88}{93.45} \\
\midrule
\multicolumn{6}{@{}l}{\textit{Four-shot}} \\
CoOp \cite{Zhou_2022} &
\fprauc{14.60}{96.62} & \fprauc{28.48}{94.98} & \fprauc{36.49}{89.98} & \fprauc{43.13}{88.03} & \cellcolor{yellow!18}\fprauc{30.67}{91.82} \\
LoCoOp \cite{miyai2023locoopfewshotoutofdistributiondetection} &
\fprauc{16.05}{96.86} & \fprauc{23.44}{\textbf{95.94}} & \fprauc{32.87}{91.98} & \fprauc{42.28}{90.19} & \cellcolor{yellow!12}\fprauc{28.66}{93.74} \\
ID-Like \cite{bai2024idlikepromptlearningfewshot} &
\fprauc{8.98}{98.19} & \fprauc{42.03}{91.64} & \fprauc{44.00}{90.57} & \fprauc{25.27}{94.32} & \cellcolor{yellow!16}\fprauc{30.07}{93.68} \\
Local-Prompt \cite{zeng2025localpromptextensiblelocalprompts} &
\fprauc{9.65}{97.87} & \fprauc{\textbf{20.40}}{95.57} & \fprauc{29.39}{92.67} & \fprauc{51.20}{88.08} & \cellcolor{green!12}\fprauc{27.66}{93.53} \\
\rowcolor{tableblue}
\textbf{\name{}} &
\fprauc{\textbf{3.93}}{\textbf{98.78}} & \fprauc{21.65}{95.13} & \fprauc{\textbf{21.88}}{\textbf{93.48}} & \fprauc{27.96}{90.58} & \cellcolor{green!34}\fprauc{\textbf{18.86}}{\textbf{94.49}} \\
\bottomrule
\end{tabular}%
}
\endgroup
\vspace{-1.0em}
\end{table}

\noindent\textbf{ImageNet-1K.}
Table~\ref{tab:main_results} compares zero-shot, prompt-learning, and few-shot detectors on the standard ImageNet-1K OOD suite. With only 4-shot ID and WebOE, the deployed $N{=}5$ \name{} ensemble reaches the best macro-average result: \textbf{18.86} FPR@95 and \textbf{94.49} AUROC (Table~\ref{tab:configs_used}). This improves over the strongest prompt-learning baseline, Local-Prompt (27.66 FPR@95), by $8.8$ points, and over the static intermediate-layer fusion baseline MoD (30.70) by $11.8$ points. Since MoD already uses intermediate layers but fixes one fusion rule, the gap isolates the value of input-conditioned layer selection beyond access to intermediate representations. The residual gaps are also explicit: Local-Prompt remains better on SUN, and ID-Like remains better on Texture. Thus the claim is macro-average and complementary improvement, not per-dataset dominance over every detector.

\noindent\textbf{Pascal-VOC.}
Table~\ref{tab:voc_results} tests transfer to a multi-label ID taxonomy using the MoD/SeTAR VOC protocol~\cite{delajara2025mysteriesdeeproleintermediate}. SeTAR and MoD fit their configurations on the VOC ID validation split; \name{} uses the same ID-data budget (94 images) and evaluates on the disjoint VOC test split. Under this matched protocol, \name{} lowers the best training-free average FPR@95 from $26.74$ to \textbf{18.41} and wins five of six OOD sets. COCO remains the hardest column because the OOD set shares object categories with the VOC ID taxonomy. This transfer result is important because WebOE is not tied to ImageNet: we reuse the construction procedure unchanged and build a separate proxy pool from the Pascal-VOC ID labels.

\begin{table}[!ht]
\centering
\caption{
\textbf{Pascal-VOC OOD detection (ViT-B/16).}
Each cell reports FPR@95$\downarrow$/AUROC$\uparrow$; Avg.\ is over six OOD sets. Best values are bold.
}
\vspace{-0.2em}
\label{tab:voc_results}

\begingroup
\footnotesize
\setlength{\tabcolsep}{1.7pt}
\renewcommand{\arraystretch}{1.02}
\setlength{\aboverulesep}{0.25ex}
\setlength{\belowrulesep}{0.25ex}
\newcommand{\fprauc}[2]{#1{\scriptsize/}#2}
\resizebox{\linewidth}{!}{%
\begin{tabular}{@{}lccccccc@{}}
\toprule
\rowcolor{headerblue}
\textbf{Method} & \textbf{iNat.} & \textbf{SUN} & \textbf{Places} & \textbf{DTD} & \textbf{IN-22K} & \textbf{COCO} & \textbf{Avg.} \\
\midrule
\multicolumn{8}{@{}l}{\textit{Training-free (MCM/GL-MCM zero-shot; SeTAR/MoD val-fitted)}} \\
MCM &
\fprauc{10.51}{97.93} & \fprauc{30.45}{94.25} & \fprauc{36.11}{91.86} & \fprauc{53.21}{91.77} & \fprauc{53.82}{91.12} & \fprauc{57.10}{89.02} & \cellcolor{orange!20}\fprauc{40.20}{92.66} \\
SeTAR+MCM &
\fprauc{4.38}{98.70} & \fprauc{26.24}{94.95} & \fprauc{28.67}{93.28} & \fprauc{50.32}{92.32} & \fprauc{44.61}{92.63} & \fprauc{49.80}{89.68} & \cellcolor{yellow!18}\fprauc{34.00}{93.59} \\
GL-MCM &
\fprauc{4.33}{98.81} & \fprauc{22.94}{94.63} & \fprauc{26.20}{93.11} & \fprauc{41.61}{92.88} & \fprauc{37.88}{93.17} & \fprauc{43.70}{90.71} & \cellcolor{yellow!16}\fprauc{29.44}{93.88} \\
SeTAR+GL &
\fprauc{3.01}{99.04} & \fprauc{21.76}{94.98} & \fprauc{24.00}{93.73} & \fprauc{37.61}{93.87} & \fprauc{33.46}{94.24} & \fprauc{\textbf{40.60}}{\textbf{91.48}} & \cellcolor{green!12}\fprauc{26.74}{94.56} \\
MoD~\cite{delajara2025mysteriesdeeproleintermediate} &
\fprauc{2.19}{98.92} & \fprauc{19.70}{95.77} & \fprauc{19.53}{95.11} & \fprauc{44.08}{92.06} & \fprauc{34.70}{91.15} & \fprauc{41.60}{88.09} & \cellcolor{green!10}\fprauc{26.97}{93.52} \\
\midrule
\multicolumn{8}{@{}l}{\textit{Few-shot, WebOE router}} \\
\name{} 1-shot &
\fprauc{0.89}{99.44} & \fprauc{15.83}{96.26} & \fprauc{20.84}{95.51} & \fprauc{32.80}{93.56} & \fprauc{19.84}{96.17} & \fprauc{44.40}{88.80} & \cellcolor{green!24}\fprauc{22.43}{94.96} \\
\rowcolor{tableblue}
\textbf{\name{} val} &
\fprauc{\textbf{0.57}}{\textbf{99.68}} & \fprauc{\textbf{13.22}}{\textbf{96.94}} & \fprauc{\textbf{15.12}}{\textbf{96.59}} & \fprauc{\textbf{23.79}}{\textbf{94.86}} & \fprauc{\textbf{13.88}}{\textbf{97.13}} & \fprauc{43.90}{88.75} & \cellcolor{green!34}\fprauc{\textbf{18.41}}{\textbf{95.66}} \\
\bottomrule
\end{tabular}%
}
\endgroup
\vspace{-1.0em}
\end{table}

\begin{table}[t]
\centering
\caption{
\textbf{Additional ViT-B/16 ImageNet-100 benchmark.}
Each cell reports FPR@95$\downarrow$/AUROC$\uparrow$; Avg. is the macro-average over the four OOD shifts. Published baselines are from Local-Prompt~\cite{zeng2025localpromptextensiblelocalprompts}; GL-MCM is our recomputation in the matched protocol.
}
\vspace{-0.2em}
\label{tab:subset_results_main}

\begingroup
\footnotesize
\setlength{\tabcolsep}{2.0pt}
\renewcommand{\arraystretch}{1.02}
\setlength{\aboverulesep}{0.25ex}
\setlength{\belowrulesep}{0.25ex}
\newcommand{\fprauc}[2]{#1{\scriptsize/}#2}
\resizebox{\linewidth}{!}{%
\begin{tabular}{@{}lccccc@{}}
\toprule
\rowcolor{headerblue}
\textbf{Method} &
\textbf{iNat.} & \textbf{SUN} & \textbf{Places} & \textbf{DTD} & \textbf{Avg.} \\
\midrule
MCM &
\fprauc{18.13}{96.77} & \fprauc{36.45}{94.54} & \fprauc{34.52}{94.36} & \fprauc{41.22}{92.25} & \cellcolor{yellow!18}\fprauc{32.58}{94.48} \\
GL-MCM (ours) &
\fprauc{17.26}{96.90} & \fprauc{34.52}{94.77} & \fprauc{33.69}{94.54} & \fprauc{38.33}{93.38} & \cellcolor{yellow!18}\fprauc{30.95}{94.90} \\
LoCoOp &
\fprauc{44.94}{93.31} & \fprauc{30.70}{94.70} & \fprauc{34.31}{93.93} & \fprauc{55.92}{90.01} & \cellcolor{orange!20}\fprauc{41.47}{92.99} \\
Local-Prompt &
\fprauc{6.99}{98.05} & \fprauc{24.34}{96.03} & \fprauc{26.04}{95.49} & \fprauc{36.32}{93.67} & \cellcolor{green!12}\fprauc{23.42}{95.81} \\
\rowcolor{tableblue}
\textbf{\name{}} &
\fprauc{\textbf{1.94}}{\textbf{98.98}} & \fprauc{\textbf{19.75}}{\textbf{96.14}} & \fprauc{\textbf{15.90}}{\textbf{96.53}} & \fprauc{\textbf{12.23}}{\textbf{97.30}} & \cellcolor{green!34}\fprauc{\textbf{12.46}}{\textbf{97.24}} \\
\bottomrule
\end{tabular}%
}
\endgroup
\vspace{-1.0em}
\end{table}

\noindent\textbf{ImageNet-100.}
Table~\ref{tab:subset_results_main} changes the ID taxonomy with a ViT-B/16 ImageNet-100 check. Against the published Local-Prompt baselines~\cite{zeng2025localpromptextensiblelocalprompts}, \name{} lowers average FPR@95 from 23.42 to \textbf{12.46}. The gain is largest on Texture, where the WebOE router favors lower-level experts more often than the ImageNet-1K default. Together with Pascal-VOC, this stress test supports the claim that WebOE is a retrieval-based construction for a chosen ID distribution rather than an ImageNet-specific proxy.

\begin{figure}[t]
\centering
\includegraphics[width=\linewidth]{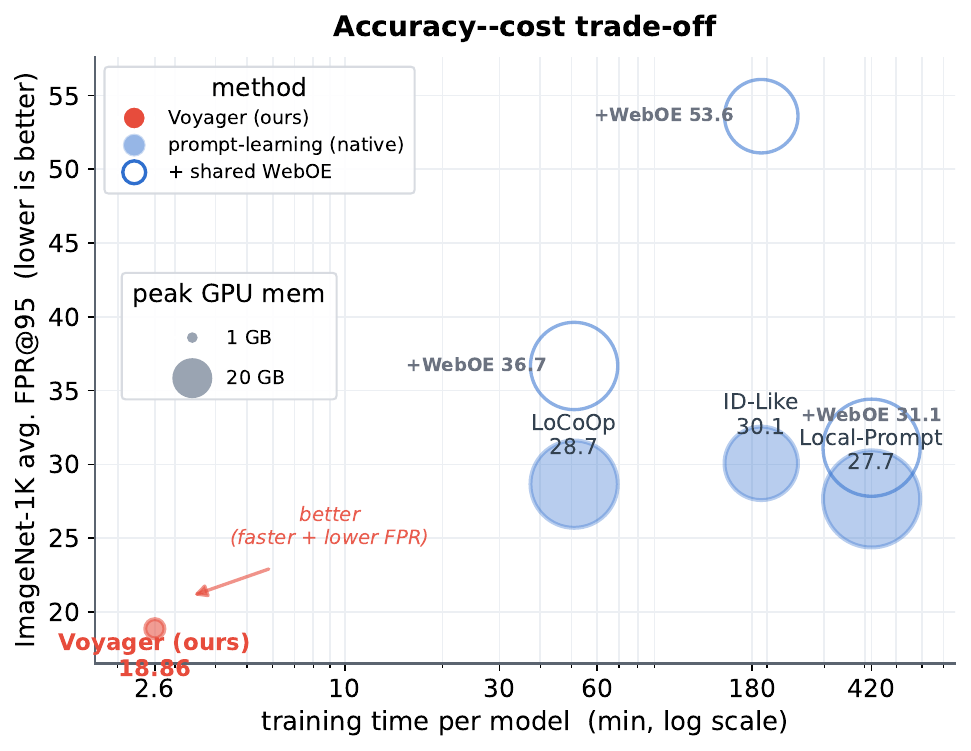}
\vspace{-1.6em}
\caption{\textbf{Efficiency--accuracy trade-off (ImageNet-1K, ViT-B/16).} Solid markers use Table~\ref{tab:main_results} FPR@95 values (Voyager: 18.86); hollow markers show prompt baselines retrained with shared WebOE. Bubble area denotes peak GPU memory.}
\label{fig:pareto}
\end{figure}

\noindent\textbf{Cost and determinism.}
Because frozen CLIP features are cached once, Figure~\ref{fig:pareto} plots the deployed Table~\ref{tab:main_results} accuracy values against measured training cost: \name{} reaches 18.86 FPR@95 while training in about 2 minutes and under 1 GB, versus 50.9 minutes to 7.1 hours and 13.1--22.9 GB for prompt-learning baselines. Hollow controls retrain the prompt baselines with the same WebOE context; adding the proxy alone does not close the gap, while routing on frozen experts moves the method to the lower-left frontier. The reported detector fixes seeds 0–4 in advance (not selected by benchmark performance), yielding 18.86 deterministically; re-drawing the ensemble with fresh seeds gives 19.91 ± 0.79 FPR@95, so even an adversarially re-seeded ensemble stays within ~1 point.

\begin{table}[t]
\centering
\caption{\textbf{Selector ablation.} Same expert family, frozen scores, and WebOE context; only the expert-selection mechanism changes. Avg.\ FPR@95$\downarrow$ on ImageNet-1K.}
\label{tab:selector_ablation_main}
\vspace{-0.2em}
\begingroup
\footnotesize
\setlength{\tabcolsep}{3.4pt}
\renewcommand{\arraystretch}{1.08}
\setlength{\aboverulesep}{0.25ex}
\setlength{\belowrulesep}{0.25ex}
\begin{tabular}{@{}>{\raggedright\arraybackslash}p{0.36\linewidth}>{\raggedright\arraybackslash}p{0.39\linewidth}>{\centering\arraybackslash}p{0.18\linewidth}@{}}
\toprule
\rowcolor{headerblue}
\textbf{Selector} & \textbf{Selection signal} & \textbf{FPR@95}$\downarrow$ \\
\midrule
Last-layer & Fixed final-layer expert & \cellcolor{orange!20}45.64 \\
Fixed context & One global expert selected on WebOE & \cellcolor{yellow!22}34.60 \\
Best score rule & Per-input max expert score & \cellcolor{yellow!14}28.93 \\
\rowcolor{tableblue}
\textbf{Learned \name{} router} & \textbf{Image feature $\rightarrow$ expert} & \cellcolor{green!34}\textbf{18.86} \\
Label-aware oracle & Analysis-only upper bound & \cellcolor{gray!15}7.06 \\
\bottomrule
\end{tabular}
\endgroup
\vspace{-0.8em}
\end{table}

\noindent\textbf{Selector.}
Table~\ref{tab:selector_ablation_main} shows that, with identical experts, scores, and WebOE context, the learned selector beats the strongest score rule by $10.18$ FPR@95 and the fixed context-selected expert by $15.85$. This is the main mechanism check: the WebOE signal is useful because it trains an image-conditioned routing policy, not because it simply identifies one globally good layer.

\noindent\textbf{Routing diagnostics.}

\begin{figure}[t]
    \centering
    \includegraphics[width=0.82\linewidth]{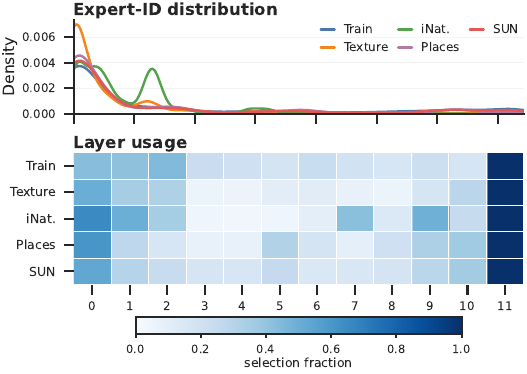}
    \vspace{-1.2em}
    \caption{\textbf{Default \name{} routing.} Top: selected expert distributions. Bottom: layer usage. ViT-B/16, WebOE, 4-shot ID, $N{=}5$.}
    \label{fig:expert-distribution}
    \vspace{-0.8em}
\end{figure}


\definecolor{blue1}{HTML}{3498DB}

\begin{figure}[t]
\centering
\begin{tikzpicture}
\pgfplotsset{
    modern_axis/.style={
        width=0.175\linewidth,   
        height=2.0cm,            
        scale only axis=true,    
        axis line style={thick, black!80},
        tick style={thick, black!80},
        tick align=outside,
        label style={font=\scriptsize, color=black!90},
        tick label style={font=\tiny, color=black!80},
        every axis plot/.append style={thick, mark size=1.0pt},
        enlarge x limits=0.08,
        ymajorgrids=true,
        grid style={dashed, gray!30},
        axis x line*=bottom,
        axis y line*=left,
    }
}

\begin{axis}[
    modern_axis,
    name=plot1,
    xlabel={Mandatory Layer},
    ylabel={FPR@95 $\downarrow$},
    ylabel style={yshift=-1.5ex}, 
    xmin=1, xmax=12,
    ymin=0.15, ymax=0.9,
    xtick={1,4,8,12},
    ytick={0.2, 0.4, 0.6, 0.8},
    yticklabel style={/pgf/number format/precision=1, /pgf/number format/fixed},
]
    \addplot[color=blue1, mark=*] coordinates {
        (1, 0.8509) (2, 0.2946) (3, 0.2351) (4, 0.2783) (5, 0.2737)
        (6, 0.2946) (7, 0.2662) (8, 0.2539) (9, 0.2643) (10, 0.2589)
        (11, 0.2840) (12, 0.1886)
    };
\end{axis}

\begin{axis}[
    modern_axis,
    name=plot2,
    at={(plot1.south east)},
    anchor=south west,
    xshift=0.075\linewidth, 
    xlabel={Sparsity Level},
    xmin=2, xmax=9,
    ymin=0.18, ymax=0.32,
    xtick={2,5,9},
    ytick={0.18, 0.22, 0.26, 0.30},
    yticklabel style={/pgf/number format/precision=2, /pgf/number format/fixed},
]
    \addplot[color=blue1, mark=*] coordinates {
        (2, 0.3128) (3, 0.2662) (4, 0.2378) (5, 0.2201) (6, 0.2133)
        (7, 0.2031) (8, 0.2036) (9, 0.1886)
    };
\end{axis}

\begin{axis}[
    modern_axis,
    name=plot3,
    at={(plot2.south east)},
    anchor=south west,
    xshift=0.075\linewidth,
    xlabel={Router Depth},
    xmin=1, xmax=5,
    ymin=0.18, ymax=0.22,
    xtick={1,2,3,4,5},
    ytick={0.18, 0.20, 0.22},
    yticklabel style={/pgf/number format/precision=2, /pgf/number format/fixed},
]
    \addplot[color=blue1, mark=*] coordinates {
        (1, 0.2056) (2, 0.1973) (3, 0.1886) (4, 0.2114) (5, 0.1995)
    };
\end{axis}

\begin{axis}[
    modern_axis,
    name=plot4,
    at={(plot3.south east)},
    anchor=south west,
    xshift=0.075\linewidth,
    ybar,
    bar width=7pt, 
    xlabel={Fusion},
    ymin=0.18, ymax=0.22,
    enlarge x limits=0.2,
    symbolic x coords={Last, Max, Concat, Mean},
    xtick=data,
    xticklabel style={rotate=45, anchor=north east, font=\tiny, inner sep=1pt},
    ytick={0.18, 0.20, 0.22},
    yticklabel style={/pgf/number format/precision=2, /pgf/number format/fixed},
]
    \addplot[fill=blue1!50, draw=blue1, thick] coordinates {
        (Last, 0.2107)
        (Max, 0.2021)
        (Concat, 0.2000)
        (Mean, 0.1886)
    };
\end{axis}
\end{tikzpicture}
\vspace{-2.5em}
\caption{\textbf{Ablation study (FPR@95 $\downarrow$).} Each panel varies one design axis with the rest locked to the deployed config; the deployed setting is best on every panel.}
\label{fig:ablation_study}
\end{figure}
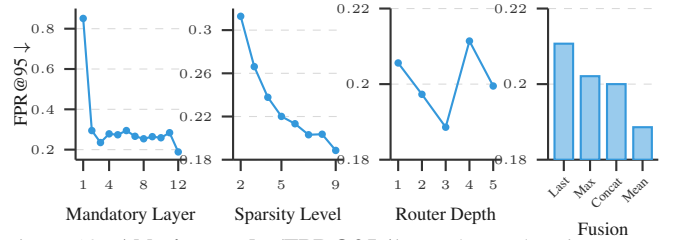

Figure~\ref{fig:expert-distribution} shows that \name{} learns distinct routing policies across OOD datasets rather than collapsing to a single expert. While the final CLIP layer is consistently selected as a semantic anchor, the use of intermediate layers varies across shifts, indicating that different OOD distributions benefit from different representations. This behavior directly supports the motivation for input-conditioned layer routing over a fixed fusion rule. Figure~\ref{fig:ablation_study} further validates the deployed design: keeping the final-layer anchor, using $s{=}9$, routing from mean-pooled multilayer features, and a three-layer MLP consistently minimizes FPR@95.

\section*{Conclusion}

\name{} introduces input-conditioned layer routing for frozen CLIP OOD detection. Rather than adapting prompts or using a fixed intermediate-layer fusion rule, \name{} learns to select a sparse, final-layer-anchored expert for each input using WebOE, an automatically retrieved source of weak OOD supervision. Across three benchmarks, \name{} consistently improves OOD detection while training approximately $20\times$ more efficiently than prompt-learning approaches. Extensive ablations show that the gains arise from dynamic layer routing rather than the supervision source or static fusion alone.
\FloatBarrier

{\small
\bibliographystyle{ieee_fullname}
\bibliography{main}
}

\appendix
\appendix
\clearpage 
\counterwithin{figure}{section}
\counterwithin{algorithm}{section}

\section{Appendix Overview and Organization}
\label{appendix:overview}

This appendix provides additional material that complements the main paper and further supports the empirical and methodological foundations of \name{}. In particular, it expands the discussion of datasets, architecture details, limitations, expert complementarity, random layer selection, cross-source OOD generalization, retrieval-based OOD supervision, theoretical motivation, layer-importance diagnostics, and interpretability.

For clarity, the appendix is organized as follows:
\begin{itemize}
    \item \textbf{Sec.~\ref{appendix:datasets}} details the in-distribution, evaluation, and proxy OOD datasets used throughout the paper.
    \item \textbf{Sec.~\ref{appendix:WebOE}} describes \textit{WebOE}, a retrieval-based strategy for constructing proxy OOD context from Wikimedia Commons.
    \item \textbf{Sec.~\ref{appendix:architecture}} records the default router architecture and score-flow used in the main ImageNet-1K experiments.
    \item \textbf{Sec.~\ref{sec:overhead}} reports inference runtime and latency overhead.
    \item \textbf{Sec.~\ref{appendix:selector}} ablates the selection mechanism, showing the learned router beats every score-based selection rule and the static baselines.
    \item \textbf{Sec.~\ref{appendix:target}} ablates the routing-target sharpness (hard top-1, top-$k$, and fully soft) used by the deployed detector.
    \item \textbf{Sec.~\ref{appendix:fewshot}} reports how \name{} scales with the number of ID shots, showing that 1--2 shots are sufficient.
    \item \textbf{Sec.~\ref{appendix:ensemble}} reports how score-level router averaging reduces variance, justifying the deployed $N{=}5$ ensemble.
    \item \textbf{Sec.~\ref{appendix:baseline-training}} details how the prompt-learning baselines are re-trained under the same WebOE context as \name{}, and the harness used to measure training time and memory.
    \item \textbf{Sec.~\ref{appendix:backbones}} extends \name{} to additional backbones and detection frameworks.
    \item \textbf{Sec.~\ref{appendix:extra-results-complementarity}} analyzes complementarity between experts across sparsity levels and backbones.
    \item \textbf{Sec.~\ref{app:random_selection}} studies random layer selection as a zero-shot baseline and examines why it can already provide gains.
    \item \textbf{Sec.~\ref{sec:theory}} provides complementary theoretical justification for the value of intermediate layers and dynamic routing.
    \item \textbf{Sec.~\ref{appendix:layer_importance}} provides a post-hoc layer-importance diagnostic for the fixed Table~\ref{tab:main_results} detector.
    \item \textbf{Sec.~\ref{app:sliding_window_heatmaps}} presents additional interpretability analyses through sliding-window OOD heatmaps.
    \item \textbf{Sec.~\ref{appendix:limitation}} discusses the main limitations of \name{}, with emphasis on the dependence on proxy OOD supervision.
\end{itemize}

Taken together, these sections broaden and reinforce the main paper along two complementary dimensions. Empirically, they show that the benefits of \name{} are not tied to a particular proxy OOD source, training recipe, or backbone, but instead persist across varied supervision settings and evaluation conditions. Conceptually, they provide additional intuition and theoretical support for our central claim: that intermediate-layer diversity, when exploited through dynamic routing, is a key ingredient for robust OOD detection.

\section{Dataset description}
\label{appendix:datasets}


\subsection*{In-Distribution Dataset}

\paragraph{ImageNet-1K.}
ImageNet-1K~\cite{deng2009imagenet} is the primary in-distribution (ID) dataset for the standard CLIP OOD suite. Following the few-shot setting, we sample a small number of images per class (1-shot and 4-shot) from the ImageNet training set to train the \name{} router. These samples are used only for router supervision; the CLIP backbone remains frozen.

For evaluation, we follow the standard ImageNet-based OOD protocol~\cite{huang2021mosscalingoutofdistributiondetection, ming2022delvingoutofdistributiondetectionvisionlanguage} and use the full ImageNet-1K validation split, consisting of 50,000 images across 1,000 categories. All OOD detection results reported in the main paper are computed using this zero-shot evaluation setting.

\paragraph{Pascal-VOC.}
We also evaluate on the multi-label Pascal-VOC ID benchmark used by SeTAR and MoD. Following that protocol, the ID fitting pool is the SeTAR validation split (94 images over 14 classes), while ID evaluation is performed on the disjoint VOC test split (906 images). The \name{}~(val) row in Table~\ref{tab:voc_results} uses the full 94-image validation pool; the 1-shot row uses one image per class.

\paragraph{ID split separation.}
For every ID benchmark, the ID images used for router training are distinct from the ID images used for evaluation. On ImageNet-1K, the router receives 4-shot samples from the ImageNet training split, while evaluation uses the ImageNet validation split. On Pascal-VOC, the \name{}~(val) row uses the SeTAR validation split as the ID fitting pool and evaluates on the disjoint VOC test split. This matches the validation-fitted protocol used by SeTAR and MoD on Pascal-VOC: those methods also select or fit their configuration on the ID validation pool rather than on VOC test images. Table~\ref{tab:data_split_accounting} summarizes the exact split roles and sample counts.

\begin{table*}[t]
\centering
\scriptsize
\setlength{\tabcolsep}{3.0pt}
\renewcommand{\arraystretch}{1.03}
\caption{\textbf{Data split accounting.} ID train/fitting images are disjoint from ID evaluation images. WebOE is built from each benchmark's ID labels; no OOD evaluation image is used for router training or WebOE construction.}
\label{tab:data_split_accounting}
\resizebox{\textwidth}{!}{%
\begin{tabular}{@{}llclp{6.6cm}@{}}
\toprule
\textbf{Benchmark} & \textbf{Role} & \textbf{Split/source} & \textbf{\# images} & \textbf{Use in this paper} \\
\midrule
ImageNet-1K & ID router train & ImageNet train & $4{,}000$ & 4-shot ID supervision and ID-boundary estimation for the router; sampled from the training split only. \\
ImageNet-1K & Weak OOD context & WebOE & $5{,}078$ & Retrievals from ImageNet-1K ID names; router context only. \\
ImageNet-1K & ID evaluation & ImageNet val & $50{,}000$ & ID side of final FPR@95/AUROC evaluation; never used for router fitting. \\
ImageNet-1K & OOD evaluation & iNat./SUN/Places/DTD & $10{,}000/10{,}000/10{,}000/5{,}640$ & Held-out OOD benchmarks used only for reporting final metrics. \\
\midrule
ImageNet-100 & ID router train & ImageNet-100 train subset & $400$ & 4-shot ID supervision for the subset stress test in Table~\ref{tab:subset_results_main}. \\
ImageNet-100 & Weak OOD context & WebOE & $5{,}078$ & Retrievals from ImageNet-100 ID names; router context only. \\
ImageNet-100 & ID evaluation & ImageNet-100 val subset & $5{,}000$ & ID side of the ImageNet-100 FPR@95/AUROC evaluation; disjoint from the 4-shot train subset. \\
ImageNet-100 & OOD evaluation & iNat./SUN/Places/DTD & $10{,}000/10{,}000/10{,}000/5{,}640$ & Held-out OOD benchmarks used only for reporting final metrics. \\
\midrule
Pascal-VOC & ID router train & SeTAR val split & $94$ & Full-val \name{} row uses the same ID validation pool used to fit SeTAR/MoD configurations; the 1-shot row uses 14 images. \\
Pascal-VOC & Weak OOD context & WebOE & $5{,}078$ & Retrievals from Pascal-VOC ID names; router context only, not VOC test data. \\
Pascal-VOC & ID evaluation & VOC test & $906$ & ID side of the Pascal-VOC evaluation; disjoint from the SeTAR validation pool. \\
Pascal-VOC & OOD evaluation & SUN/Places/DTD/iNat./COCO/IN-22K & $10{,}000/10{,}000/5{,}640/10{,}000/1{,}000/10{,}000$ & Held-out OOD benchmarks from the SeTAR/MoD protocol; used only for reporting final metrics. \\
\bottomrule
\end{tabular}}
\end{table*}

\begin{table}[t]
\centering
\scriptsize
\setlength{\tabcolsep}{2.2pt}
\renewcommand{\arraystretch}{1.08}
\caption{\textbf{Configurations used in the paper.} All rows use frozen CLIP ViT-B/16, GL-MCM ($\lambda{=}0.5$), and the same router architecture unless stated otherwise. Deployed WebOE is the $5{,}078$-image Wikimedia pool; the sweep samples from a larger reservoir built with the same procedure.}
\label{tab:configs_used}
\resizebox{\linewidth}{!}{%
\begin{tabular}{@{}lcccc@{}}
\toprule
\textbf{Use} & \textbf{WebOE pool} & \textbf{Routers} & \textbf{ID shots} & \textbf{Reported value} \\
\midrule
Main result (Tab.~\ref{tab:main_results}) & WebOE ($5{,}078$) & $N{=}5$ & $4$ & $18.86$ FPR \\
WebOE sweep (Fig.~\ref{fig:weboe_proportion}) & $250$--$8{,}000$ subsamples & $N{=}8$ & $4$ & $19.19$ best diagnostic point \\
Shared-cost diagnostic (Tab.~\ref{tab:baseline_training}) & WebOE ($5{,}078$) & $N{=}8$ & $4$ & $19.12$ FPR \\
Few-shot curve (Fig.~\ref{fig:shots_curve}) & WebOE ($5{,}078$) & $N{=}8$ & $1$--$16$ & $19.77$ at $2$ shots \\
Selector ablation (Fig.~\ref{fig:selector_ablation}) & WebOE ($5{,}078$) & $N{=}8$ & $4$ & $18.75$ learned gate \\
\bottomrule
\end{tabular}}
\end{table}

\subsection*{Out-of-Distribution Datasets}

We adopt the four OOD datasets filtered by the MoS protocol~\cite{huang2021mosscalingoutofdistributiondetection}, which removes all ImageNet-overlapping categories to ensure a clean distribution shift.

\paragraph{iNaturalist.}
~\cite{Horn2023iNaturalist} contains fine-grained flora and fauna categories. We use the 10,000-image filtered version from~\cite{huang2021mosscalingoutofdistributiondetection}, which excludes any classes that map to ImageNet-1K.

\paragraph{SUN.}
~\cite{xiao2010sun} consists of indoor and outdoor scene images across thousands of semantic categories. Following~\cite{huang2021mosscalingoutofdistributiondetection}, we use the de-duplicated 10,000-image subset with all ImageNet-overlapping classes removed.

\paragraph{Places.}
~\cite{zhou2017places} provides large-scale scene categories. We use the 10,000-image subset filtered by~\cite{huang2021mosscalingoutofdistributiondetection} to ensure no semantic overlap with ImageNet-1K.

\paragraph{Texture (DTD).}
~\cite{cimpoi2014describing} includes 5,640 texture images annotated using 47 human-interpretable attributes. This dataset introduces low-level distribution shifts distinct from object-centric ImageNet categories.

\paragraph{Pascal-VOC OOD suite.}
For Pascal-VOC, we follow the SeTAR/MoD protocol and evaluate against six held-out OOD sets: the four datasets above plus COCO~\cite{lin2014microsoft} and an ImageNet-22K subset~\cite{deng2009imagenet}. These OOD images are evaluation-only; they are not used to fit \name{}, SeTAR, or MoD in the comparison.

\subsection{Weak OOD Supervision for \name{}}

\name{} uses a small amount of weak OOD supervision to train the dynamic router; this proxy context is \emph{not} used during evaluation. The supervision is provided by \textit{WebOE}, a retrieval-based proxy context constructed automatically from Wikimedia Commons using only the ID label names (no hand-curated outlier data); we detail its construction in Sec.~\ref{appendix:WebOE}.

\section{WebOE (Web External Outlier): Retrieval-Based Weak OOD Supervision}
\label{appendix:WebOE}

Constructing a proxy OOD context dataset can be challenging in practice, particularly in domain-specific settings where curated outlier data may not be readily available. To explore a more accessible alternative, we introduce \textit{WebOE}, a retrieval-based strategy that constructs weak OOD supervision from web images for a given labeled ID distribution. The goal of WebOE is to demonstrate that the router in \name{} can learn effective routing behavior from retrieval-based weak OOD supervision and is not restricted to curated, hand-collected benchmark outlier sets. Algorithm~\ref{alg:weboe} gives the full construction procedure.

\begin{algorithm}[t]
\caption{\textbf{WebOE construction.} Retrieval-based weak OOD supervision from ID labels only.}
\label{alg:weboe}
\small
\begin{algorithmic}[1]
\Require ID label set $\mathcal{C}_{\mathrm{ID}}$, candidate web labels $\mathcal{C}_{\mathrm{web}}$, WordNet graph $\mathcal{G}$, Wikimedia retriever $\mathcal{R}$, distance threshold $d_{\min}{=}2$, per-label image cap $m$
\Ensure WebOE context $\mathcal{D}_{\mathrm{ctx}}$ and cached frozen-CLIP features/scores
\State Map each ID label $c\in\mathcal{C}_{\mathrm{ID}}$ to WordNet synsets $S_{\mathrm{ID}}(c)$.
\State Initialize retained query set $\mathcal{Q}\gets\emptyset$ and context set $\mathcal{D}_{\mathrm{ctx}}\gets\emptyset$.
\ForAll{candidate label $q\in\mathcal{C}_{\mathrm{web}}$}
    \State Map $q$ to candidate synsets $S(q)$.
    \State Compute $d(q,\mathcal{C}_{\mathrm{ID}})$, the shortest WordNet path from $S(q)$ to any ID synset.
    \If{$S(q)$ overlaps an ID synset or $d(q,\mathcal{C}_{\mathrm{ID}})<d_{\min}$}
        \State discard $q$
    \Else
        \State add $q$ to $\mathcal{Q}$
    \EndIf
\EndFor
\ForAll{retained query $q\in\mathcal{Q}$}
    \State Retrieve images $\mathcal{I}_q\gets\mathcal{R}(q)$ from Wikimedia Commons.
    \State Keep at most $m$ images that pass resolution, file-integrity, and deduplication checks.
    \State Add the filtered images to $\mathcal{D}_{\mathrm{ctx}}$ with binary context label $y{=}1$.
\EndFor
\State Cache frozen CLIP features and per-layer scores for $\mathcal{D}_{\mathrm{ctx}}$.
\State Train the router with domain labels $y{=}0$ for few-shot ID images and $y{=}1$ for WebOE context images.
\State \textbf{return} $\mathcal{D}_{\mathrm{ctx}}$ and cached features/scores.
\Statex \textbf{Constraint:} no target-OOD label, target-OOD training image, or evaluation image is an input to WebOE.
\end{algorithmic}
\end{algorithm}

WebOE first identifies candidate labels that are semantically separated from the in-distribution classes. To this end, we leverage the WordNet hierarchy and measure the shortest-path distance between candidate synsets and the synsets corresponding to the ID labels supplied by the user or benchmark. Labels whose semantic distance exceeds a small threshold (at least two hops in our experiments) are retained as potential OOD categories. As illustrated in Fig.~\ref{fig:wordnet_dist}, this heuristic encourages semantic separation while still allowing a mixture of relatively nearby and more distant concepts. Including semantically closer categories is particularly useful for generating challenging near-OOD examples, which provide informative supervision for learning robust routing behavior. Importantly, this procedure is intended to encourage semantic disjointness with the in-distribution classes, although constructing perfectly disjoint OOD proxy sets remains an open problem.

\begin{figure*}[t]
    \centering
    \includegraphics[width=0.78\textwidth]{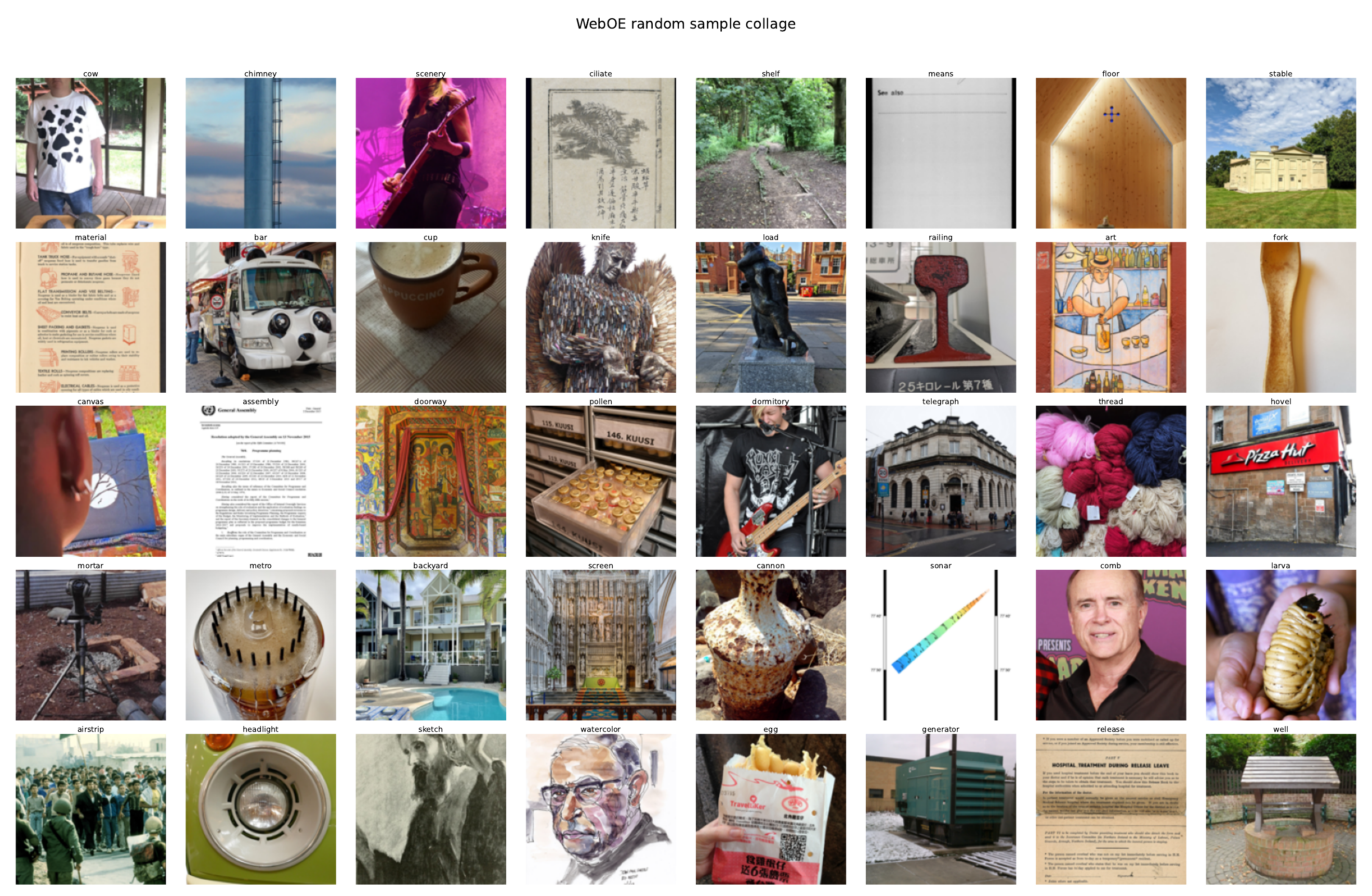}
    \caption{\textbf{Example WebOE retrievals.} Random samples from the Wikimedia Commons proxy context used only for router supervision.}
    \label{fig:examplesofimages}
    \vspace{-1.2em}
\end{figure*}

Once the candidate labels are selected, we retrieve images from Wikimedia Commons\footnote{\url{https://commons.wikimedia.org/wiki/Main_Page}}, an open repository containing over 136 million images, and use them to construct the proxy OOD context for router training. Sampling is balanced across the selected labels so that the proxy context is not dominated by a small number of visual concepts. Example retrieved images are shown in Fig.~\ref{fig:examplesofimages}.

\subsection{Why retrieval-based augmentation works}

\begin{figure}[H]
    \centering
    \captionsetup[subfigure]{skip=1pt}
    \begin{subfigure}{0.49\linewidth}
        \centering
        \includegraphics[width=\linewidth]{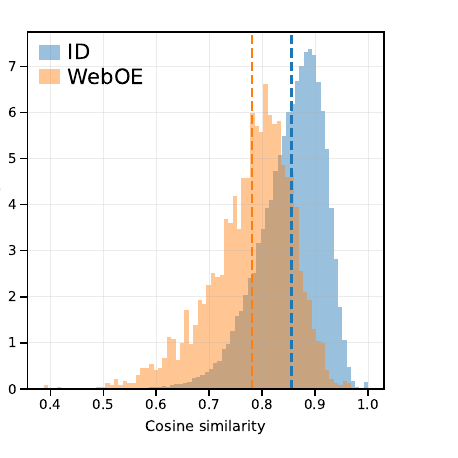}
        \caption{Retrieval similarity}
        \label{fig:sim_b32}
    \end{subfigure}
    \hspace{-0.10\linewidth}
    \begin{subfigure}{0.49\linewidth}
        \centering
        \includegraphics[width=\linewidth]{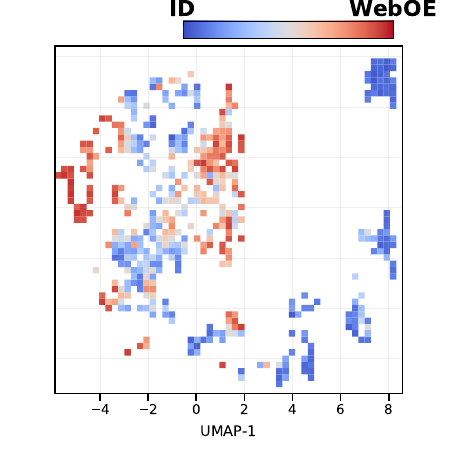}
        \caption{Local structure (UMAP)}
        \label{fig:umap_b32}
    \end{subfigure}
    \vspace{-0.35em}
    \caption{\textbf{WebOE retrieval analysis (ViT-B/32).} Same pattern as ViT-B/16: WebOE samples lie near the ID manifold but are less similar than ID matches. UMAP colors indicate ID-dominated (blue), WebOE-dominated (red), and mixed (pale) neighborhoods.}
    \label{fig:combined_visualizations_b32}
    \vspace{-0.8em}
\end{figure}

To understand why WebOE provides useful supervision for training the router, we analyze the representation similarity between retrieved samples and in-distribution data. Fig.~\ref{fig:sim_b16} and Fig.~\ref{fig:sim_b32} show the cosine similarity between image embeddings and their closest ID neighbors (ImageNet-1K in our main experiments). While ID samples naturally exhibit very high similarity to other ID examples, WebOE images remain relatively close to the ID manifold but are consistently shifted toward lower similarity values. This indicates that retrieved images tend to form \emph{near-OOD} examples: they share semantic or visual structure with the in-distribution classes while still deviating from them. Such samples provide informative supervision, as they expose the model to ambiguous regions where the boundary between ID and OOD is difficult to determine.

We further visualize the embedding space using UMAP projections (Fig.~\ref{fig:umap_b16} and Fig.~\ref{fig:umap_b32}). The resulting maps reveal regions where ID and WebOE samples overlap as well as regions dominated by OOD examples. These mixed regions correspond to areas of high uncertainty where distinguishing ID from OOD requires more nuanced representation cues. Training the router with WebOE therefore encourages the model to learn routing policies that are sensitive to these local shifts in the representation space, allowing it to better exploit intermediate-layer signals when encountering distribution shifts at test time. The ViT-B/16 similarity and UMAP panels appear in the main paper (Fig.~\ref{fig:combined_visualizations}); the ViT-B/32 counterparts (Fig.~\ref{fig:combined_visualizations_b32}) show the same qualitative behavior, confirming that the effect is not backbone-specific.

\subsection*{WebOE construction details (reproducibility)}

We document the exact construction settings used to build the deployed WebOE pool so the proxy context can be reproduced from the released builder.

\paragraph{Candidate vocabulary.}
The candidate OOD vocabulary is drawn from WordNet noun synsets (\texttt{wordnet.all\_synsets(pos="n")}). A synset is retained only if it is semantically disjoint from the ID label set: we form the union of the ID synsets and their full WordNet ancestor and descendant closures (hypernyms/instance-hypernyms and hyponyms/instance-hyponyms), and discard any candidate falling in that union. Surviving candidates are further filtered to a minimum WordNet depth of $5$, to the allowed lexical categories \texttt{noun.animal}, \texttt{noun.plant}, and \texttt{noun.artifact}, and to a ``simple'' primary lemma (lowercase alphabetic, no underscores) whose maximum lemma frequency lies in $[5,80]$ to prefer mid-frequency concrete concepts over hyper-generic or rare ones. Selection is deterministic under a fixed seed (seeded tie-jitter), and the disjointness is verified by an overlap report asserting $|Y_{\text{WebOE}}\cap Y_{\text{ImageNet}}|=0$.

\paragraph{Query formation from labels.}
For each selected synset, search queries are formed from its WordNet lemma names: underscores are replaced by spaces, and only simple alphabetic terms of length $\ge 3$ are kept. From each base term we generate up to three query variants---the bare term, ``\texttt{<term> photo}'', and a lexname hint (e.g.\ ``\texttt{<term> animal}'', ``\texttt{<term> plant}'', ``\texttt{<term> object}'')---capped at $5$ queries per synset, with up to $80$ candidate results fetched per query from the Wikimedia Commons search API.

\paragraph{Images-per-label cap.}
Sampling is balanced across labels: each synset has a target of $20$ images, and per-synset collection stops once that quota is met. To avoid intra-synset redundancy, candidates are additionally capped at a maximum fraction of the per-synset quota from any single query and from any single aspect-ratio bucket.

\paragraph{Duplicate removal.}
Duplicates are removed at three levels: (i) image URL and Commons page id are deduplicated during candidate collection; (ii) exact duplicates are removed by SHA-256 over the re-encoded JPEG bytes; and (iii) near-duplicates are removed with an $8{\times}8$ DCT-based perceptual hash, rejecting any image within a Hamming distance of $8$ of an already-accepted image.

\paragraph{Failed-query and download handling.}
A search query that raises an API error is logged and skipped, and the synset proceeds with its remaining queries. Each image download is attempted up to $2$ times with linear backoff (base $0.4$\,s) and an $18$\,s per-request timeout, retrying on transient HTTP responses ($429$/$503$/$504$); downloads that still fail, or images that fail to decode, are skipped. Synsets that do not reach their quota are recorded as partial and used to backfill the final selection when needed.

\paragraph{Integrity and resolution filtering.}
Every accepted image must have width and height of at least $256$ pixels; SVG/vector files are rejected, EXIF orientation is normalized, transparent (RGBA/LA) images are flattened onto a white background, and images are re-encoded to RGB JPEG (quality $95$). A lightweight quality heuristic down-weights candidates whose title or description matches noisy tokens (e.g.\ logo, icon, diagram, map, flag, chart) and those with extreme aspect ratios, prioritizing photographic content.

\paragraph{Deployed pool.}
Applying this procedure yields the deployed WebOE pool of $5{,}078$ Wikimedia Commons images, materialized as an \texttt{ImageFolder}-compatible union and used only at training time.

\subsection{Proxy-context source ablation}
\label{appendix:proxy-source}

Table~\ref{tab:proxy_ablation} isolates the effect of the proxy-context \emph{source}, holding the router and recipe fixed (deployed $s{=}9$, top-32, $N{=}5$); every context is sampled to the WebOE size. Three findings emerge. (i)~\textbf{ID-only local context is already useful:} a local-region crop proxy built from freshly sampled few-shot ID images, in the spirit of ID-Like and Local-Prompt, reaches $26.71$ FPR@95 and improves over the MoD static-fusion baseline ($30.70$). It also gives the best DTD result among routing proxies ($25.82$), consistent with local ID crops acting as texture-like negative regions. A generic ID-only feature-synthesis proxy ($27.38$; best of cross-class mixup, manifold extrapolation, and noise injection) shows the same trend. (ii)~\textbf{The source still matters:} WebOE, retrieved automatically from the ID label names, remains substantially stronger ($18.86$) than ID-only proxies and every off-the-shelf external dataset, while the best single external source is Flowers ($23.74$). Thus WebOE's advantage is not merely access to arbitrary outlier images, but a useful retrieval-based weak OOD context for the chosen ID distribution. (iii)~\textbf{The supervision is what is learned, not the data:} replacing the true ID/OOD labels with a random permutation collapses performance to $51.36$, worse than the last-layer baseline ($45.64$ in Fig.~\ref{fig:selector_ablation}). Together these results show the gain comes from learning an input-conditioned routing policy under weak supervision, and that ID-derived local regions can also provide a meaningful supervision signal.

\begin{table}[!t]
\centering
\caption{
\textbf{Proxy-context source ablation (ImageNet-1K ID, ViT-B/16, 4-shot).}
Each cell reports FPR@95$\downarrow$/AUROC$\uparrow$; Avg.\ is the macro-average across OOD datasets. Every context is sampled to the WebOE size ($5{,}078$) and trained with the identical router ($s{=}9$, top-32, $N{=}5$); only the outlier-context \emph{source} changes. \name{} (deployed) uses WebOE; the generic ID-only synthetic row uses context generated from the few-shot ID features (best of mixup/extrapolation/noise); the ID-only local-crops row uses small local image regions of freshly-sampled few-shot ID images (ID-Like/Local-Prompt-style). \textbf{Bold} marks the best per column among the routing proxies; the lower block is reference (static MoD; and a shuffled-label control that destroys the ID/OOD supervision).
}
\vspace{-0.2em}
\label{tab:proxy_ablation}

\begingroup
\footnotesize
\setlength{\tabcolsep}{2.0pt}
\renewcommand{\arraystretch}{1.02}
\setlength{\aboverulesep}{0.25ex}
\setlength{\belowrulesep}{0.25ex}
\newcommand{\fprauc}[2]{#1{\scriptsize/}#2}
\resizebox{\linewidth}{!}{%
\begin{tabular}{@{}lccccc@{}}
\toprule
\rowcolor{headerblue}
\textbf{Proxy-context source} &
\textbf{iNat.} & \textbf{SUN} & \textbf{Places} & \textbf{DTD} & \textbf{Avg.} \\
\midrule
\multicolumn{6}{@{}l}{\textit{Retrieved / generated context (routing)}} \\
\rowcolor{tableblue}
\textbf{WebOE} (deployed) &
\fprauc{\textbf{3.93}}{\textbf{98.78}} & \fprauc{\textbf{21.65}}{\textbf{95.13}} & \fprauc{21.88}{93.48} & \fprauc{27.96}{90.58} & \cellcolor{green!34}\fprauc{\textbf{18.86}}{\textbf{94.49}} \\
ID-only local crops (ID-Like) &
\fprauc{22.01}{94.53} & \fprauc{33.95}{93.22} & \fprauc{25.06}{93.44} & \fprauc{\textbf{25.82}}{\textbf{92.45}} & \cellcolor{green!13}\fprauc{26.71}{93.41} \\
Generic ID-only synthetic &
\fprauc{17.37}{95.91} & \fprauc{33.66}{93.41} & \fprauc{25.98}{93.42} & \fprauc{32.52}{90.26} & \cellcolor{green!12}\fprauc{27.38}{93.25} \\
\midrule
\multicolumn{6}{@{}l}{\textit{External proxy datasets (sampled to 5{,}078)}} \\
Oxford Flowers-102 &
\fprauc{15.97}{96.13} & \fprauc{29.79}{94.05} & \fprauc{\textbf{20.83}}{\textbf{94.45}} & \fprauc{28.39}{91.23} & \cellcolor{green!20}\fprauc{23.74}{93.96} \\
Food-101 &
\fprauc{15.32}{96.28} & \fprauc{30.55}{94.09} & \fprauc{22.47}{94.02} & \fprauc{29.10}{91.25} & \cellcolor{green!18}\fprauc{24.36}{93.91} \\
Stanford Cars &
\fprauc{18.79}{95.25} & \fprauc{29.54}{94.05} & \fprauc{22.68}{93.79} & \fprauc{31.99}{89.38} & \cellcolor{green!14}\fprauc{25.75}{93.12} \\
CUB-200 &
\fprauc{19.40}{94.99} & \fprauc{35.00}{92.97} & \fprauc{24.43}{93.71} & \fprauc{35.41}{88.88} & \cellcolor{yellow!12}\fprauc{28.56}{92.64} \\
Mix (all four) &
\fprauc{18.93}{95.71} & \fprauc{35.12}{92.95} & \fprauc{25.50}{93.64} & \fprauc{32.54}{90.17} & \cellcolor{yellow!12}\fprauc{28.02}{93.12} \\
\midrule
\multicolumn{6}{@{}l}{\textit{Reference (no routing / broken supervision)}} \\
MoD static fusion &
\fprauc{15.98}{96.90} & \fprauc{45.58}{89.69} & \fprauc{35.71}{92.72} & \fprauc{25.51}{94.84} & \cellcolor{yellow!18}\fprauc{30.70}{93.54} \\
Shuffled labels (control) &
\fprauc{32.06}{93.20} & \fprauc{52.83}{88.38} & \fprauc{61.07}{84.36} & \fprauc{59.47}{82.32} & \cellcolor{orange!34}\fprauc{51.36}{87.06} \\
\bottomrule
\end{tabular}%
}
\endgroup
\vspace{-1.0em}
\end{table}

\begin{table}[t]
\centering
\caption{\textbf{Direct proxy-classifier control.} Binary ID-vs-WebOE MLPs use the same 4-shot ID data and full WebOE pool, but their classifier probability is used directly as the OOD score instead of routing over frozen experts. Avg.\ FPR@95$\downarrow$ on ImageNet-1K.}
\label{tab:proxy_classifier_control}
\vspace{-0.3em}
\begingroup
\footnotesize
\setlength{\tabcolsep}{3.2pt}
\renewcommand{\arraystretch}{1.03}
\begin{tabular}{@{}lc@{}}
\toprule
\rowcolor{headerblue}
\textbf{Detector} & \textbf{Avg.\ FPR@95}$\downarrow$ \\
\midrule
Direct MLP, final-layer feature & 54.38 \\
Direct MLP, mean-pooled multilayer feature & 56.96 \\
\rowcolor{tableblue}
\textbf{\name{} router, same WebOE context} & \textbf{18.86} \\
\bottomrule
\end{tabular}
\endgroup
\vspace{-0.9em}
\end{table}

Table~\ref{tab:proxy_classifier_control} tests a stricter alternative explanation: perhaps WebOE works simply because a binary ID-vs-WebOE classifier transfers to the benchmark OOD sets. It does not. Direct MLP classifiers trained on the final-layer feature or on the same mean-pooled multilayer feature used by the router both exceed $54$ FPR@95, while \name{} remains at $18.86$. Thus WebOE is useful primarily as supervision for choosing among frozen layer experts, not as a proxy-domain classifier used directly at test time.

\FloatBarrier

\section{Default Architecture and Implementation Details}
\label{appendix:architecture}


\begin{figure*}[t]
\centering
\resizebox{\textwidth}{!}{%
\begin{tikzpicture}[
  font=\small, >={Latex[length=2mm]},
  node distance=7mm and 8mm,
  box/.style={draw, rounded corners, align=center, minimum height=9mm, inner sep=4pt, fill=white},
  proc/.style={box, fill=tableblue},
  score/.style={box, fill=headerblue},
  train/.style={box, fill=insightfill, draw=insightaccent},
  op/.style={draw, circle, inner sep=0pt, minimum size=7mm, fill=white, font=\footnotesize},
  flow/.style={draw, ->, thick},
  trainflow/.style={draw, ->, thick, dashed, insightaccent}]

  \node[box] (img) {image\\$x$};
  \node[proc, right=of img] (clip) {frozen CLIP\\ViT-B/16};
  \node[proc, right=of clip] (layers) {12 visual layers\\$f_0(x),\ldots,f_{11}(x)$};
  \node[score, right=of layers] (scores) {per-layer GL-MCM\\scores $S_0,\ldots,S_{11}$};
  \node[proc, right=of scores] (experts) {anchored sparse\\expert table\\$K{=}1981$, $s{=}9$};
  \node[score, right=of experts] (selected) {selected expert\\$E_{\hat i_r}(x)$};
  \node[op, right=7mm of selected] (avg) {$\frac{1}{5}\sum$};
  \node[score, right=7mm of avg] (out) {final OOD score\\$\mathrm{Score}(x)$};

  \node[proc, below=8mm of layers] (feat) {mean-pooled\\layer features\\$\phi(x)\in\mathbb{R}^{512}$};
  \node[train, below=8mm of experts] (router) {router MLP\\$512{\to}256{\to}128{\to}32{\to}K$\\5 independent seeds};
  \node[train, below=8mm of router] (ctx) {training only\\4-shot ID + WebOE\\FPR@95-aligned soft target};

  \draw[flow] (img) -- (clip);
  \draw[flow] (clip) -- (layers);
  \draw[flow] (layers) -- (scores);
  \draw[flow] (scores) -- (experts);
  \draw[flow] (experts) -- (selected);
  \draw[flow] (selected) -- (avg);
  \draw[flow] (avg) -- (out);

  \draw[flow] (layers) -- (feat);
  \draw[flow] (feat) -- (router);
  \draw[flow] (router) -- (selected);
  \draw[trainflow] (ctx) -- (router);
  \draw[trainflow] (ctx.west) -- ++(-16mm,0) |- node[pos=0.25, left, font=\scriptsize, align=right] {ID\\boundaries} (experts.south);

  \node[below=1mm of experts, font=\scriptsize, align=center] {each expert averages a layer subset\\and always includes layer 11};
  \node[below=1mm of out, font=\scriptsize, align=center] {no WebOE or OOD context\\is used at inference};
\end{tikzpicture}}
\caption{\textbf{Default \name{} architecture.} The frozen scoring path produces layer-wise GL-MCM scores and a fixed dictionary of final-layer-anchored sparse experts. A small router MLP, trained only from 4-shot ID and WebOE context, selects one expert per router; the deployed detector averages the five selected expert scores.}
\label{fig:appendix-architecture}
\end{figure*}
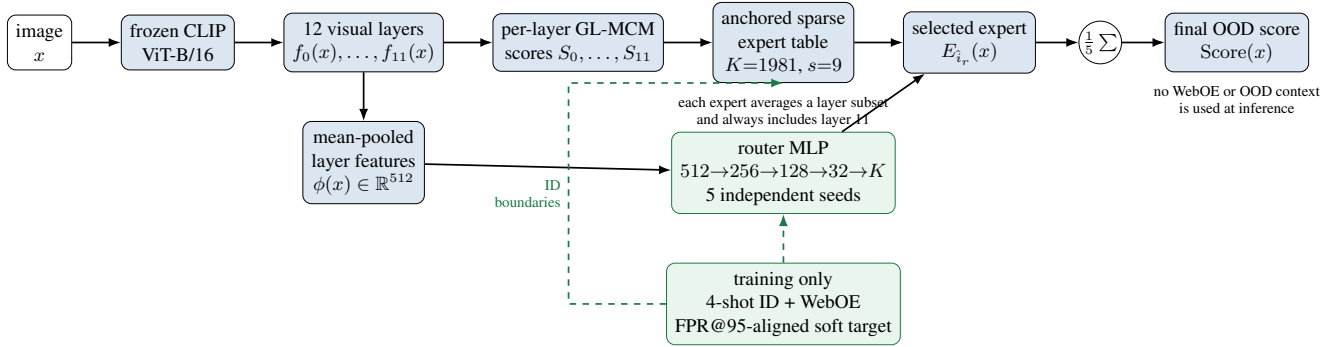

Figure~\ref{fig:appendix-architecture} expands the default \name{} detector used for the main ImageNet-1K results. The architecture is intentionally split into a frozen scoring path and a small trainable routing path. The CLIP image encoder, text encoder, prompts, layer-wise GL-MCM scores, ID thresholds, and expert table are all fixed before router training; only the router MLP parameters are learned.

\paragraph{Frozen scoring path.}
For the default ViT-B/16 setting, CLIP provides $L{=}12$ visual layers. We compute a GL-MCM confidence score at each layer and construct every sparse expert as a mean over a layer subset that always includes the final layer $\ell^\star{=}11$. With sparsity budget $s{=}9$, the candidate family contains
\[
K=\sum_{r=0}^{8}\binom{11}{r}=1981
\]
experts. Each expert has its own ID boundary estimated from the few-shot ID features, and expert scores are simple frozen reductions of the cached layer scores.

\paragraph{Router and ensemble.}
The router input is the mean-pooled layer feature vector $\phi(x)\in\mathbb{R}^{512}$. The default MLP has hidden widths $256,128,32$ and outputs a distribution over the $K$ experts. It is trained for 6 epochs with AdamW, learning rate $10^{-2}$, weight decay $10^{-4}$, batch size 64, and the top-$k$ FPR@95-aligned soft target from Eq.~\ref{eq:soft_target} (deployed $k{=}32$; ablated in Sec.~\ref{appendix:target}). The deployed detector trains five routers with independent seeds and averages their selected expert scores, as in Eq.~\ref{eq:ensemble_score}. This ensemble changes only the scalar score aggregation; it does not require additional CLIP forward passes.

\paragraph{Training versus inference.}
WebOE and the 4-shot ID set are used only to train the router and estimate ID boundaries. At test time, \name{} receives a single image, runs frozen CLIP once, computes the frozen expert table, selects one expert per router, and returns the average selected score. No WebOE image, target-domain OOD example, or proxy-dataset score is used during inference.

\paragraph{Reproducibility and environment.}
All reported numbers use PyTorch~2.10 (CUDA~12.8) on a single GPU, with the five fixed router seeds $\{0,\dots,4\}$ and the cached 4-shot ID draws. Because the router is trained on \emph{cached} per-layer features and GL-MCM scores (the frozen CLIP forward is computed once and stored), the full training-and-evaluation pipeline is deterministic and reproduces run-to-run bit-for-bit within a fixed environment; we release the environment specification and seeds with the code. As is typical of GPU training, the hard $\arg\max$ over the $K{=}1981$ near-tied expert candidates makes the \emph{absolute} FPR@95 mildly sensitive (well under one point) to the CUDA/cuDNN version, so we fix and report the environment to reproduce the headline numbers exactly; the routing conclusions and rankings are unaffected.

\insightbox{Architecture insight: The learned part is small}
{The trainable component is only the router ensemble. The backbone, prompts, per-layer GL-MCM scores, expert definitions, and ID boundaries remain fixed, so \name{} acts as a lightweight selection layer over an already computed frozen-CLIP score table.}

\FloatBarrier

\section{Inference Runtime and Latency}
\label{sec:overhead}

\begin{table}[t]
\centering
\caption{
\textbf{Single-image inference cost (ViT-B/16, one GPU).}
End-to-end latency (raw image to OOD score, \emph{no} cached features), throughput, and peak allocated GPU memory over $300$ runs. \name{} adds only $0.10$\,ms over fixed 12-layer GL-MCM and no measurable peak-memory increase.
}
\vspace{-0.2em}
\label{tab:latency_overhead}

\begingroup
\footnotesize
\setlength{\tabcolsep}{3.4pt}
\renewcommand{\arraystretch}{1.06}
\setlength{\aboverulesep}{0.25ex}
\setlength{\belowrulesep}{0.25ex}
\begin{tabular}{@{}lccc@{}}
\toprule
\rowcolor{headerblue}
\textbf{Detector} & \textbf{Latency (ms)$\downarrow$} & \textbf{Throughput$\uparrow$} & \textbf{Peak GPU$\downarrow$} \\
\midrule
MCM & \cellcolor{green!18}$3.21$ & \cellcolor{green!18}$312$ & \cellcolor{green!18}$0.39$\,GB \\
GL-MCM fixed & \cellcolor{green!10}$4.08$ & \cellcolor{green!10}$245$ & \cellcolor{yellow!10}$0.43$\,GB \\
\midrule
\rowcolor{tableblue}
\textbf{\name{} deployed} & \cellcolor{green!10}$\mathbf{4.18}$ & \cellcolor{green!10}$\mathbf{239}$ & \cellcolor{yellow!10}$\mathbf{0.43}$\,GB \\
\bottomrule
\end{tabular}
\endgroup
\vspace{-1.0em}
\end{table}

Table~\ref{tab:latency_overhead} reports the single-image inference cost of \name{} on ViT-B/16, measured end-to-end from the raw image with no cached features. The deployed detector runs in $4.18$\,ms ($239$ images/s) with a $0.43$\,GB peak GPU footprint. Compared with a fixed 12-layer GL-MCM detector, the input-conditioned router adds only $0.10$\,ms ($+2.5\%$) and no measurable peak-memory increase, because the heavy CLIP forward and per-layer scores are shared across experts.

\FloatBarrier
\section{Selection Mechanism: Learned Routing vs.\ Rules}
\label{appendix:selector}

\definecolor{blue1}{HTML}{3498DB}
\definecolor{green1}{HTML}{1B7A47}
\definecolor{gray1}{HTML}{8A8F98}
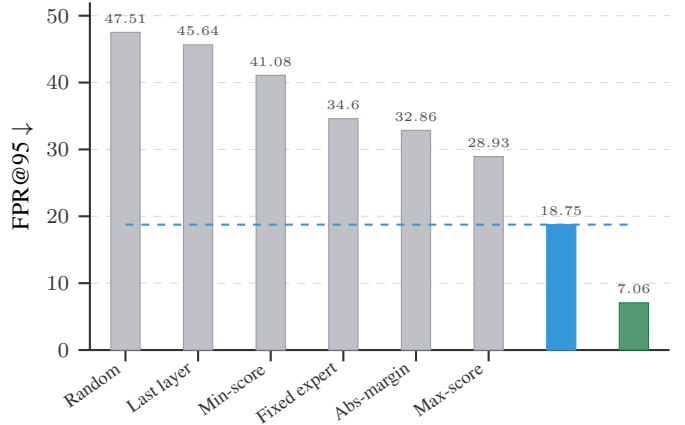
\begin{figure}[t]
\centering
\begin{tikzpicture}
\begin{axis}[
    width=0.92\linewidth, height=4.6cm, scale only axis=true,
    ybar, bar width=11pt,
    symbolic x coords={Random,Last layer,Min-score,Fixed expert,Abs-margin,Max-score,Learned,Oracle}, xtick=data,
    xticklabel style={font=\scriptsize, rotate=35, anchor=east, color=black!80},
    ylabel={FPR@95 $\downarrow$}, ylabel style={font=\small, yshift=-0.4ex},
    ymin=0, ymax=52, ytick={0,10,20,30,40,50},
    tick label style={font=\footnotesize, color=black!80},
    axis line style={thick, black!80}, tick style={thick, black!80}, tick align=outside,
    ymajorgrids=true, grid style={dashed, gray!30},
    axis x line*=bottom, axis y line*=left,
    nodes near coords, nodes near coords style={font=\tiny, color=black!70},
    every node near coord/.append style={anchor=south},
    enlarge x limits=0.07,
]
\addplot[fill=gray1!55, draw=gray1!90, bar shift=0pt] coordinates {(Random,47.51) (Last layer,45.64) (Min-score,41.08) (Fixed expert,34.6) (Abs-margin,32.86) (Max-score,28.93)};
\addplot[fill=blue1, draw=blue1, bar shift=0pt] coordinates {(Learned,18.75)};
\addplot[fill=green1!75, draw=green1, bar shift=0pt] coordinates {(Oracle,7.06)};
\draw[blue1, dashed, thick] (axis cs:Random,18.75) -- (axis cs:Oracle,18.75);
\end{axis}
\end{tikzpicture}
\vspace{-0.8em}
\caption{\textbf{Selection mechanism, not the score (ViT-B/16, WebOE).} All bars use the same
anchored sparse expert family and the same context; only the per-input \emph{selection mechanism}
changes. \textcolor{gray1}{\textbf{Gray}}: score-based selection rules (pick an expert per input by a
heuristic on the expert scores) and the static last-layer/fixed-context experts.
\textcolor{blue1}{\textbf{Blue}}: the learned router (dashed line). \textcolor{green1}{\textbf{Green}}:
the per-input oracle (label-aware headroom). The learned gate beats the best rule (max-score,
28.9) by ${\sim}10$ FPR@95 and is far below the static baselines, confirming that the gain comes
from input-conditioned routing rather than from reading off a max/min/margin of the expert scores.
Configuration: $N{=}8$ routers, 4-shot ID, top-32 target.}
\label{fig:selector_ablation}
\end{figure}

A central question is whether \name{}'s gain comes from \emph{input-conditioned routing} or merely from access to the sparse expert family and its scores. Figure~\ref{fig:selector_ablation} isolates this: every bar uses the same anchored sparse experts, the same frozen GL-MCM scores, and the same WebOE context; only the per-input selection mechanism changes. We compare (i)~static experts (the last layer, and the single best expert chosen on the context), (ii)~score-based selection rules that pick one expert per input from a heuristic on the expert scores (random, min/max expert score, and the largest absolute margin from the per-expert ID boundary), (iii)~the learned router, and (iv)~the per-input label-aware oracle.

The learned router reaches $18.8$ FPR@95, beating the strongest rule (max-score, $28.9$) by roughly ten points and the static last-layer baseline by more than twenty-five. Crucially, the router does not consume the expert scores---it selects from the image feature alone---so these score-based rules serve as representative non-learned baselines, not upper bounds over all non-learned selectors. That they all trail the router shows the benefit is the learned, input-conditioned policy, not the scores themselves. The oracle ($7.1$) marks the remaining headroom of the expert family under perfect per-input selection. (Here the router is the extended-budget $N{=}8$ ensemble, evaluated on cached features; the deployed detector uses $N{=}5$.)

\insightbox{Routing insight: The gate is the mechanism}
{Given identical experts, scores, and context, a learned input-conditioned gate beats every score-based selection rule by ${\sim}10$ FPR@95, the fixed-context expert by ${\sim}16$, and the last-layer baseline by ${>}25$. The improvement is the routing policy, not access to the expert scores.}

\section{Routing Target Ablation}
\label{appendix:target}

\definecolor{blue1}{HTML}{3498DB}
\definecolor{gray1}{HTML}{8A8F98}
\begin{figure}[t]
\centering
\begin{tikzpicture}
\begin{axis}[
    width=0.84\linewidth, height=4.4cm, scale only axis=true,
    ybar, bar width=12pt,
    symbolic x coords={Hard,top-2,top-4,top-8,top-16,top-32,Soft}, xtick=data,
    xticklabel style={font=\scriptsize, rotate=30, anchor=east, color=black!80},
    ylabel={Test FPR@95 $\downarrow$}, ylabel style={font=\small, yshift=-0.4ex},
    ymin=18.8, ymax=22.5,
    tick label style={font=\footnotesize, color=black!80},
    axis line style={thick, black!80}, tick style={thick, black!80}, tick align=outside,
    ymajorgrids=true, grid style={dashed, gray!30},
    axis x line*=bottom, axis y line*=left,
    nodes near coords, nodes near coords style={font=\tiny, color=black!70},
    every node near coord/.append style={anchor=south, yshift=2pt},
    enlarge x limits=0.09,
]
\addplot[fill=gray1!55, draw=gray1!90, bar shift=0pt, error bars/.cd, y dir=both, y explicit,
    error bar style={black!55}] coordinates {(Hard,21.48) +- (0,0.06) (top-2,20.2) +- (0,0.44) (top-4,20.16) +- (0,0.29) (top-8,20.88) +- (0,0.12) (top-16,20.1) +- (0,0.32) (Soft,20.5) +- (0,0.07)};
\addplot[fill=blue1, draw=blue1, bar shift=0pt, error bars/.cd, y dir=both, y explicit,
    error bar style={blue1!70}] coordinates {(top-32,19.66) +- (0,0.24)};
\draw[blue1, dashed, thick] (axis cs:Hard,19.66) -- (axis cs:Soft,19.66);
\end{axis}
\end{tikzpicture}
\vspace{-0.8em}
\caption{\textbf{Routing target ablation (ViT-B/16, WebOE).}
Mean FPR@95 over three independent $N{=}8$ ensembles. A hard top-1 target is worst (21.48), while the top-32 target (dashed line, 19.66) beats the full soft target (20.50) by about 0.8 FPR@95.}
\label{fig:target_ablation}
\end{figure}
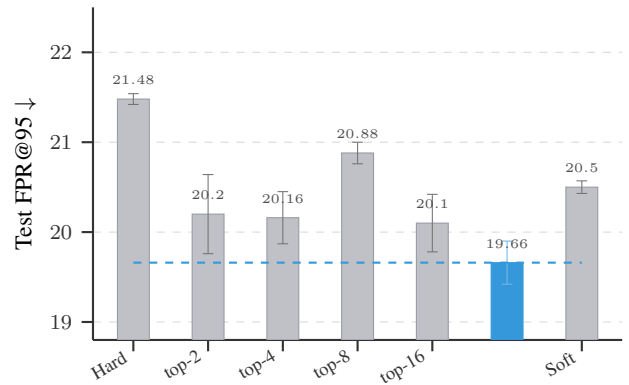

The router is trained with a soft, FPR@95-aligned target rather than a hard one-hot choice of the single best expert. Figure~\ref{fig:target_ablation} ablates the target sharpness by varying the number of target experts $k$ while keeping the rest of the recipe fixed (three independent $N{=}8$ ensembles per setting). A hard top-1 target is robustly worst: at the FPR@95 boundary many experts are near-tied, so committing the target to a single expert is brittle. Relaxing the target helps, but a fully soft target over all $1981$ experts spreads its mass too widely. The top-32 target gives the best result in this diagnostic, beating the full soft target by ${\sim}0.8$ FPR@95 ($19.7$ vs.\ $20.5$) with non-overlapping error bars. We therefore use top-32 in the deployed recipe.

\insightbox{Target insight: Soft, but not too soft}
{A hard one-hot target is brittle because many experts are near-tied at the FPR@95 boundary, while a fully soft target over all experts spreads mass too widely. A soft target restricted to the top-32 near-optimal experts is strongest in this diagnostic.}

\section{Few-shot Scaling of \name{}}
\label{appendix:fewshot}

\definecolor{blue1}{HTML}{3498DB}
\definecolor{green1}{HTML}{1B7A47}
\definecolor{gray1}{HTML}{8A8F98}
\begin{figure}[t]
\centering
\begin{tikzpicture}
\begin{axis}[
    width=0.84\linewidth, height=4.4cm, scale only axis=true,
    xmode=log, log basis x=2,
    xlabel={\# ID shots per class}, ylabel={FPR@95 $\downarrow$}, ylabel style={yshift=-0.5ex},
    ymin=19, ymax=31,
    axis line style={thick, black!80}, tick style={thick, black!80}, tick align=outside,
    label style={font=\small, color=black!90}, tick label style={font=\footnotesize, color=black!80},
    ymajorgrids=true, grid style={dashed, gray!30},
    axis x line*=bottom, axis y line*=left, enlarge x limits=0.08,
    xtick={1,2,4,8,16}, xticklabels={1,2,4,8,16},
    legend style={draw=none, fill=none, font=\scriptsize, at={(0.98,0.98)}, anchor=north east, row sep=-1pt},
    every axis plot/.append style={thick},
]
\addplot[name path=sup, draw=none, forget plot] coordinates {(1,30.06) (2,26.89) (4,26.65) (8,26.93) (16,27.09)};
\addplot[name path=slo, draw=none, forget plot] coordinates {(1,24.04) (2,22.35) (4,22.71) (8,23.37) (16,23.85)};
\addplot[gray1!22, forget plot] fill between[of=sup and slo];
\addplot[gray1, dashed, mark=square*, mark size=1.4pt, mark options={fill=gray1, draw=gray1}] coordinates {(1,27.05) (2,24.62) (4,24.68) (8,25.15) (16,25.47)};
\addlegendentry{single router ($\pm$std)}
\addplot[name path=eup, draw=none, forget plot] coordinates {(1,21.1) (2,19.92) (4,20.52) (8,21.94) (16,22.47)};
\addplot[name path=elo, draw=none, forget plot] coordinates {(1,20.82) (2,19.62) (4,20.34) (8,21.04) (16,22.01)};
\addplot[blue1!18, forget plot] fill between[of=eup and elo];
\addplot[blue1, very thick, mark=*, mark size=1.8pt, mark options={fill=white, draw=blue1},
    error bars/.cd, y dir=both, y explicit,
    error bar style={blue1, thick}, error mark options={blue1, mark size=1.3pt, line width=0.7pt}]
    coordinates {(1,20.96) +- (0,0.14) (2,19.77) +- (0,0.15) (4,20.43) +- (0,0.09) (8,21.49) +- (0,0.45) (16,22.24) +- (0,0.23)};
\addlegendentry{\name{} (N=8 ens., ext.\ budget, $\pm$std)}
\addplot[green1, only marks, mark=star, mark size=2.4pt, mark options={fill=green1, draw=green1}, forget plot] coordinates {(2,19.77)};
\node[font=\scriptsize, text=green1, anchor=north] at (axis cs:2,19.77) {best 19.77};
\end{axis}
\end{tikzpicture}
\vspace{-0.9em}
\caption{\textbf{Few-shot scaling of \name{} (ViT-B/16, WebOE).}
ImageNet-1K FPR@95 vs.\ the number of ID shots per class. \textcolor{gray1}{\textbf{Gray}}: a single router
(mean, with the $\pm$std band over seeds); \textcolor{blue1}{\textbf{blue}}: the extended-budget N=8 ensemble
(error bars/band are $\pm 1$ std over 3 independent ensembles; the deployed detector uses N=5). The ensemble both \emph{lowers}
FPR@95 by ${\sim}4$--$6$ points and \emph{collapses} the seed variance (std ${\sim}0.1$--$0.5$ vs.\ ${\sim}2$--$3$
for a single router). Performance is best at only \textbf{1--2 shots}; more ID data lets the router over-fit
the ID-vs-WebOE boundary, which transfers worse to the evaluation shifts.
Configuration: $N{=}8$ routers, top-32 target; ID shots vary along the $x$-axis.}
\label{fig:shots_curve}
\end{figure}
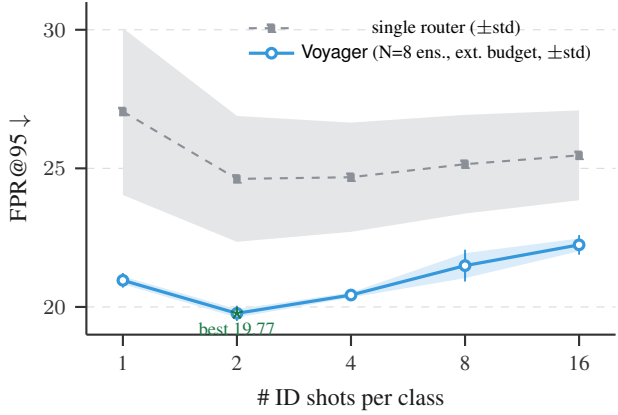

Figure~\ref{fig:shots_curve} reports how \name{} scales with the number of ID shots per class used to train the router, with the WebOE context held fixed. Performance is strongest at \textbf{1--2 shots} and degrades slightly as more ID data is added. The router is trained using ID-versus-context domain supervision to predict an expert-selection policy, so additional ID supervision can let it over-fit the boundary between ID and the WebOE proxy, which transfers less well to the held-out evaluation shifts. This confirms that \name{} needs only minimal labeled ID data, consistent with the few-shot setting used throughout the paper, and mirrors the proxy-size behavior in Sec.~\ref{appendix:WebOE}: a moderate amount of supervision is enough, and more is not always better. This diagnostic uses the extended-budget $N{=}8$ ensemble; the main paper reports the deployed $N{=}5$ detector at the standard $4$-shot setting (Table~\ref{tab:main_results}, $18.86$ FPR@95), so the absolute values should be read through the configuration map in Table~\ref{tab:configs_used}.

\insightbox{Few-shot insight: A little ID data is enough}
{\name{} reaches its best ImageNet-1K FPR@95 with only 1--2 ID shots per class; adding more shots does not help and can mildly hurt, because the router over-fits the ID-vs-proxy boundary rather than the true OOD decision boundary.}

\section{Ensemble Size and Variance Reduction}
\label{appendix:ensemble}

\definecolor{blue1}{HTML}{3498DB}
\definecolor{green1}{HTML}{1B7A47}
\begin{figure}[t]
\centering
\begin{tikzpicture}
\begin{axis}[
    width=0.86\linewidth, height=3.6cm,
    scale only axis=true,
    xlabel={Ensemble size $N$ (routers averaged)},
    ylabel={FPR@95 $\downarrow$},
    ylabel style={yshift=-0.5ex},
    axis line style={thick, black!80},
    tick style={thick, black!80}, tick align=outside,
    label style={font=\small, color=black!90},
    tick label style={font=\footnotesize, color=black!80},
    ymajorgrids=true, grid style={dashed, gray!30},
    axis x line*=bottom, axis y line*=left,
    enlarge x limits=0.04,
    xmin=0.5, xmax=15.5, ymin=18.0, ymax=28.5,
    xtick={1,2,3,4,5,6,7,8,9,10,11,12,13,14,15},
    xticklabel style={font=\scriptsize, color=black!80},
    legend style={draw=none, fill=none, font=\scriptsize, at={(0.98,0.97)}, anchor=north east, row sep=-1pt},
    every axis plot/.append style={thick},
]
\addplot[name path=up, draw=none, forget plot] coordinates {(1,27.90)(2,24.37)(3,22.31)(4,21.79)(5,21.14)(6,20.86)(7,20.22)(8,20.15)(9,20.12)(10,19.85)(11,19.65)(12,19.46)(13,19.40)(14,19.10)(15,18.88)};
\addplot[name path=lo, draw=none, forget plot] coordinates {(1,21.09)(2,19.25)(3,18.63)(4,18.65)(5,18.52)(6,18.51)(7,18.40)(8,18.45)(9,18.64)(10,18.44)(11,18.46)(12,18.62)(13,18.57)(14,18.73)(15,18.88)};
\addplot[blue1!14, forget plot] fill between[of=up and lo];
\addplot[blue1, very thick, mark=*, mark size=1.6pt, mark options={fill=white, draw=blue1}]
    coordinates {(1,21.09)(2,19.25)(3,19.19)(4,19.22)(5,18.86)(6,18.97)(7,18.99)(8,18.75)(9,18.71)(10,18.54)(11,18.46)(12,18.64)(13,18.88)(14,18.82)(15,18.88)};
\addlegendentry{\name{} ensemble}
\addlegendimage{area legend, fill=blue1!14, draw=blue1!14}
\addlegendentry{min--max}
\addplot[green1, only marks, mark=star, mark size=2.6pt, mark options={fill=green1, draw=green1}]
    coordinates {(5,18.86)};
\node[font=\scriptsize, text=green1, fill=white, fill opacity=0.85, text opacity=1,
      inner sep=1.2pt, anchor=north west] at (axis cs:5.35,18.40) {deployed $N{=}5$: 18.86};
\end{axis}
\end{tikzpicture}
\vspace{-0.9em}
\caption{\textbf{Ensemble-size variance reduction (ViT-B/16, 4-shot, top-32 target).}
ImageNet-1K FPR@95 of the \name{} ensemble as routers are added (seeds $0..N{-}1$); shading is the min--max over random $N$-router subsets of a $15$-router pool. Averaging both lowers FPR@95 and shrinks the seed-to-seed spread, with diminishing returns past $N{\approx}5$. The deployed $N{=}5$ detector (star) reaches $\mathbf{18.86}$, matching the main results (Table~\ref{tab:main_results}).}
\label{fig:ensemble_curve}
\end{figure}
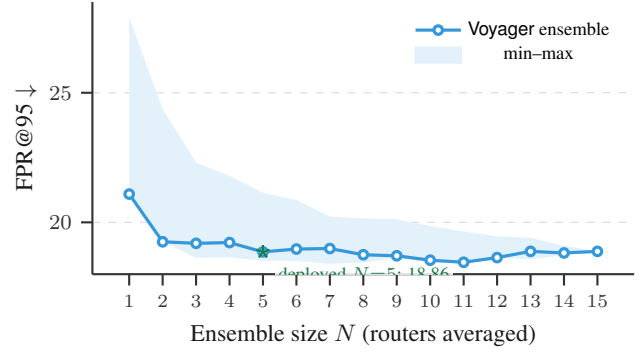

Figure~\ref{fig:ensemble_curve} measures how score-level router averaging trades ensemble size for stability, holding the deployed recipe fixed (4-shot ID, top-32 target, WebOE context). A single router carries seed variance from the random gate initialization---its FPR@95 ranges from $21$ to $28$ across $15$ seeds (std $1.7$). Averaging the per-input scores of $N$ routers shrinks this spread monotonically (std $0.5$ at $N{=}5$, ${<}0.2$ by $N{=}12$) and lowers the FPR@95, which plateaus past $N{\approx}5$. The deployed detector therefore uses $N{=}5$---the seeds-0--4 ensemble, fixed in advance and not selected by benchmark performance, reaches $18.86$ (Table~\ref{tab:main_results})---capturing most of the variance reduction at a fraction of the cost, since the frozen backbone and per-layer scores are shared across routers and only the small MLP heads are replicated. This is the mechanism behind the deterministic test-time inference reported in the main paper: a fixed five-router ensemble yields one scalar score per input.

\insightbox{Ensemble insight: Cheap averaging buys stability}
{Score-level averaging of a few routers converts the seed variance of a single gate into a stable, near-deterministic detector. Five routers capture most of the gain; the cost is negligible because only the tiny router heads are replicated over the shared frozen-CLIP score table.}

\section{Training Baselines under a Shared WebOE Context}
\label{appendix:baseline-training}

To isolate the contribution of \emph{routing} from the contribution of the \emph{supervision signal}, we re-train the prompt-learning baselines using the exact same weak OOD context as \name{}. In this controlled comparison every method receives: (i) the same frozen CLIP ViT-B/16 backbone; (ii) the same 4-shot ImageNet-1K ID supervision; (iii) the same WebOE proxy context (WebOE: $5{,}078$ Wikimedia Commons images built from ID label names and used only at training time); and (iv) the same evaluation protocol (FPR@95/AUROC on iNaturalist, SUN, Places, Texture), with each method scored in its native convention. Only the \emph{method} differs, so any gap reflects how well each approach exploits identical retrieval-based weak supervision rather than differences in the outlier data.

\paragraph{How each baseline consumes the WebOE context.}
The prompt-learning baselines were not originally designed to ingest an arbitrary external outlier pool; each instead derives its negatives internally. We therefore adapt each method \emph{minimally}, leaving its loss and architecture intact and substituting only the source of outliers:
\begin{itemize}
    \item \textbf{LoCoOp}~\cite{miyai2023locoopfewshotoutofdistributiondetection} normally mines OOD regions from the background of the few-shot ID images. We replace this internally mined OOD-region set with the WebOE pool through its external-context branch ($16$ context tokens, $50$ epochs, outlier weight $\lambda{=}0.25$, outlier batch size $4$), keeping its global+local (GL-MCM) test-time scoring.
    \item \textbf{ID-Like}~\cite{bai2024idlikepromptlearningfewshot} normally \emph{synthesizes} ID-like outliers via learnable crops of the ID images. We instead supply WebOE images as the external negative set ($4$-shot ID, $3$ epochs, ID batch size $32$, outlier batch size $32$). Its crop-based outlier construction makes each optimization step expensive, dominating its training cost.
    \item \textbf{Local-Prompt}~\cite{zeng2025localpromptextensiblelocalprompts} normally contrasts positive prompts against learned negative prompts; we attach the WebOE pool as the negative-region context ($16$ context tokens, class-specific contexts, $30$ epochs, $8$ positive / $1$ negative prompts, $300$ negative-prompt bank). We reduce its batch size from $256$ to $64$ so the added outlier batch fits in memory.
\end{itemize}

\paragraph{Cost-measurement protocol.}
All methods are measured with a single identical harness: end-to-end wall-clock training time (single seed), peak GPU memory (sampled from \texttt{nvidia-smi} at $3$\,s intervals), and peak system RAM (resident set of the training processes). \name{} is measured the same way, timing the training of its extended-budget $N{=}8$ router ensemble on the cached frozen features under the same WebOE context (the deployed $N{=}5$ ensemble is cheaper still). This places every point of the efficiency--accuracy comparison on a common footing rather than mixing measured and reported costs.

\paragraph{Reproducibility notes.}
We record three practical considerations for exact reproduction. (i) The WebOE pool is materialized as an \texttt{ImageFolder}-compatible union of the retrieval batches; the per-folder labels in the union are unused, since the pool serves only as an undifferentiated outlier set. (ii) For LoCoOp, the post-training ID-accuracy check in the underlying Dassl trainer expects a single logit tensor and is incompatible with LoCoOp's global+local tuple output; this does not affect the saved prompt checkpoint or the OOD evaluation, which we run as a separate step. (iii) For Local-Prompt, the default batch size ($256$) does not fit on a $24$\,GB GPU once the WebOE outlier batch is added; we reduce it to $32$ (which still uses ${\sim}22.9$\,GB) and enable expandable CUDA memory segments to avoid fragmentation. This changes only the batch size, not the optimization recipe.

\paragraph{Training cost spans more than two orders of magnitude.}
Table~\ref{tab:baseline_training} reports the measured cost of every method under identical supervision, and the spread is large. Training \emph{time} ranges from ${\sim}2.6$ minutes for \name{} to ${\sim}50.9$ minutes (LoCoOp), ${\sim}3.2$ hours (ID-Like), and ${\sim}7$ hours (Local-Prompt)---a factor of roughly $160\times$ between the cheapest and most expensive method. Peak \emph{GPU memory} ranges from under $1$\,GB for \name{} to $13$--$23$\,GB for the prompt-learning baselines. Peak \emph{system RAM} shows a similar gap ($2.8$\,GB vs.\ $28$--$35$\,GB), because \name{} trains on a small cache of frozen per-layer scores while the prompt-learning methods stream the full ImageNet-1K training set through their data loaders.

The reason is structural: prompt learning optimizes \emph{through} the frozen CLIP encoder, so every step must retain the activations needed to back-propagate the full image---and, for the methods studied here, several inner forward passes and an extra outlier batch. \name{} instead trains a small router on cached frozen scores and never runs a CLIP forward or backward pass during training. As a concrete illustration, Local-Prompt's default batch size of $256$ does not fit on a $24$\,GB GPU at all once the WebOE outlier batch is added; even reduced to $32$ it occupies ${\sim}22.9$\,GB ($97\%$ of the card) and takes about seven hours, whereas \name{} trains an extended-budget $N{=}8$ router ensemble in under three minutes using less than one gigabyte (the deployed detector uses $N{=}5$ and is cheaper still).

\paragraph{Result.}
Crucially, this large extra cost does \emph{not} buy the prompt-learning methods a better detector under the shared context. Because their losses are built around ID-derived or synthesized outliers, substituting the same retrieval-based WebOE pool yields detectors weaker than their native recipes (LoCoOp $36.67$, ID-Like $53.60$ FPR@95), whereas the shared-cost \name{} diagnostic reaches $19.12$ with an extended-budget $N{=}8$ router ensemble. This is separate from the deployed Table~\ref{tab:main_results} detector plotted in Fig.~\ref{fig:pareto}, which uses the fixed $N{=}5$ configuration and reaches $18.86$ (Table~\ref{tab:configs_used}); the two rows are not intended as a router-count ablation. This supports our central claim that the gain comes from input-conditioned routing over the frozen score table, not merely from access to weak outlier data---and that it is obtained at a small fraction of the training budget.

\insightbox{Appendix insight: Cheaper \emph{and} better under the same context}
{Given the same retrieval-based WebOE pool, \name{} is both the strongest detector and by far the cheapest to train: ${\sim}160\times$ less wall-clock time and ${>}20\times$ less GPU and system memory than the most expensive baseline. Prompt-learning methods pay a heavy cost to optimize through CLIP, yet do not convert the external context into a better detector.}

\begin{table}[t]
\centering
\footnotesize
\setlength{\tabcolsep}{2.3pt}
\renewcommand{\arraystretch}{1.12}
\caption{\textbf{Shared WebOE training cost.} ViT-B/16, 4-shot ID, WebOE ($5{,}078$ images). Avg.\ FPR@95 is averaged over iNaturalist, SUN, Places, and Texture; the $N{=}8$ \name{} row is the shared-cost diagnostic, not the deployed Table~\ref{tab:main_results} configuration.}
\label{tab:baseline_training}
\begin{tabular}{@{}lcccc@{}}
\toprule
\textbf{Method} & \textbf{Train time} & \textbf{Peak GPU} & \textbf{Peak RAM} & \textbf{Avg.\ FPR@95}$\downarrow$ \\
\midrule
LoCoOp        & $50.9$\,min        & $18.5$\,GB & $28.3$\,GB & $36.67$ \\
ID-Like       & ${\sim}3.2$\,h      & $13.1$\,GB & $35.3$\,GB & $53.60$ \\
Local-Prompt  & $7.1$\,h            & $22.9$\,GB & $11.7$\,GB & $31.13$ \\
\rowcolor{tableblue}
\textbf{\name{} ($N{=}8$)} & ${\sim}\mathbf{2.6}$\,min & $\mathbf{0.9}$\,GB & $\mathbf{2.8}$\,GB & $\mathbf{19.12}$ \\
\bottomrule
\end{tabular}
\end{table}

\FloatBarrier

\section{Additional Backbones and Detection Frameworks}
\label{appendix:backbones}

Table~\ref{tab:backbone_results} extends \name{} beyond the ViT-B/16 main results along two axes: additional backbones and additional detection frameworks. With a four-shot router, the learned selector improves the average FPR@95 over the static MoD fusion on both ViT-B/32 ($28.62$ vs.\ $29.01$) and the RN50x4 convolutional backbone ($40.31$ vs.\ $41.61$). The ViT-B/32 margin is narrow---MoD remains stronger on iNaturalist---whereas on ViT-B/16 the gap is large ($18.86$ vs.\ $30.70$ for MoD), with ViT-B/32's optimum at a slightly smaller expert budget ($s{=}8$): the routing gain scales with the intermediate-layer diversity a backbone exposes, so it is largest on the stronger ViT-B/16. As a plug-in, \name{} also lowers the FPR@95 of the prompt-learning detector LoCoOp ($28.66{\to}24.79$) and the negative-label detector CSP ($17.51{\to}15.41$), indicating that input-conditioned routing composes with stronger base detectors rather than competing with them.

\begin{table}[!t]
\centering
\caption{
\textbf{Additional CLIP backbones and OOD-detection frameworks (ImageNet-1K ID, 4-shot).}
Each cell reports FPR@95$\downarrow$/AUROC$\uparrow$; Avg.\ is the macro-average across OOD datasets. \name{} acts as a four-shot routing layer within each group; the ViT-B/32 \name{} row uses the deployed recipe ($s{=}8$, top-32, $N{=}5$). Baselines are from the respective source methods. \textbf{Bold} marks the best per column within a group.
}
\vspace{-0.2em}
\label{tab:backbone_results}

\begingroup
\footnotesize
\setlength{\tabcolsep}{2.0pt}
\renewcommand{\arraystretch}{1.02}
\setlength{\aboverulesep}{0.25ex}
\setlength{\belowrulesep}{0.25ex}
\newcommand{\fprauc}[2]{#1{\scriptsize/}#2}
\resizebox{\linewidth}{!}{%
\begin{tabular}{@{}lccccc@{}}
\toprule
\rowcolor{headerblue}
\textbf{Method} &
\textbf{iNat.} & \textbf{SUN} & \textbf{Places} & \textbf{DTD} & \textbf{Avg.} \\
\midrule
\multicolumn{6}{@{}l}{\textit{Frozen-backbone, ViT-B/32}} \\
MCM &
\fprauc{33.85}{93.62} & \fprauc{40.99}{91.56} & \fprauc{46.71}{89.25} & \fprauc{60.90}{85.03} & \cellcolor{orange!22}\fprauc{45.61}{89.87} \\
\textbf{MoD} &
\fprauc{\textbf{12.02}}{\textbf{97.64}} & \fprauc{28.98}{\textbf{93.37}} & \fprauc{35.69}{\textbf{91.61}} & \fprauc{39.36}{\textbf{91.07}} & \cellcolor{yellow!16}\fprauc{29.01}{\textbf{93.42}} \\
\rowcolor{tableblue}
\textbf{\name{}} &
\fprauc{21.41}{95.58} & \fprauc{\textbf{26.56}}{93.17} & \fprauc{\textbf{34.54}}{90.79} & \fprauc{\textbf{31.95}}{90.00} & \cellcolor{yellow!22}\fprauc{\textbf{28.62}}{92.39} \\
\midrule
\multicolumn{6}{@{}l}{\textit{Frozen-backbone, RN50x4}} \\
MCM &
\fprauc{44.00}{91.59} & \fprauc{35.13}{92.83} & \fprauc{44.30}{89.38} & \fprauc{57.22}{85.99} & \cellcolor{orange!22}\fprauc{45.16}{89.95} \\
\textbf{MoD} &
\fprauc{41.79}{92.00} & \fprauc{\textbf{31.83}}{\textbf{93.48}} & \fprauc{\textbf{40.46}}{\textbf{90.36}} & \fprauc{\textbf{52.38}}{87.37} & \cellcolor{orange!14}\fprauc{41.61}{\textbf{90.80}} \\
\rowcolor{tableblue}
\textbf{\name{}} &
\fprauc{\textbf{34.18}}{\textbf{92.66}} & \fprauc{32.80}{92.79} & \fprauc{40.62}{89.01} & \fprauc{53.67}{\textbf{88.10}} & \cellcolor{orange!10}\fprauc{\textbf{40.31}}{90.64} \\
\midrule
\multicolumn{6}{@{}l}{\textit{Frozen-backbone, ViT-L/14}} \\
MCM &
\fprauc{46.19}{91.42} & \fprauc{32.58}{93.34} & \fprauc{37.02}{91.86} & \fprauc{54.95}{86.19} & \cellcolor{orange!16}\fprauc{42.68}{90.70} \\
\rowcolor{tableblue}
\textbf{\name{}} &
\fprauc{\textbf{16.08}}{\textbf{96.94}} & \fprauc{\textbf{24.67}}{\textbf{94.02}} & \fprauc{\textbf{26.53}}{\textbf{92.51}} & \fprauc{\textbf{43.90}}{\textbf{88.05}} & \cellcolor{green!12}\fprauc{\textbf{27.80}}{\textbf{92.88}} \\
\midrule
\multicolumn{6}{@{}l}{\textit{Prompt-learning, ViT-B/16}} \\
LoCoOp$_\text{GL}$ &
\fprauc{16.05}{96.86} & \fprauc{\textbf{23.44}}{\textbf{95.07}} & \fprauc{\textbf{32.87}}{\textbf{91.98}} & \fprauc{42.28}{90.19} & \cellcolor{yellow!16}\fprauc{28.66}{93.52} \\
\rowcolor{tableblue}
\textbf{LoCoOp\,+\,\name{}} &
\fprauc{\textbf{4.75}}{\textbf{98.91}} & \fprauc{33.29}{92.03} & \fprauc{35.31}{90.67} & \fprauc{\textbf{25.82}}{\textbf{93.57}} & \cellcolor{green!14}\fprauc{\textbf{24.79}}{\textbf{93.80}} \\
\midrule
\multicolumn{6}{@{}l}{\textit{Negative-label, ViT-B/16}} \\
CSP &
\fprauc{1.54}{99.60} & \fprauc{13.66}{96.66} & \fprauc{29.32}{92.90} & \fprauc{25.52}{93.86} & \cellcolor{green!26}\fprauc{17.51}{95.76} \\
\rowcolor{tableblue}
\textbf{CSP\,+\,\name{}} &
\fprauc{\textbf{1.26}}{\textbf{99.64}} & \fprauc{\textbf{11.26}}{\textbf{97.22}} & \fprauc{\textbf{25.43}}{\textbf{94.21}} & \fprauc{\textbf{23.69}}{\textbf{95.55}} & \cellcolor{green!32}\fprauc{\textbf{15.41}}{\textbf{96.65}} \\
\bottomrule
\end{tabular}%
}
\endgroup
\vspace{-1.0em}
\end{table}

\paragraph{ViT-L/14.}
We additionally verified \name{} on the larger ViT-L/14 backbone ($24$ transformer layers), at our general $T{=}1$ scoring temperature with the standard CLIP prompt ensemble (which benefits L/14, whereas a single prompt is stronger on ViT-B/16). Because L/14's early layers are noisier---a static average over all $24$ layers degrades sharply---the optimal expert budget stays modest relative to the $24$ layers ($s{=}4$, versus $s{=}9$ on the $12$-layer ViT-B/16). On this setup, \name{} ($N{=}5$, 4-shot) attains $\mathbf{27.8}$ FPR@95, a $14.9$-point improvement over the static last-layer MCM ($42.7$), showing that input-conditioned routing continues to help---and helps more---on a larger backbone.

\FloatBarrier
\section{Complementarity Between Experts}
\label{appendix:extra-results-complementarity}

\begin{figure*}[t]
    \centering
    \includegraphics[width=\textwidth]{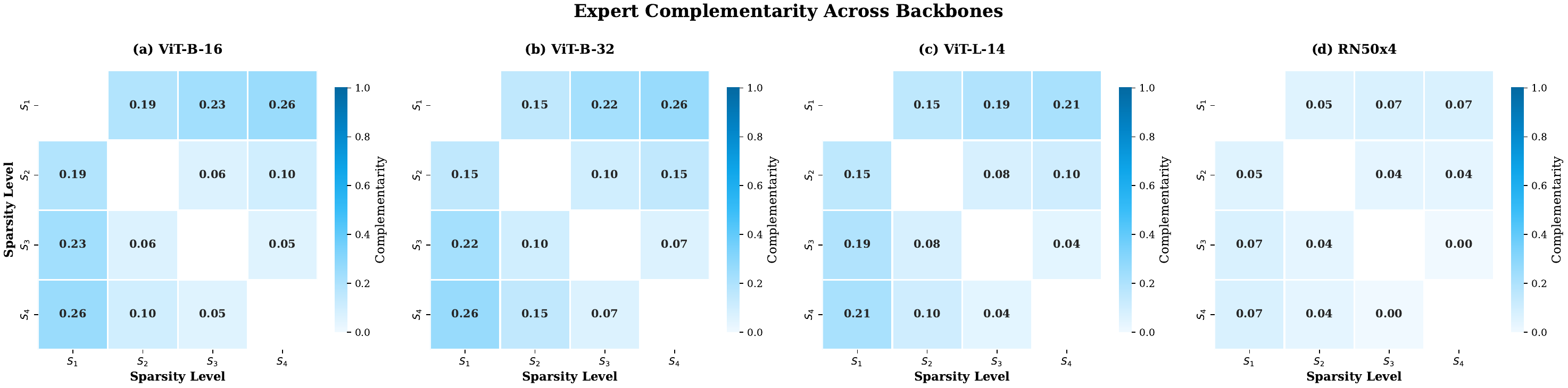}
    \caption{\textbf{Expert complementarity across CLIP backbones.}
    Pairwise complementarity matrices for ViT-B/16, ViT-B/32, ViT-L/14, and RN50x4.
    All backbones exhibit non-trivial complementarity structure, confirming that 
    inter-expert diversity is a general and backbone-agnostic property.}
    \label{fig:backbone_complementarity}
\end{figure*}

We analyze the complementarity between experts generated at different sparsity levels, using ImageNet-1K as the in-distribution dataset and SUN, Places, iNaturalist, and Textures as OOD benchmarks. The \textbf{complementarity score} between two experts is defined as:

\begin{equation}
C =
\frac{\text{Unique Detections}}{\text{Total Detections}}
=
\frac{\text{only}_1 + \text{only}_2}{\text{only}_1 + \text{only}_2 + \text{both}},
\end{equation}

where $\text{only}_1$ and $\text{only}_2$ denote samples detected exclusively by one expert, and $\text{both}$ denotes samples detected by both. A value of $C=0$ indicates identical behavior, whereas $C=1$ corresponds to fully complementary experts.

To characterize expert diversity, we evaluate complementarity under a \textbf{retrospective per-instance selector}, where the best expert is chosen independently for each instance using labels that are unavailable at deployment. This is an analysis diagnostic rather than a practical detector. Figure~\ref{fig:backbone_complementarity} reports the pairwise complementarity matrices across sparsity levels for four CLIP backbones.

Several consistent trends emerge. First, experts associated with \emph{distant sparsity levels} exhibit the strongest complementarity. For example, ViT-B/16 and ViT-B/32 reach complementarity scores up to $C\approx0.26$ between the most distant sparsity configurations, indicating that these experts frequently detect different subsets of OOD samples. In contrast, nearby sparsity levels show substantially lower complementarity (e.g., $C\approx0.05$--$0.10$), suggesting partial redundancy.

Second, transformer backbones exhibit substantially richer complementarity patterns than convolutional ones. ViT-B/16 and ViT-B/32 display the highest diversity across experts, while ViT-L/14 shows slightly reduced complementarity (maximum $C\approx0.21$), suggesting that deeper transformer models produce somewhat more homogeneous expert behavior. In contrast, RN50x4 exhibits consistently low complementarity (typically $C\leq0.07$), indicating that experts derived from different sparsity levels capture more similar signals.

\paragraph{Retrospective sparsity sweep.}
Figure~\ref{fig:sparsity_vs_fpr} reports the main sparsity diagnostic deferred from the paper body. Increasing the sparsity budget enlarges the candidate expert family and consistently lowers retrospective FPR@95 relative to the last-layer baseline, confirming that useful intermediate evidence is not exhausted by a single fixed subset.

\definecolor{vColor}{RGB}{53,92,125}   
\definecolor{mColor}{RGB}{242,142,43}  
\definecolor{oColor}{RGB}{27,158,119}  

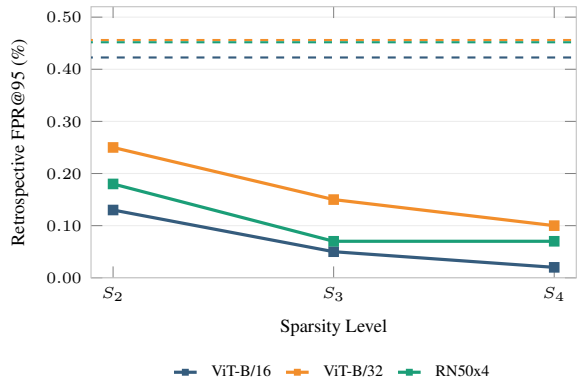
\begin{figure}[t]
    \centering
    \begin{tikzpicture}
    \begin{axis}[
        width=0.96\linewidth,
        height=0.62\linewidth,
        xlabel={Sparsity Level},
        ylabel={Retrospective FPR@95 (\%)},
        xmin=1.9, xmax=4.1,
        xtick={2,3,4},
        xticklabels={$S_2$, $S_3$, $S_4$},
        ymin=0, ymax=0.52,
        ytick={0, 0.1, 0.2, 0.3, 0.4, 0.5},
        y tick label style={
            /pgf/number format/.cd, fixed, fixed zerofill, precision=2,
            /tikz/.cd, font=\fontsize{6}{7}\selectfont
        },
        ymajorgrids,
        grid style={gray!15},
        axis line style={gray!50},
        label style={font=\fontsize{7}{8}\selectfont},
        tick label style={font=\fontsize{6}{7}\selectfont},
        legend style={
            at={(0.5,-0.28)},
            anchor=north,
            legend columns=-1,
            draw=none,
            fill=none,
            font=\fontsize{6}{7}\selectfont,
            column sep=3pt, 
            nodes={inner xsep=2pt}, 
        },
        legend image post style={scale=0.5}
    ]
    
    \addplot[vColor, dashed, line width=0.8pt, forget plot] coordinates {(1.8,0.4226) (4.2,0.4226)};
    \addplot[mColor, dashed, line width=0.8pt, forget plot] coordinates {(1.8,0.4561) (4.2,0.4561)};
    \addplot[oColor, dashed, line width=0.8pt, forget plot] coordinates {(1.8,0.4518) (4.2,0.4518)};
    
    \addplot[vColor, line width=1.2pt, mark=square*, mark size=1.4pt]
        coordinates {(2,0.13) (3,0.05) (4,0.02)};
    \addlegendentry{ViT-B/16}

    \addplot[mColor, line width=1.2pt, mark=square*, mark size=1.4pt]
        coordinates {(2,0.25) (3,0.15) (4,0.10)};
    \addlegendentry{ViT-B/32}

    \addplot[oColor, line width=1.2pt, mark=square*, mark size=1.4pt]
        coordinates {(2,0.18) (3,0.07) (4,0.07)};
    \addlegendentry{RN50x4}
    
    \end{axis}
    \end{tikzpicture}
    \vspace{-0.9em}
    \caption{\textbf{Sparsity diagnostic.} Comparison across architectures. Dashed lines: last-layer reference; solid lines: retrospective per-instance selection, used only to measure routing opportunity.}
    \label{fig:sparsity_vs_fpr}
    \vspace{-0.8em}
\end{figure}

\section{Random Dynamic Selection as a Zero-Shot Method}
\label{app:random_selection}

\begin{figure}[t]
    \centering    
    \begin{subfigure}[t]{\linewidth}
        \centering
        \includegraphics[width=\linewidth]{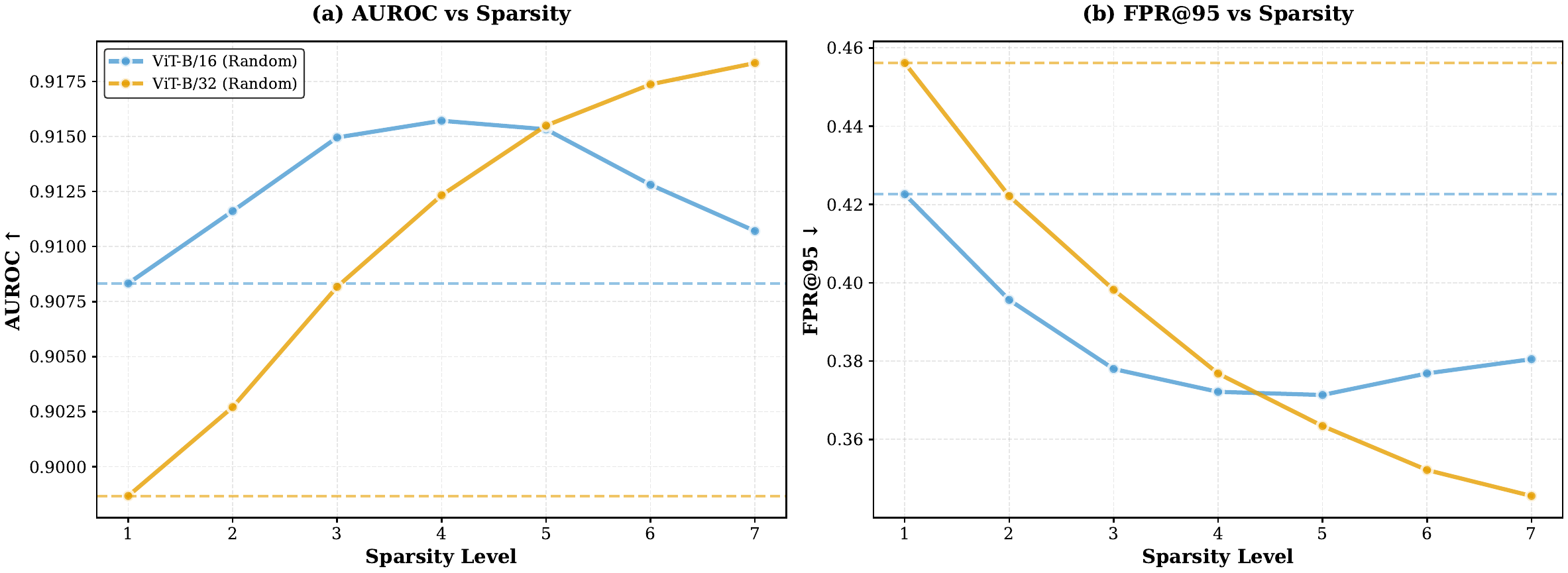}
        \caption{Effect of sparsity on performance for each backbone.}
        \label{fig:vit_sparsity_analysis}
    \end{subfigure}
    \caption{The impact of sparsity differs across architectures: ViT-B/32 benefits most at higher sparsity, while ViT-B/16 reaches its optimum near $s{=}5$.}
    \label{fig:vit_sparsity_backbone_analysis}
    \vspace{-2em}
\end{figure}

As shown in the previous sections, dynamic layer selection improves OOD detection by exploiting the complementary information distributed across intermediate representations. This naturally raises the question of whether similar gains could arise \emph{without learning}, simply by selecting layers at random. To examine this, we randomly sample layer combinations under different sparsity levels and evaluate their effectiveness and variability for OOD detection.

Figure~\ref{fig:vit_sparsity_backbone_analysis} shows that random selection can already provide noticeable gains for some backbones, confirming that intermediate-layer diversity is itself informative. In particular, both ViT-B/16 and ViT-B/32 achieve improvements of more than $12$ points in FPR@95 over the last-layer baseline on most OOD datasets. However, these gains are less stable than those obtained with learned routing, and the variance tends to increase with sparsity, especially on SUN. This suggests that, while many random combinations are beneficial, their effectiveness depends strongly on which layers are selected.

The effect also differs across architectures, reflecting backbone-specific patterns of layer complementarity. Although ViT-B/32 starts from a weaker last-layer baseline, it benefits the most from random selection at high sparsity, indicating greater untapped diversity across its intermediate layers. By contrast, ViT-B/16 reaches its best performance at $s=5$, after which the gains saturate.

The relative improvement further depends on the evaluation dataset. DTD and iNaturalist show the largest gains, suggesting that random fusion is particularly effective when OOD cues are tied to texture and fine-grained visual structure. In contrast, SUN and Places exhibit smaller and less stable improvements, likely because scene-centric shifts require more precise layer selection to capture higher-level semantic variation. Overall, these results show that part of the benefit comes from the richness of intermediate representations themselves, but also highlight that learning how to select layers remains important for achieving consistent performance.

\subsection{Why Can Random Layer Selection Work?}

To understand the empirical effectiveness of random selection, we analyze the performance distribution of all possible layer combinations across sparsity levels.
Let $\mathcal{C}_s$ denote the set of all layer combinations of size $s$ (always including the last layer), and let $\Delta_i$ represent the improvement in AUROC (or reduction in FPR95) of combination $i$ over the last layer baseline.
The distribution of $\{\Delta_i \mid i \in \mathcal{C}_s\}$ exhibits a right-skewed shape, meaning that while most combinations perform near the mean, a non-negligible fraction achieve substantial gains.

Formally, the probability that a random combination improves over the baseline is:
\begin{equation}
    P_{\text{better}}(s) = \mathbb{P}[\Delta_i > 0], \quad i \sim \mathcal{U}(\mathcal{C}_s),
\end{equation}
and the expected improvement from random sampling is:
\begin{equation}
    \mathbb{E}[\Delta \mid s] = \frac{1}{|\mathcal{C}_s|} \sum_{i \in \mathcal{C}_s} \Delta_i.
\end{equation}

\begin{figure}[t]
    \centering
    \begin{adjustbox}{width=\columnwidth}
    \definecolor{vitblue}{RGB}{31,119,180}
\definecolor{vitorange}{RGB}{255,127,14}
\begin{tikzpicture}
\begin{groupplot}[
  group style={
    group size=2 by 1,
    horizontal sep=0.12cm,
  },
  grid=both,
  grid style={line width=0.2pt, draw=gray!40},
  major grid style={line width=0.25pt, draw=gray!50},
  minor grid style={line width=0.15pt, draw=gray!30},
  width=0.46\columnwidth,
  height=0.31\columnwidth,
  xmin=1, xmax=12,
  ymin=0, ymax=1.0,
  xtick={1,3,5,7,9,11},
  xlabel={Sparsity $s$},
  ylabel={Prob.},
  tick label style={font=\scriptsize},
  label style={font=\scriptsize},
  title style={font=\scriptsize, yshift=-2pt},
  legend style={
    font=\tiny,
    cells={anchor=west},
    fill=white,
    fill opacity=0.8,
    text opacity=1,
    draw=gray!60,
  },
  scaled y ticks=false
]

\nextgroupplot[
  title={(a)~ViT-B/32},
  legend columns=1,
  legend style={
    font=\tiny,
    cells={anchor=west},
    fill=white,
    fill opacity=0.8,
    text opacity=1,
    draw=gray!60,
    at={(0.98,0.04)},
    anchor=south east,
  },
]
\addplot[vitblue, thick, mark=*, mark size=1.6pt] coordinates {
  (1,0.00)(2,0.69)(3,0.79)(4,0.85)(5,0.87)(6,0.87)
  (7,0.86)(8,0.86)(9,0.85)(10,0.85)(11,0.85)(12,0.85)
};
\addlegendentry{AUROC $>$ Base}

\addplot[vitorange, thick, dashed, mark=square*, mark size=1.5pt] coordinates {
  (1,0.00)(2,0.67)(3,0.83)(4,0.88)(5,0.90)(6,0.91)
  (7,0.91)(8,0.91)(9,0.91)(10,0.91)(11,0.91)(12,0.91)
};
\addlegendentry{FPR@95 $<$ Base}

\nextgroupplot[
  title={(b)~ViT-B/16},
  yticklabels=\empty,
  ylabel={},
]
\addplot[vitblue, thick, mark=*, mark size=1.6pt] coordinates {
  (1,0.00)(2,0.62)(3,0.70)(4,0.69)(5,0.67)(6,0.66)
  (7,0.64)(8,0.63)(9,0.62)(10,0.61)(11,0.61)(12,0.61)
};
\addplot[vitorange, thick, dashed, mark=square*, mark size=1.5pt] coordinates {
  (1,0.00)(2,0.62)(3,0.72)(4,0.71)(5,0.69)(6,0.66)
  (7,0.63)(8,0.62)(9,0.60)(10,0.60)(11,0.60)(12,0.60)
};

\end{groupplot}

\end{tikzpicture}
    \end{adjustbox}
    
    \caption{\textbf{Probability of selecting a better expert across sparsity levels for ViT-B/32 and ViT-B/16.}
    (\textbf{a}) ViT-B/32 shows the strongest effect, with probabilities exceeding \textbf{0.9} for higher sparsity levels.
    (\textbf{b}) For ViT-B/16, the probability that a randomly selected expert outperforms the last layer remains above $0.6$ across most sparsity levels.}
    \label{fig:vit_sparsity_backbone_analysis_2}
\end{figure}
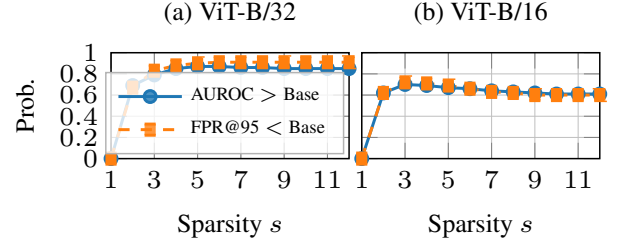

Empirically from~\ref{fig:vit_sparsity_backbone_analysis_2}, we find $P_{\text{better}}(s) > 0.6$ for moderate $s$, showing that even naive sampling frequently selects combinations outperforming the fixed-layer baseline.
Empirically, random sparse subsets that include the final layer often retain useful separation; we make no claim that a uniformly sampled subset is guaranteed to improve, and present these as empirical observations.

\begin{figure}[!t]
    \centering
    \includegraphics[width=1\linewidth]{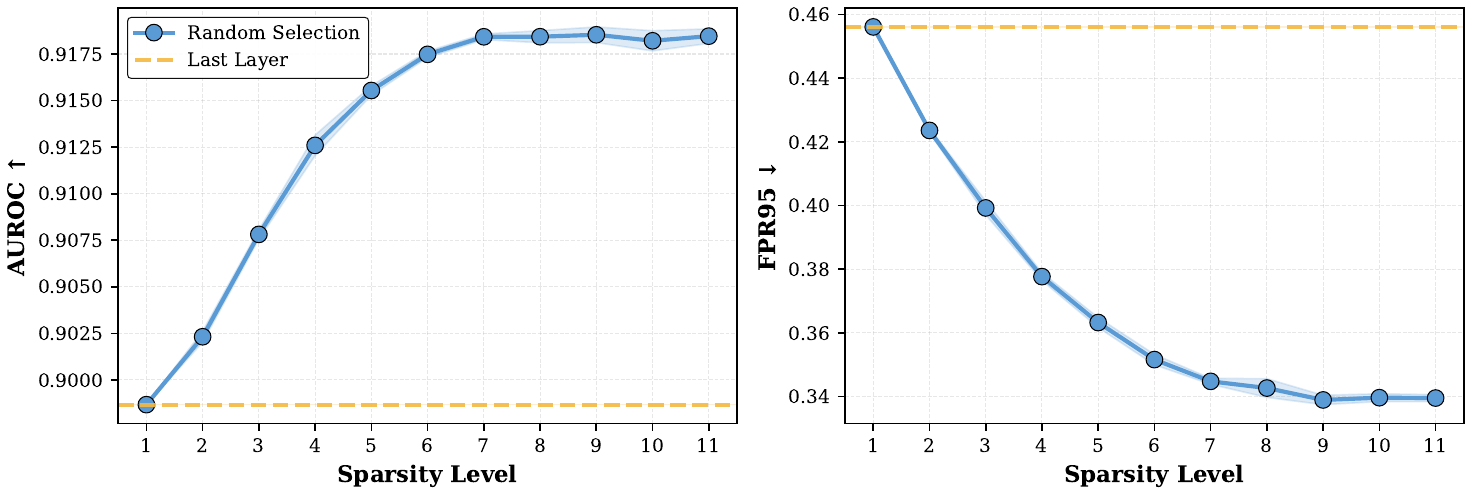}
    \caption{\textbf{AUROC and FPR95 as a function of sparsity.} Random selection consistently improves with higher sparsity and reliably outperforms the last layer across both metrics.}
    \label{fig:randomness_sparsity_general}
\end{figure}

\begin{figure}[!t]
    \centering
    \includegraphics[width=1\linewidth]{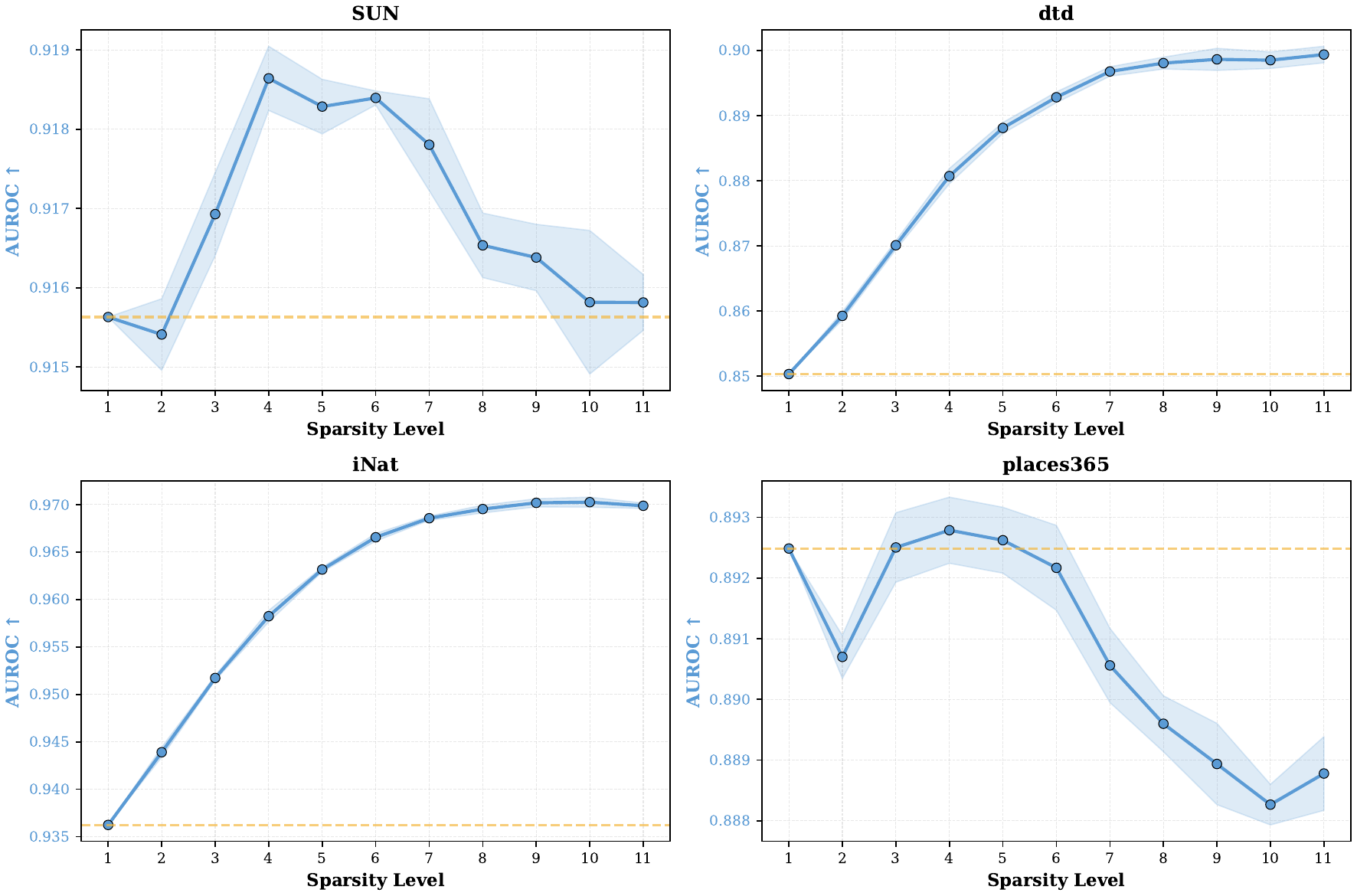}
    \caption{\textbf{AUROC across sparsity levels for SUN, DTD, iNaturalist, and Places365.}}
    \label{fig:randomness_sparsity_datasets}
\end{figure}

\subsection{Randomness in the Random Selection}

We repeat the random-selection experiment five times, and the resulting curves in Fig.~\ref{fig:randomness_sparsity_general} are almost identical across runs, confirming that the method is highly stable. This stability reflects the high probability of sampling at least one informative expert, which allows random combinations to consistently outperform the last layer. The more nuanced behavior appears when analyzing each dataset independently. As shown in Fig.~\ref{fig:randomness_sparsity_datasets}, the variance increases slightly for SUN, DTD, iNaturalist, and Places365, but this variation remains confined to a narrow performance band. This indicates that although datasets respond differently to sparsity, the random strategy remains reliable and does not collapse under dataset-specific fluctuations.

\subsection{Distribution of Experts Across Sparsity Levels}
\label{app:distribution_experts_sparsity}

\begin{figure}[!h]
    \centering
    \begin{minipage}{0.48\textwidth}
        \centering
        \includegraphics[width=\linewidth]{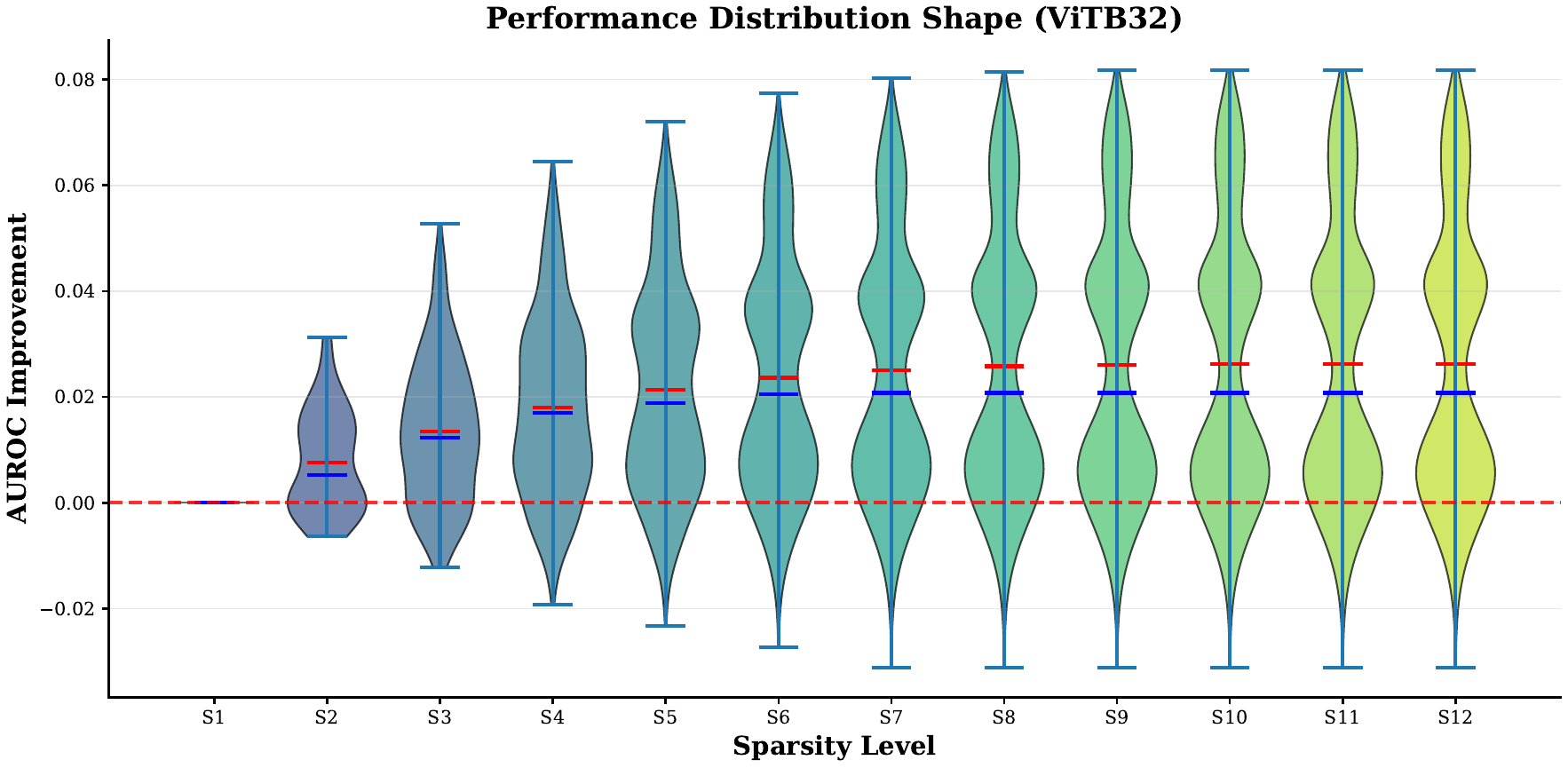}
        \caption{\textbf{Distribution of expert improvements for ViT-B/32.} 
        Violin plots show AUROC improvements $\Delta$ for all experts at each sparsity level.}
        \label{fig:vitb32_violin}
    \end{minipage}
    \hfill
    \begin{minipage}{0.48\textwidth}
        \centering
        \includegraphics[width=\linewidth]{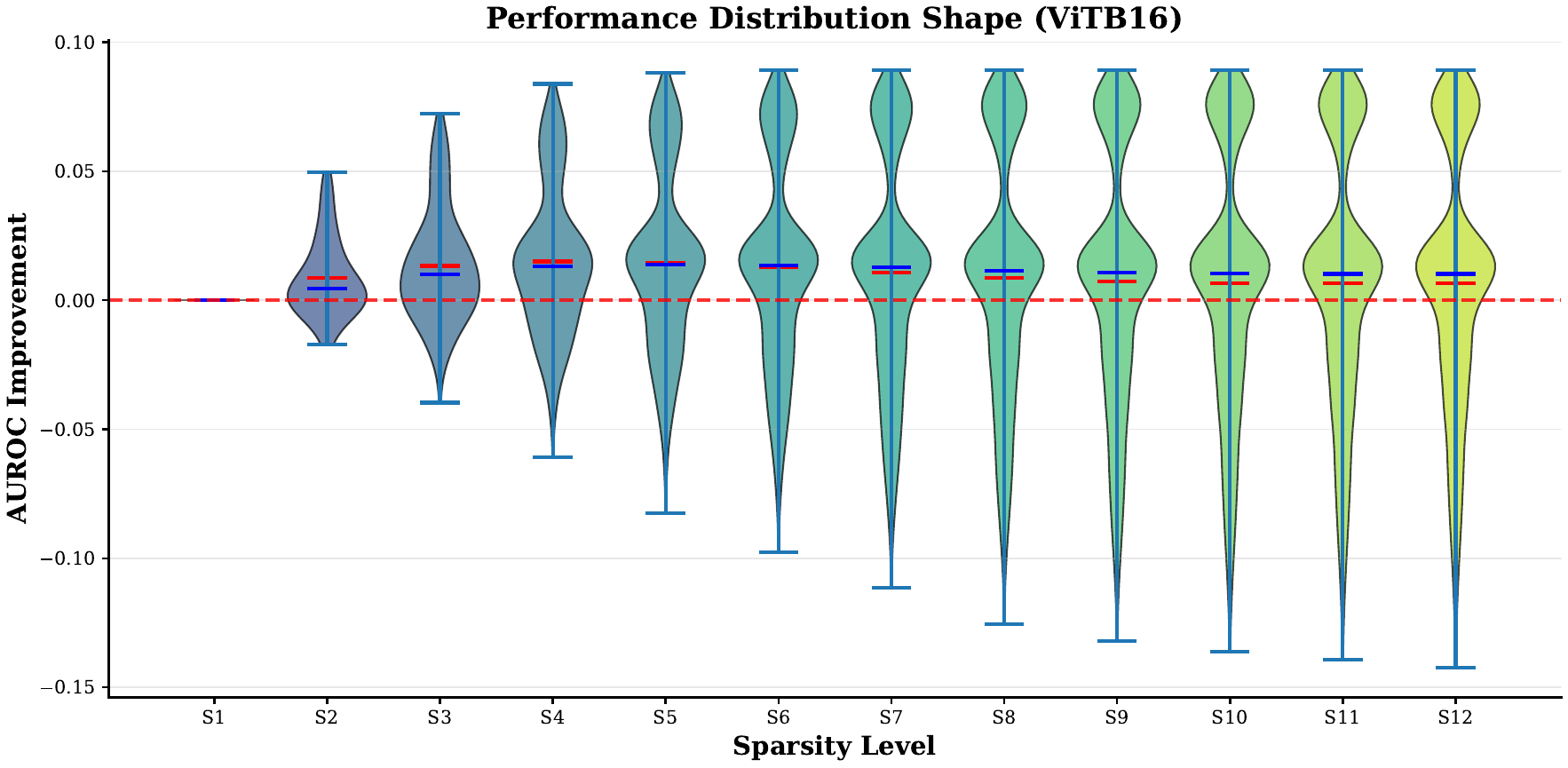}
        \caption{\textbf{Distribution of expert improvements for ViT-B/16.} 
        Improvements are smaller in magnitude compared to ViT-B/32 but remain positive and stable.}
        \label{fig:vitb16_violin}
    \end{minipage}
    \vspace{-3em}
\end{figure}

A complementary perspective on the random-selection analysis is provided by examining the performance distribution of individual experts across sparsity levels. For each sparsity level $s$, we compute the AUROC improvement $\Delta$ obtained by every expert when appended to the last layer, and summarize these distributions with violin plots. Figures~\ref{fig:vitb32_violin} and~\ref{fig:vitb16_violin} report the results for ViT-B/32 and ViT-B/16, respectively.

Two main observations emerge. First, across most sparsity levels, both the median and mean improvements are \emph{positive}, indicating that a substantial fraction of experts consistently outperform the last-layer baseline. This shows that the benefits of intermediate-layer fusion are not driven by a few isolated outliers, but instead arise from a broad set of experts that provide complementary and reliable gains. Second, the magnitude and dispersion of these improvements vary across architectures. ViT-B/32 exhibits a wider positive tail and higher median gains, suggesting stronger diversity among its experts and a greater chance that randomly selected combinations yield meaningful improvements. In contrast, ViT-B/16 shows smaller but more concentrated gains, indicating a more uniform expert population with lower variance in performance.

\FloatBarrier

\section{Theoretical Justification: The Value of Intermediate Layers}
\label{sec:theory}

This section is intended as a lightweight motivation for dynamic routing, not as a formal guarantee of FPR@95. The gap $\Delta$ below is an expected-score separation proxy that explains when adding an intermediate layer can improve an anchored expert; the empirical sections measure the actual thresholded OOD behavior.

\paragraph{Layer-wise Confidence and Separation.} 
Let $f_\ell(x)$ be the normalized representation at layer $\ell \in \{0,\dots,L-1\}$ of a frozen CLIP-style encoder and $\{t_c\}_{c=1}^C$ be text prototypes. We define the maximum-concept matching (MCM) confidence \cite{ming2022delvingoutofdistributiondetectionvisionlanguage} at temperature $\tau > 0$ as:
\begin{equation}
S_\ell(x;\tau) = \max_{c} \frac{\exp(\langle f_\ell(x), t_c \rangle / \tau)}{\sum_{c'} \exp(\langle f_\ell(x), t_{c'} \rangle / \tau)}.
\label{eq:mcm_layer}
\end{equation}
To quantify OOD detection performance, we define the \textit{separation contribution} of layer $\ell$ as the expected difference between in-distribution ($\mathbb{P}_{\text{ID}}$) and OOD ($\mathbb{P}_{\text{OOD}}$) scores:
\begin{equation}
\Delta(\ell) = \mathbb{E}_{x \sim \mathbb{P}_{\text{ID}}}[S_\ell(x;\tau)] - \mathbb{E}_{x \sim \mathbb{P}_{\text{OOD}}}[S_\ell(x;\tau)].
\label{eq:gap_layer}
\end{equation}

\paragraph{Anchored Expert Scores.} 
We anchor our selection to the final layer $\ell^\star = L-1$, which contains the strongest semantic signal. For any auxiliary subset $A \subseteq \{0,\dots,L-2\}$ where $|A| \le s-1$, the anchored expert score and its corresponding population separation gap $\Delta(A)$ are:
\begin{equation}
S_{A}^{\text{anch}}(x;\tau) = \frac{S_{\ell^\star} + \sum_{\ell \in A} S_\ell}{1+|A|}, \quad \Delta(A) = \frac{\Delta(\ell^\star) + \sum_{\ell \in A} \Delta(\ell)}{1+|A|}.
\label{eq:gap_decomp}
\end{equation}
While a \textit{static} method fixes $A$ for all $x$, our \textit{dynamic} approach selects $A(x)$ per-input. Larger $\Delta(A)$ implies better threshold separability for OOD detection.

\paragraph{Theorem 1 (Exact Improvement Condition).} 
\textit{For any $A \subseteq \{0,\dots,L-2\}$ and $\ell \notin A$, adding layer $\ell$ improves the separation gap if and only if its individual contribution exceeds the current anchored average:}
\begin{equation}
\Delta(A \cup \{\ell\}) > \Delta(A) \iff \Delta(\ell) > \Delta(A).
\label{eq:improvement_condition}
\end{equation}

\begin{proof}
Let $B = \Delta(\ell^\star) + \sum_{j \in A} \Delta(j)$. By \eqref{eq:gap_decomp}, we have $\Delta(A) = \frac{B}{1+|A|}$ and $\Delta(A \cup \{\ell\}) = \frac{B + \Delta(\ell)}{2+|A|}$. Then:
\begin{equation*}
\begin{aligned}
\frac{B + \Delta(\ell)}{2+|A|} > \frac{B}{1+|A|} &\iff (1+|A|)(B + \Delta(\ell)) > B(2+|A|) \\
&\iff (1+|A|)\Delta(\ell) > B \\
&\iff \Delta(\ell) > \frac{B}{1+|A|} = \Delta(A).
\end{aligned}
\end{equation*}
This completes the proof.
\end{proof}

\paragraph{Implications: why intermediate layers help.}
Equation~\eqref{eq:improvement_condition} formalizes the key mechanism: intermediate layers are beneficial \emph{only} when they provide higher ID--OOD separation than the current anchored expert.
Under distribution shift, some intermediate depths can increase separability (large $\Delta(\ell)$) while others can be neutral or harmful (small or negative $\Delta(\ell)$). Consequently, any \emph{static} subset $A$ inevitably underperforms on shifts where its chosen layers do not satisfy \eqref{eq:improvement_condition}. Dynamic routing addresses this by selecting, per input, auxiliary layers whose contributions are predicted to be beneficial using minimal outlier context.

\paragraph{Combinatorial Sparsity Motivation and a Theoretical Plot}
\label{sec:math_sparsity_plot}

Equation \eqref{eq:expert_count_main} shows that selecting up to $s$ layers from $L$ induces a combinatorial hypothesis class of size
$
K=\sum_{r=1}^{s}\binom{L-1}{r-1},
$
even under the mandatory inclusion constraint of the last layer. This motivates dynamic routing as an \emph{amortized} selection mechanism: a static method must commit to a single subset (or a fixed fusion rule) globally, whereas routing learns a mapping $x\mapsto \widehat{i}(x)$ that selects among many sparse candidates without exhaustive search.

\section{Layer-Importance Diagnostic}
\label{appendix:layer_importance}

To connect the preceding motivation to the deployed model, we compute a post-hoc sparsity-constrained Shapley diagnostic for the fixed Table~\ref{tab:main_results} \name{} recipe (ViT-B/16, WebOE, 4-shot ID, $s{=}9$, top-$32$, $N{=}5$ seeds $0$--$4$). The final layer is held as the mandatory anchor, and the diagnostic measures the marginal contribution of intermediate layers within the same expert family used by the detector. This analysis is explanatory only: it is run after fixing the detector and is not used for hyperparameter selection.

\insightbox{Layer value is shift-dependent}
{Texture and fine-grained natural shifts benefit from early and middle layers, while scene datasets penalize late intermediate layers. This supports dynamic routing over a fixed intermediate-layer subset: useful auxiliary evidence changes with the OOD family.}

\begin{figure}[H]
    \centering
    \includegraphics[width=\linewidth]{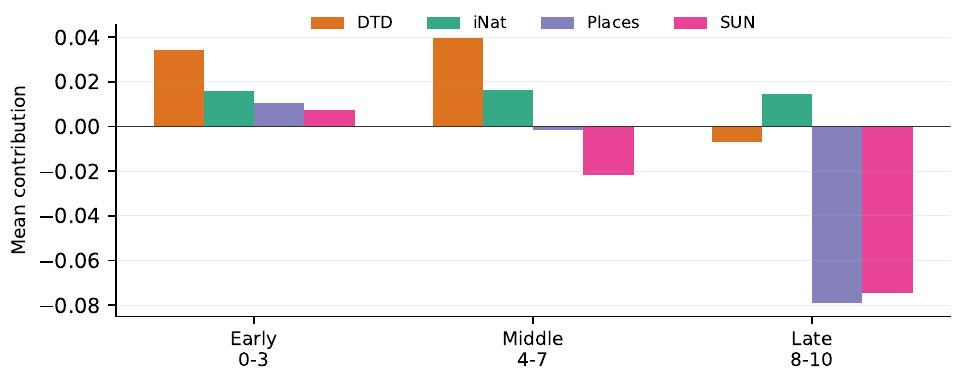}
    \caption{\textbf{Intermediate-layer contribution by OOD family.} Bars report the mean Shapley contribution of early, middle, and late intermediate layers to the Table~\ref{tab:main_results} \name{} separation objective. Positive values indicate improved ID--OOD separation relative to the final-layer anchor; negative values indicate harmful auxiliary evidence.}
    \label{fig:layer_shapley_depth}
    \vspace{-0.8em}
\end{figure}

\section{Interpretability: Sliding-Window OOD Heatmaps with \name{}}
\label{app:sliding_window_heatmaps}

To qualitatively assess where \name{} detects OOD evidence, we introduce a sliding-window
heatmap procedure that produces spatial OOD maps from a frozen CLIP backbone and our trained router. 
Given an input image, we extract overlapping crops on a regular grid
(window size $w=224$, stride $24$), and process each crop independently to avoid GPU
memory issues. For each crop, we obtain all-layer CLIP logits, compute expert scores via
the MCM confidence for each expert, and apply \name{} router to produce a scalar
OOD score. In parallel, we compute a baseline score using only the last-layer MCM.

To make scores comparable across images and between \name{} and the baseline, 
we normalize them using their respective decision thresholds 
$\tau_{\text{Dyn}}$ and $\tau_{\text{Base}}$, each defined as the 
$\mathbf{5^{\text{th}}}$ percentile of the ID score distribution, which retains 95\% of ID samples as ID under a confidence-style score. 
For a crop-level OOD score $s$, we compute a signed distance
\[
    \tilde{s} = \frac{s - \tau}{Z},
\]
where $Z$ rescales positive (ID) and negative (OOD) deviations into the interval $[-1, 1]$. 
Positive values ($\tilde{s}>0$) correspond to ID-like regions, while negative values 
($\tilde{s}<0$) indicate OOD evidence; the magnitude reflects the confidence of the assignment. 
The normalized scores are then projected back onto the image plane by averaging across 
overlapping windows and applying a mild Gaussian blur, yielding a dense spatial heatmap 
that is resized to the original resolution and overlaid on the RGB image 
	(red = ID, blue = OOD).

\begin{figure}[t]
    \centering
    \includegraphics[width=1\linewidth]{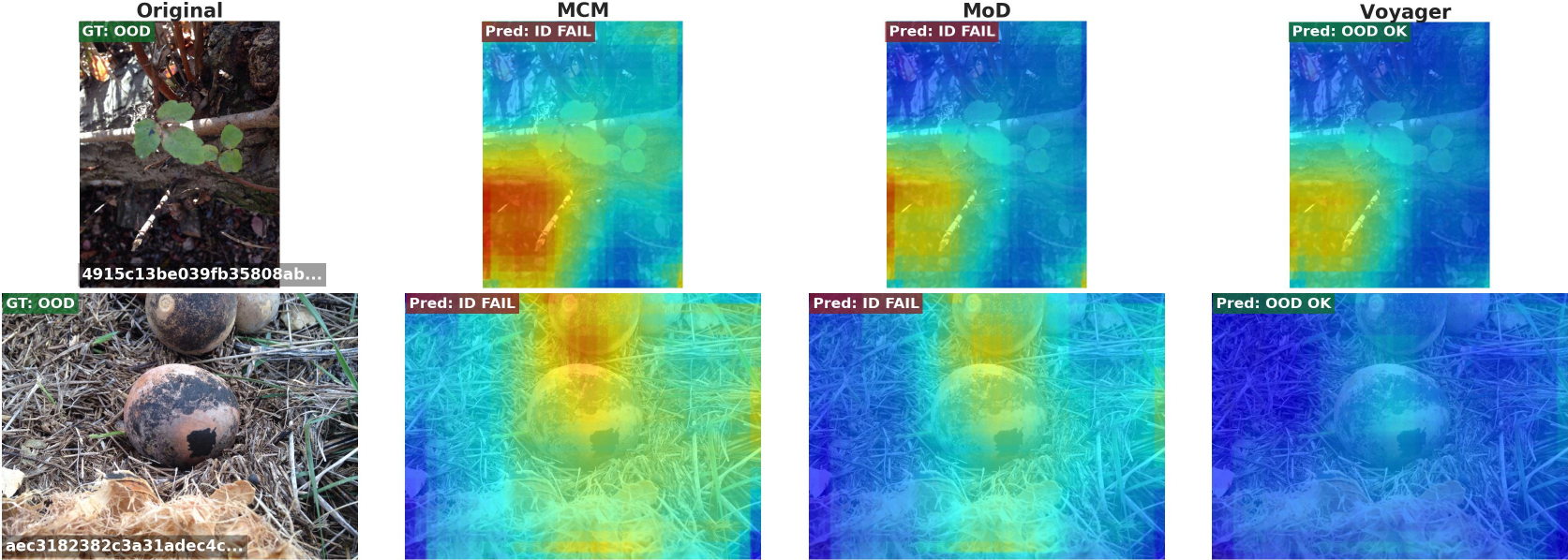}
    \caption{\textbf{Sliding-window OOD maps.} MCM, MoD, and \name{} on benchmark OOD images where MCM/MoD predict ID but \name{} predicts OOD at the FPR@95 threshold. Red is ID-like evidence; blue is OOD evidence.}
    \label{fig:sliding_heatmaps_appendix}
    \vspace{-1.0em}
\end{figure}

Figure~\ref{fig:sliding_heatmaps_appendix} highlights cases where older detectors over-trust local ID-like plant and texture evidence. MCM and MoD keep enough red evidence to classify the full image as ID, while the \name{} ensemble lowers the global score below the OOD threshold.

\subsection{Extra Results}

\begin{figure}
    \centering
    \includegraphics[width=1\linewidth]{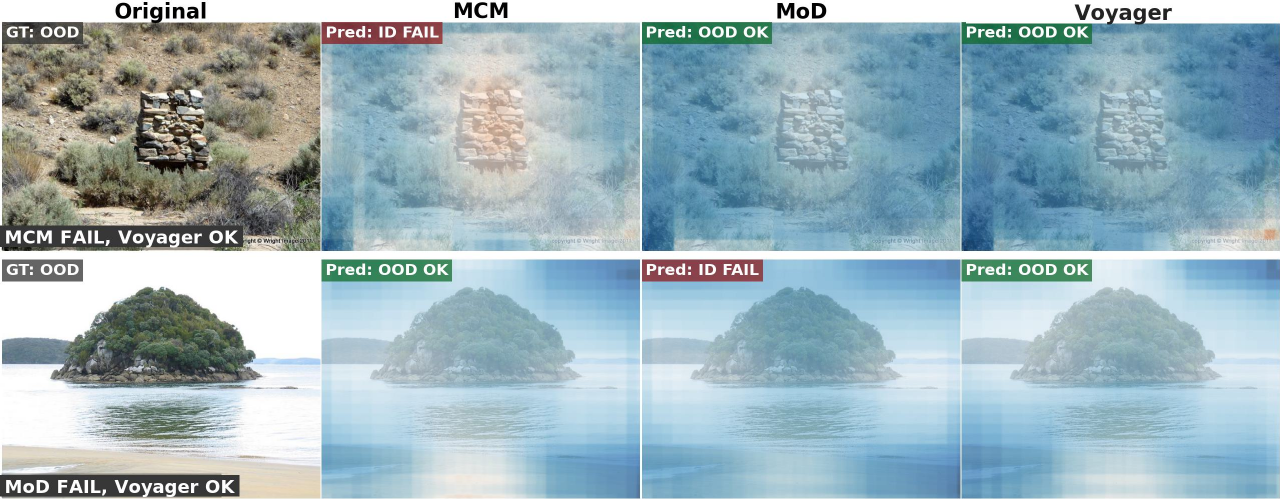}
    \caption{\textbf{Sliding-window OOD maps (representative WebOE-trained ViT-B/16 router) -- additional benchmark examples.}}
    \label{fig:sliding_window_extra_1}
\end{figure}

\begin{figure}
    \centering
    \includegraphics[width=1\linewidth]{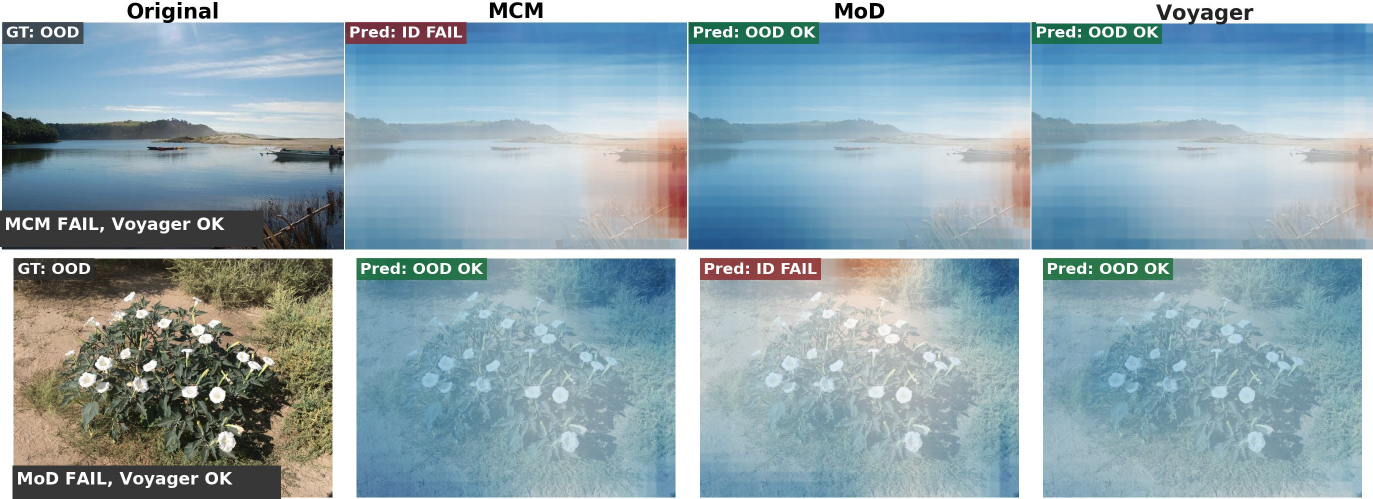}
    \caption{\textbf{Sliding-window OOD maps (representative WebOE-trained ViT-B/16 router) -- additional benchmark examples.}}
    \label{fig:sliding_window_extra_2}
\end{figure}

\begin{figure}
    \centering
    \includegraphics[width=1\linewidth]{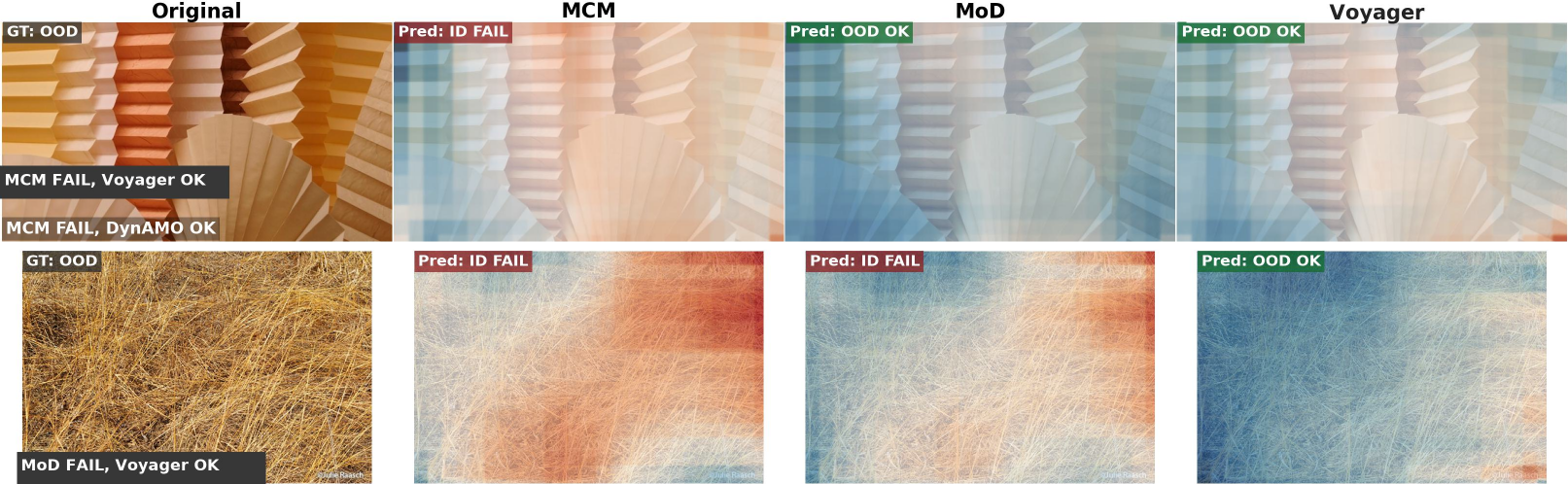}
    \caption{\textbf{Sliding-window OOD maps (representative WebOE-trained ViT-B/16 router) -- additional texture examples.}}
    \label{fig:sliding_window_extra_3}
\end{figure}

\FloatBarrier

\section{Limitations}
\label{appendix:limitation}

While \name{} consistently improves out-of-distribution detection, it is a few-shot detector rather than a zero-shot one: it trains a lightweight router on a few labeled ID images together with a proxy OOD context. Crucially, that context requires no manual curation---\textit{WebOE} retrieves it automatically from Wikimedia Commons using only the ID label names---and our experiments indicate that the gain is not an artifact of the proxy: the selection-mechanism ablation (Sec.~\ref{appendix:selector}) shows that, given the \emph{same} proxy context, the learned input-conditioned router ($18.8$ FPR@95) substantially outperforms the best score-based selection rule ($28.9$), the fixed-context expert ($34.6$), and the last-layer baseline ($45.6$). The improvement therefore comes from the learned gate rather than from access to curated outlier data. The residual cost is the modest supervision the router still needs---a handful of ID labels and the automatically retrieved proxy context---together with its reliance on the frozen backbone exposing informative intermediate-layer scores; relaxing these requirements toward a fully self-supervised router is a natural direction for future work.

\end{document}